\documentclass[10pt,twocolumn,letterpaper]{article}
\usepackage[table,dvipsnames]{xcolor}
\usepackage[pagenumbers]{cvpr} % To force page numbers, e.g. for an arXiv version

\definecolor{cvprblue}{rgb}{0.21,0.49,0.74}

\newlength\savewidth\newcommand\shline{\noalign{\global\savewidth\arrayrulewidth
  \global\arrayrulewidth 1pt}\hline\noalign{\global\arrayrulewidth\savewidth}}
\newcommand{\tablestyle}[2]{\setlength{\tabcolsep}{#1}\renewcommand{\arraystretch}{#2}\centering\footnotesize}

\definecolor{cvprblue}{rgb}{0.21,0.49,0.74}
\usepackage[pagebackref,breaklinks,colorlinks,allcolors=cvprblue]{hyperref}

\usepackage{array}
\usepackage{multirow}
\usepackage{amssymb}
\usepackage{caption}
\usepackage{subcaption}
\usepackage{pgfplots}
\pgfplotsset{compat=1.18}

\usepackage{algpseudocode}
\usepackage[ruled,vlined]{algorithm2e} 
\usepackage{multicol}
\usepackage{setspace}

\usepackage{listings}
\usepackage{placeins}

\usepackage{xcolor}
\definecolor{codegreen}{rgb}{0,0.6,0}
\definecolor{codegray}{rgb}{0.5,0.5,0.5}
\definecolor{codepurple}{rgb}{0.58,0,0.82}
\definecolor{backcolour}{rgb}{0.95,0.95,0.92}

\lstdefinestyle{mystyle}{
    commentstyle=\color{codegreen},
    keywordstyle=\color{magenta},
    numberstyle=\tiny\color{codegray},
    stringstyle=\color{codepurple},
    basicstyle=\ttfamily\footnotesize,
    breakatwhitespace=false,         
    breaklines=true,                 
    captionpos=b,                    
    keepspaces=true,                 
    numbers=left,                    
    numbersep=5pt,                  
    showspaces=false,                
    showstringspaces=false,
    showtabs=false,                  
    tabsize=2
}

\usepackage{booktabs}
\usepackage{pifont}
\usepackage{longtable}
\usepackage{supertabular}

\newcommand{\cmark}{\ding{51}}%
\newcommand{\xmark}{\ding{55}}%

\newcolumntype{x}[1]{>{\centering\arraybackslash}p{#1pt}}
\newcolumntype{y}[1]{>{\raggedright\arraybackslash}p{#1pt}}
\newcolumntype{z}[1]{>{\raggedleft\arraybackslash}p{#1pt}}

\newcommand{\app}{\raise.17ex\hbox{$\scriptstyle\sim$}}

\definecolor{baselinecolor}{gray}{.9}

\RequirePackage[breakable,skins]{tcolorbox}
\RequirePackage{xcolor}
\definecolor{colorcommentfg}{RGB}{0,63,87}
\colorlet{colorchangebg}{black!2}
\colorlet{colorchangeframe}{black!20}
\definecolor{lightpastelpurple}{rgb}{0.69, 0.61, 0.85}
\definecolor{mediumpurple}{rgb}{0.58, 0.44, 0.86}

\definecolor{colorcommentbg}{HTML}{CCCCFF}
\definecolor{colorcommentframe}{HTML}{76608A}

\newcounter{reviewcomment@counter}[section]
\newenvironment{contentagnosticlayout}[1][]{\refstepcounter{reviewcomment@counter}
	\begin{tcolorbox}[adjusted title={Anonymous Layout Example}, fonttitle={\bfseries\footnotesize}, fontupper=\footnotesize, colback={colorcommentbg!40}, colframe={colorcommentframe!90},coltitle={white},#1]
}{\end{tcolorbox}}

\definecolor{colorcommentbg_qualityprompt}{HTML}{A0522D}
\definecolor{colorcommentframe_qualityprompt}{HTML}{6D1F00}

\definecolor{colorcommentbg_pkprompt}{HTML}{5D8AA8}
\definecolor{colorcommentframe_pkprompt}{HTML}{0093AF}

\newcommand{\designbenchmark}{\textsc{Design-Multi-Layer-Bench}\xspace}
\newcommand{\photobenchmark}{\textsc{Photo-Multi-Layer-Bench}\xspace}

\newcommand\codeurl[1]{{{\color{blue}{\url{#1}}}}}

\title{
ART: Anonymous Region Transformer for \\ Variable Multi-Layer Transparent Image Generation
}

\author{
\small Yifan Pu\textsuperscript{$\dagger$} \; Yiming Zhao\textsuperscript{$\dagger$} \; Zhicong Tang\; Ruihong Yin \; Haoxing Ye \; Yuhui Yuan\textsuperscript{$\dagger\ddagger$}  \; Dong Chen\textsuperscript{$\dagger\ddagger$}   \; Jianmin Bao \\\small Sirui Zhang \;\; Yanbin Wang\;\; Lin Liang \; Lijuan Wang \; Ji Li \; \; Xiu Li \; Zhouhui Lian \; Gao Huang\;  \; Baining Guo \\
{\small $^\dagger$equal technical contribution \qquad $^\ddagger$project lead}\\[1mm]
\small Microsoft Research Asia \;  Tsinghua University \;  Peking University \; University of Science and Technology of China \\
{\small\codeurl{https://art-msra.github.io}\vspace{-4mm}}
}

\begin{document}
\maketitle

\begin{abstract}
Multi-layer image generation is a fundamental task that enables users to isolate, select, and edit specific image layers, thereby revolutionizing interactions with generative models. In this paper, we introduce the Anonymous Region Transformer (ART), which facilitates the direct generation of variable multi-layer transparent images based on a global text prompt and an anonymous region layout.
Inspired by Schema theory\footnote{Schema theory~\cite{bartlett1995remembering,rumelhart2017schemata} suggests that knowledge is organized in frameworks (schemas) that enable people to interpret and learn from new information by linking it to prior knowledge.}, this anonymous region layout allows the generative model to autonomously determine which set of visual tokens should align with which text tokens, which is in contrast to the previously dominant semantic layout for the image generation task.
In addition, the layer-wise region crop mechanism, which only selects the visual tokens belonging to each anonymous region, significantly reduces attention computation costs and enables the efficient generation of images with numerous distinct layers (\eg, 50+). When compared to the full attention approach, our method is over 12 times faster and exhibits fewer layer conflicts.
Furthermore, we propose a high-quality multi-layer transparent image autoencoder that supports the direct encoding and decoding of the transparency of variable multi-layer images in a joint manner.
By enabling precise control and scalable layer generation, ART establishes a new paradigm for interactive content creation.
\end{abstract}    
\section{Introduction}
\label{sec:intro}

Diffusion-based generative models have shown tremendous success in producing high-quality images from given text prompts~\cite{podell2023sdxl,esser2024scaling,betker2023improving,saharia2022photorealistic}. These models are typically limited to producing entire images in a single, unified layer, which restricts the ability to edit or manipulate specific elements independently. This limitation presents significant challenges in fields like graphic design and digital art, where creators frequently rely on layer-by-layer control to construct and refine complex compositions.

\begin{figure}[!t]
\begin{minipage}[!t]{1\linewidth}
\begin{subfigure}[b]{1\textwidth}
\centering
\includegraphics[width=1.0\textwidth]{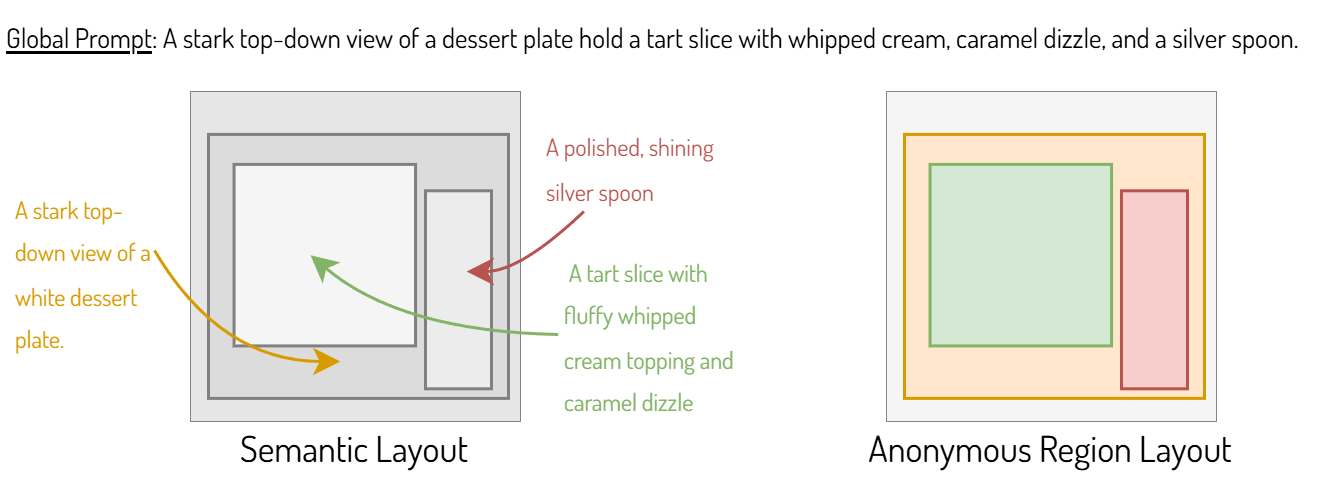}
\vspace{-3mm}
\end{subfigure}
\end{minipage}
\vspace{-3mm}
\caption{\footnotesize{\textbf{Semantic Layout \vs Anonymous Region Layout}. The conventional semantic layout requires specifying what objects to generate in each given region, whereas our anonymous region layout only identifies where the important regions are. 
People can leverage the prior knowledge, activated by the global prompt, to intuitively infer the semantic label of each anonymous region. The generative model also learns to harness this capability and autonomously determine what to generate in each region.
}}
\label{fig:anonymous_layout}
\vspace{-3mm}
\end{figure}

This paper presents Anonymous Region Transformer for multi-layer transparent image generation. The key ingredient of the anonymous region transformer is the anonymous region layout, which solely consists of a set of anonymous rectangular regions without any region-wise prompt annotations, as shown in \Cref{fig:anonymous_layout}. This is unlike the conventional semantic layout for text-to-image generation~\cite{li2023gligen,yang2024mastering,yang2023reco}, which requires clearly specify both the global prompt for the entire image and the location and region-wise prompts for each region\footnote{We use `region' and `layer' interchangeably in this paper.}. The drawback of the conventional layout is that it heavily relies on human labor for creating the layout and this process can be very labor intensive, especially when handling tens or even hundreds of regions on a canvas, a common scenario in graphic design generation. The anonymous region transformer significantly reduces the human labor by allowing the generative model to perform the visual planning task of determining which objects to generate in each anonymous region based on the global prompt.
The core insight behind the anonymous region layout is to \emph{give more control to the generative models, while ensuring that users have great control over manipulating the multi-layered output.}

\begin{figure}[!t]
\begin{minipage}[!t]{1\linewidth}
\begin{subfigure}[b]{0.19\textwidth}
\centering
\includegraphics[width=1\textwidth]{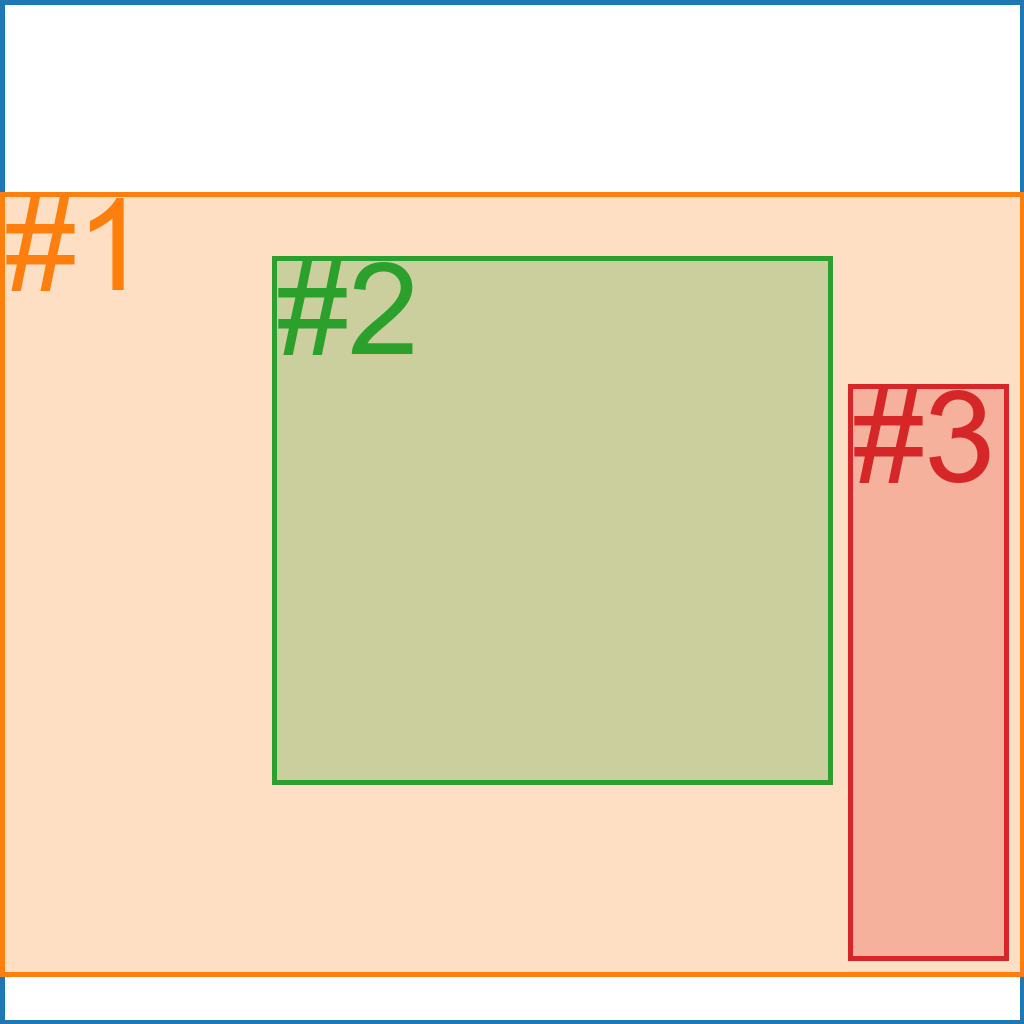}
\vspace{-3mm}
\caption*{Layout}
\end{subfigure}
\begin{subfigure}[b]{0.19\textwidth}
\centering
\includegraphics[width=1\textwidth]{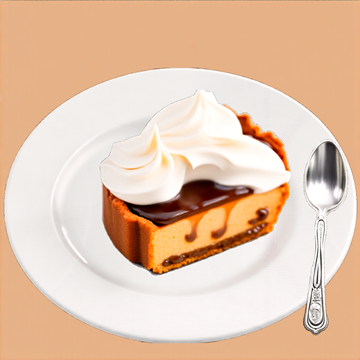}
\vspace{-3mm}
\caption*{Composed.}
\end{subfigure}
\begin{subfigure}[b]{0.19\textwidth}
\centering
\includegraphics[width=1\textwidth]{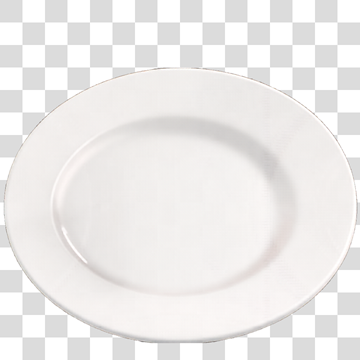}
\vspace{-3mm}
\caption*{Region\#1}
\end{subfigure}
\begin{subfigure}[b]{0.19\textwidth}
\centering
\includegraphics[width=1\textwidth]{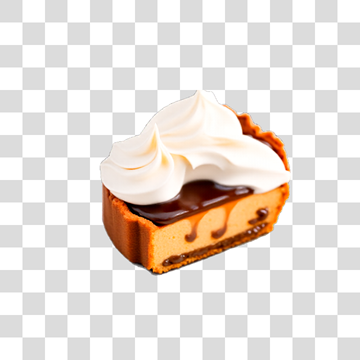}
\vspace{-3mm}
\caption*{Region\#2}
\end{subfigure}
\begin{subfigure}[b]{0.19\textwidth}
\centering
\includegraphics[width=1\textwidth]{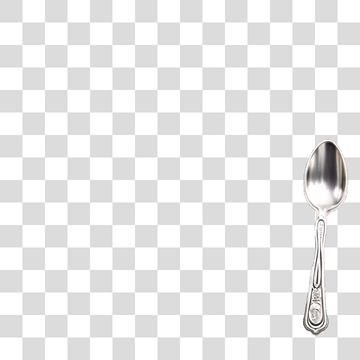}
\vspace{-3mm}
\caption*{Region\#3}
\end{subfigure}
\hfill
\begin{subfigure}[b]{1\textwidth}
\centering
\includegraphics[width=1\textwidth,height=14mm]{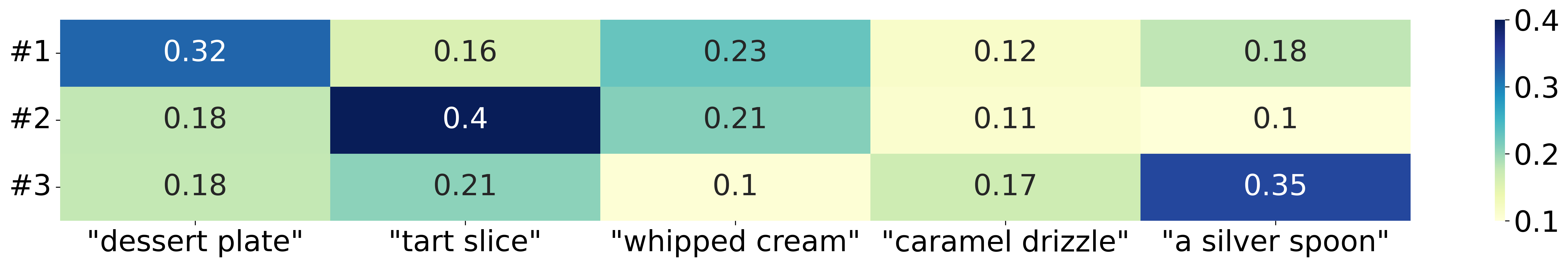}
\vspace{-4mm}
\caption*{Attention Maps between Anonymous Region and Text}
\end{subfigure}
\end{minipage}
\vspace{-2mm}
\caption{\footnotesize{
\textbf{Visual planning capability of our Anonymous Region Transformer}. We visualize the averaged attention maps of all visual tokens within the same anonymous region (as Query) attending to the entities within the global prompt text tokens (as Key and Value). These attention maps reveal that each anonymous region assigns the majority of attention weights to one of the major objects identified in the given text prompt. }}
\label{fig:attention_analysis}
\vspace{-3mm}
\end{figure}

A natural question arises regarding how the anonymous region layout can function effectively without region-wise prompts, especially given that these prompts are central to conventional semantic layout approaches. This effectiveness can be explained by Schema Theory~\cite{bartlett1995remembering,rumelhart2017schemata,kant1934critique,axelrod1973schema}, a well-established cognitive framework that helps bridge the gap between abstract concepts (such as \emph{plate} or \emph{spoon}) and specific sensory experiences (such as \emph{layout}). It suggests that people can infer each region's semantic label based on their prior knowledge activated by a global prompt. In our case, we find that the effectiveness of the anonymous-region layout for multi-layer image generation tasks stems from the Transformer model’s ability to autonomously identify semantic labels for each layer through interactions between text tokens and visual tokens. The generative model learns to capture the prior knowledge similar to Schema Theory, enabling it to determine which set of visual tokens (from an anonymous region) attends to which text tokens (representing different entities), as shown in \Cref{fig:attention_analysis}. Our experiments further demonstrate that adding additional region-wise prompts for each layer does not necessarily improve the results and can even diminish coherence across layers.

The anonymous region transformer offers several key advantages over the conventional approach for multi-layer transparent image generation. \underline{First}, it ensures better coherence across different layers. We observe that, in the semantic layout, regional visual tokens struggle to balance attention weights between region-wise text tokens (to ensure \emph{prompt following}) and the corresponding global visual tokens located at the same position (ensure \emph{coherence}). This difficulty arises from a semantic gap between the global visual tokens and region-wise visual tokens as they are forced to attend different text tokens. In contrast, our anonymous region layout enables all regional visual tokens and global visual tokens to attend to the same set of global text tokens, thereby closing this gap. \underline{Second}, annotating the anonymous-region layout is more scalable, especially for native multi-layer graphic design images. We can easily generate a large number of high-quality anonymous-region layouts, whereas recaptioning each region is non-trivial and often suffers from significant noise due the semantic gap between captioning a crop conditioned on an entire image and captioning only a small crop. \underline{Third}, by focusing on the anonymous regions within each layer, we can significantly reduce computation costs and enables the efficient generation of images with numerous distinct layers (e.g., 50+).

\begin{figure}[!t]
\begin{minipage}[!t]{1\linewidth}
\begin{subfigure}[b]{1\textwidth}
\centering
\includegraphics[width=1\textwidth]{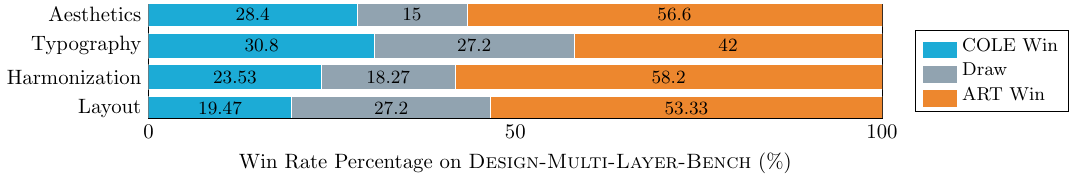}
\vspace{-3mm}
\end{subfigure}
\begin{subfigure}[b]{1\textwidth}
\centering
\includegraphics[width=1\textwidth]{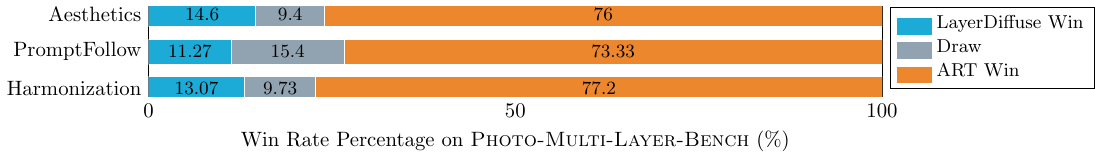}
\vspace{-3mm}
\end{subfigure}
\end{minipage}
\vspace{-3mm}
\caption{\footnotesize{
\textbf{ART \vs previous SOTA} in multi-layer transparent image generation: user study results across different domains. ART significantly outperforms LayerDiffuse~\cite{zhang2024transparent} in the photorealistic domain and COLE~\cite{jia2023cole} in the graphic-design domain across multiple aspects.
}}
\label{fig:user_study}
\vspace{-3mm}
\end{figure}

Our methodology consists of three key components: the Multi-layer Transparent Image Autoencoder, the Anonymous Region Transformer, and the Anonymous Region Layout Planner. The Multi-layer Transparent Autoencoder encodes and decodes a variable number of transparent layers at different resolutions using a sequence of latent visual tokens. The Anonymous Region Transformer concurrently generates a global reference image, a background image, and multiple cropped transparent foreground layers from Gaussian noise conditioned on the anonymous region layout. The Anonymous Region Layout Planner predicts a set of anonymous bounding boxes based on the user-provided text prompt.
Compared existing methods in multi-layer image generation—such as Text2Layer~\cite{zhang2023text2layer}, LayerDiff~\cite{huang2024layerdiff}, and LayerDiffuse~\cite{zhang2024transparent}-the key difference is that these methods can produce only a limited number of transparent layers at fixed resolutions.
Additionally, unlike the COLE~\cite{jia2023cole} and OpenCOLE~\cite{inoue2024opencole}, which apply a cascade of diffusion models to generate layers sequentially, our method generates all transparent layers and the reference image simultaneously in an \textit{end-to-end} manner, ensuring a better global harmonization across different layers. The experimental results demonstrate the advantages of our approach over previous methods, and we report the user study results in \Cref{fig:user_study}.

In summary, this paper not only proposes a novel approach to multi-layer transparent image generation, but also opens up numerous possibilities for future research and applications. Our main contributions are as follows:
\begin{enumerate}
\item We are the first to propose a novel pipeline for multi-layer transparent image generation that supports generating a variable number of layers at variable resolution.
\item We introduce the anonymous region layout, which offers several key advantages over conventional semantic layout for multi-layer transparent image generation.
\item We empirically validate the effectiveness of our method. Compared to the previous state-of-the-art methods, our algorithm generates multi-layer transparent images with higher quality and a greater number of layers.
\end{enumerate}
\section{Related work}
\label{sec:related_work}
\noindent\textbf{Multi-Layer Transparent Image Generation} has primarily been approached through two different paths. The first path focuses on generating all image layers simultaneously. Along this path, Text2Layer~\cite{zhang2023text2layer} adapts the Stable Diffusion model into a two-layer generation model, enabling the simultaneous generation of a background layer accompanied by a foreground layer. LayerDiff~\cite{huang2024layerdiff} designs a layer-collaborative diffusion model to generate up to four layers at once under the guidance of both global prompts and layer prompts.
The second path generates multiple image layers sequentially. For instance, LayerDiffuse~\cite{zhang2024transparent} introduces a background-conditioned transparent layer generation model, which generates image layers iteratively. COLE~\cite{jia2023cole} and OpenCOLE~\cite{inoue2024opencole} start from a brief user-provided prompt and employ multiple LLMs and diffusion models to generate each element within the final image step by step.
Unlike most of the aforementioned works, which only support generating a limited number of transparent layers, our approach allows for the generation of tens of transparent layers using an anonymous region transformer design. We also empirically demonstrate the advantages of our approach over these methods for photorealistic and design-oriented multi-layer image generation tasks.

\noindent\textbf{Layout Generation and Layout Control} for image generation tasks have attracted significant attention due to their broader applications. We can categorize most existing efforts into two groups: designing better layout generation models and controlling image generation with a given layout prior.
The first approach focuses on generating a reasonable layout given a set of visual elements. For example, Graphist~\cite{cheng2024graphic}, Visual Layout Composer~\cite{shabani2024visual}, and MarkupDM~\cite{kikuchi2024multimodal} propose different methods to generate layouts based on a set of transparent visual layers. Readers can refer to~\cite{feng2024layoutgpt,yamaguchi2021canvasvae,inoue2023layoutdm,chai2023layoutdm,hui2023unifying,kong2022blt,cheng2023play,tang2023layoutnuwa,jiang2022coarse,jiang2023layoutformer++,weng2024desigen,wang2023dolfin,guerreiro2025layoutflow,yang2024posterllava,inoue2023towards,chen2024textlap,fontanella2024generating,braunstein2024slayr} for more discussion on the development of various layout generation models.
In the second approach, researchers focus on enhancing the compositional generation capability of diffusion models by specifying what objects to generate and where to place them on the canvas. Several representative works include GLIGEN~\cite{li2023gligen}, InstanceDiffusion~\cite{wang2024instancediffusion}, and MS-Diffusion~\cite{wang2024msdiffusion}, which introduce different methods to inject positional information into diffusion models. Other efforts, such as~\cite{bar2023multidiffusion,yang2024mastering,kim2023dense,omost,sarukkai2024collage,zhang2024itercomp}, propose training-free schemes, post-training schemes, or harmonization enhancement designs. Among these efforts, LayoutGPT~\cite{feng2024layoutgpt} and TextLap~\cite{chen2024textlap} are the closest works that support predicting the semantic layout from a global text prompt. We empirically demonstrate the advantages of our anonymous region layout planner on multi-layer transparent image generation.
\section{Approach}

\begin{figure*}[!t]
\begin{minipage}[!t]{1\linewidth}
\begin{subfigure}[b]{0.57\textwidth}
\centering
\includegraphics[width=1\textwidth]{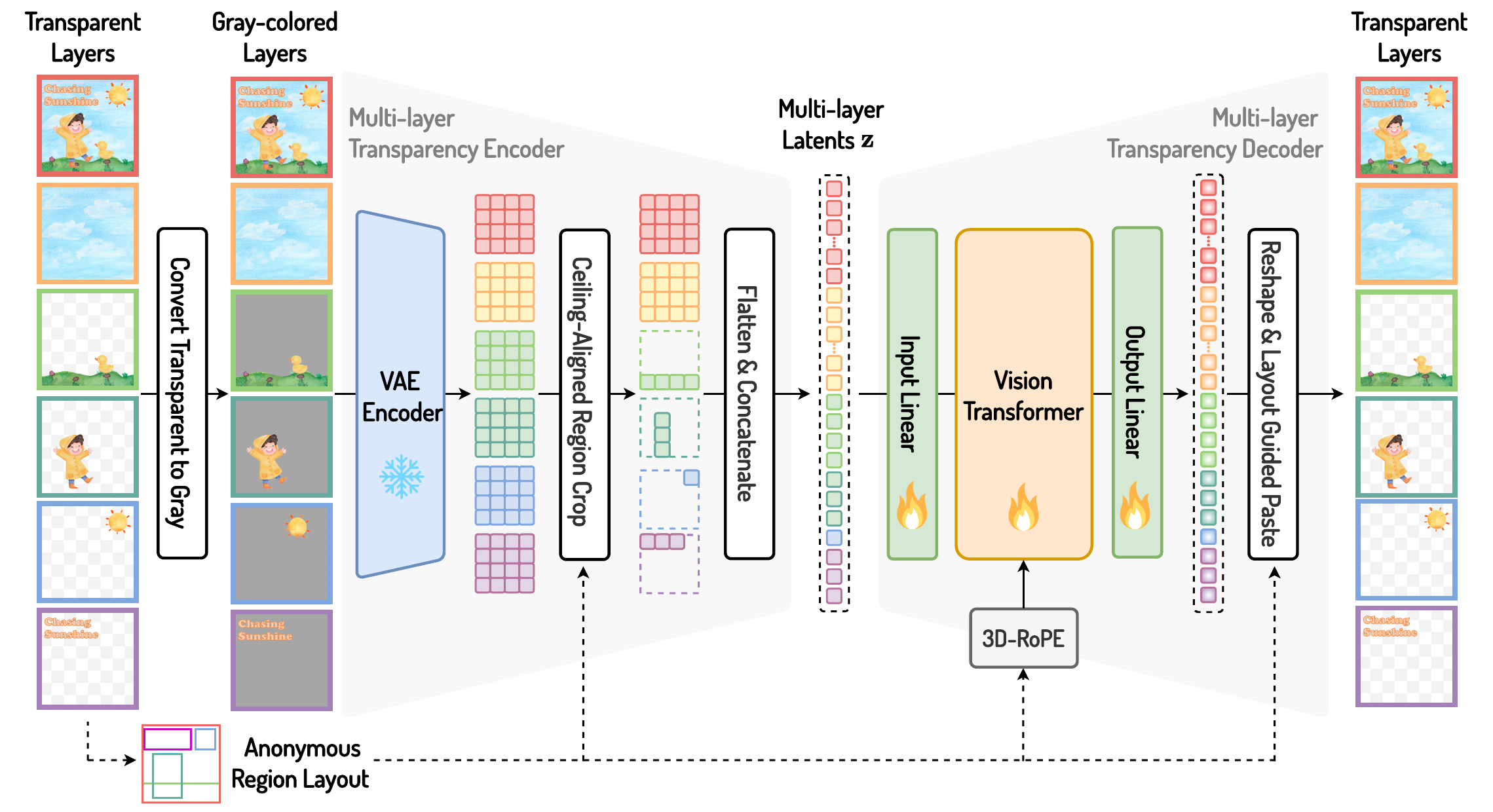}
\vspace{-3mm}
\caption{Multi-Layer Transparent Image Autoencoder}
\label{fig:framework_a}
\end{subfigure}
\hfill
\begin{subfigure}[b]{0.39\textwidth}
\centering
\includegraphics[width=1\textwidth]{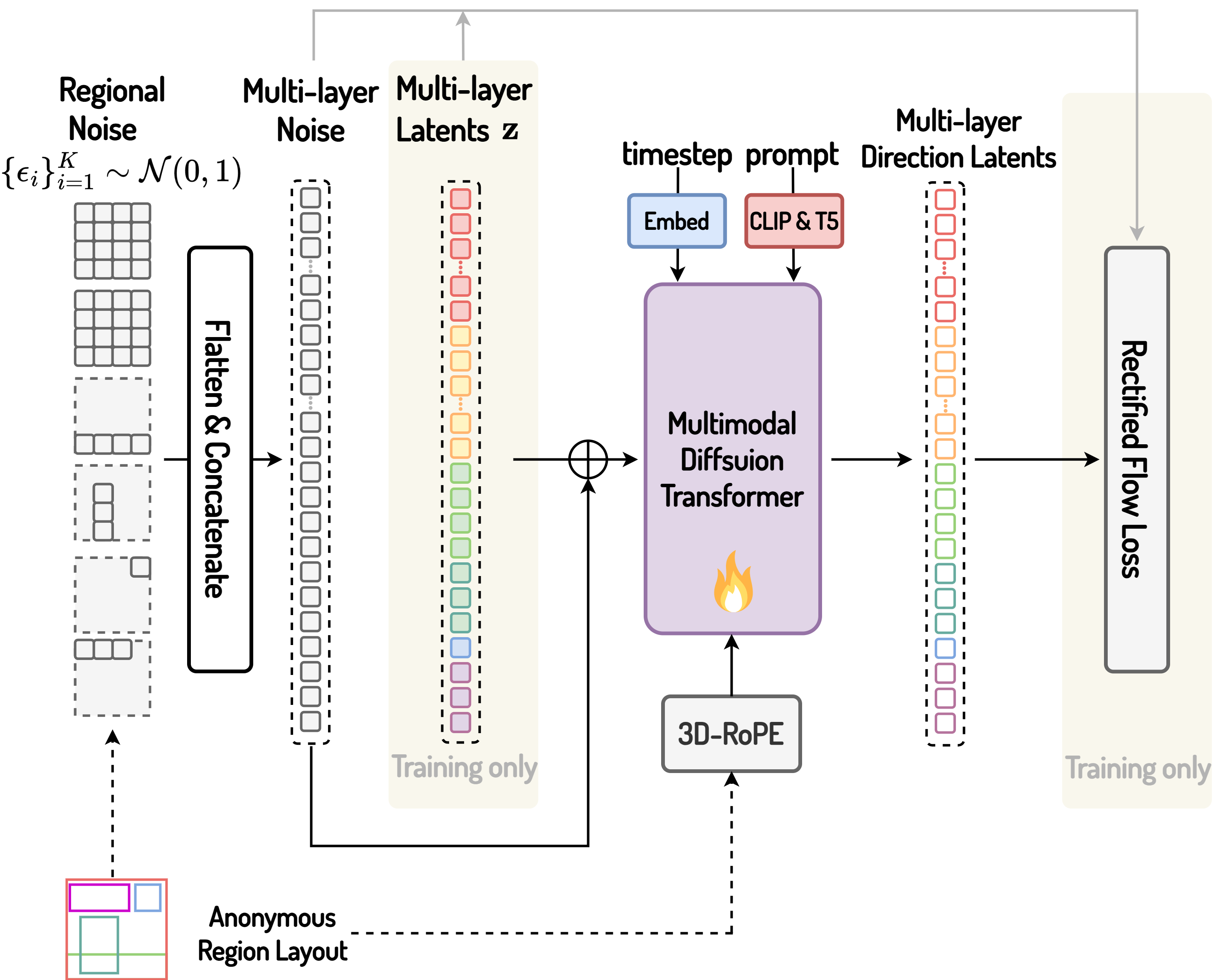}
\vspace{-3mm}
\caption{\footnotesize{Anonymous Region Transformer}}
\label{fig:framework_b}
\end{subfigure}
\end{minipage}
\vspace{-2mm}
\caption{\footnotesize{
(a)~\textbf{Multi-layer Transparent Image Autoencoder} directly encodes each layer of the multi-layer image, accompanied by the entire composed image, into latent space and jointly decodes the multi-layer latent tokens into RGBA transparent image layers.
(b)~\textbf{Anonymous Region Transformer (ART)} performs denoising diffusion on the noisy multi-layer latents corresponding to a variable number of transparent layers jointly.
}}
\label{fig:framework}
\vspace{-2mm}
\end{figure*}

The conventional text-to-image model~\cite{podell2023sdxl,esser2024scaling,betker2023improving,saharia2022photorealistic, flux} supports only a single, unified image generation from a global prompt. Our approach enables diffusion transformer-based models to jointly generate images with multiple transparent layers conditioned on an anonymous region layout provided by the user or predicted by an LLM. The entire framework consists of three key components: the \emph{Multi-layer Transparent Autoencoder} (\Cref{sec:method:ml_vae}), which jointly encodes and decodes multi-layer images and their corresponding latent representations; the \emph{Anonymous Region Transformer} (\Cref{sec:method:mmdit}), which concurrently generates a global reference image, a background image, and multiple RGBA transparent foreground image layers from a sequence of layout-guided noisy tokens; and the \emph{Anonymous Region Layout Planner} (\Cref{{sec:method:planner}}), which predicts a set of anonymous bounding boxes given the user-provided text prompt. The technical details are presented as follows.

\subsection{Multi-Layer Transparent Image Autoencoder} \label{sec:method:ml_vae}

A multi-layer transparent image consists of an RGB background layer $\mathbf{I}_\text{bg} \in \mathbb{R}^{H \times W \times 3}$, and a variable number $K$ of RGBA foreground layers, $\{\mathbf{I}_\text{fg}^{i} \in \mathbb{R}^{H_i \times W_i \times 4}\}_{i=1}^K$. The corresponding merged image $\mathbf{I}_{\text{mg}} \in \mathbb{R}^{H \times W \times 3}$ can be obtained by integrating $\mathbf{I}_\text{bg}$ as the base layer and overlaying all $\mathbf{I}_\text{fg}^{i}$ layers according to a predefined layout. We use $\mathbf{L}=\{x_c^i, y_c^i, H_i, W_i\}_{i=1}^K$ to represent the anonymous region layout of all $K$ foreground layers. Here, ${x_c^i, y_c^i}$ and ${H_i, W_i}$ denote the center coordinates and the height and width of the bounding box that encapsulates the $i$-th transparent foreground layer. It is worth noting that the anonymous region layout $\mathbf{L}$ is inherently encoded in the alpha channel of each foreground layer. Thus, $\{x_c^i, y_c^i, H_i, W_i\}$ can be obtained by computing the bounding box of the non-transparent, or opaque, region from the alpha channel of $\mathbf{I}_\text{fg}^{i}$.

\vspace{1mm}
\noindent\textbf{Transparency Encoding.}
Our method integrates the transparency in alpha channel $\mathbf{I}_{\text{fg},\alpha}^{i}$ directly into the RGB channels $\mathbf{I}_{\text{fg},\text{RGB}}^{i}$. Specifically, we compute $\hat{\mathbf{I}}_\text{fg}^{i}=(0.5\mathbf{I}_{\text{fg},\alpha}^{i} + 0.5) \times \mathbf{I}_{\text{fg},\text{RGB}}^{i}$, converting the transparent-background image $\mathbf{I}_\text{fg}^{i}$ into a gray-background image $\hat{\mathbf{I}}_\text{fg}^{i}$. All channel values are normalized to range between $-1$ to $1$. Empirically, we found that this gray background sufficient to ensure accurate transparency decoding in subsequent stages.

\vspace{1mm}
\noindent\textbf{Multi-Layer Transparency Encoder.}
In the encoder part of the Multi-layer Transparency Encoder (\Cref{fig:framework_a}), the merged reference image $\mathbf{I}_{\text{mg}}$, the background layer $\mathbf{I}_{\text{bg}}$, and all the padded gray-background image layers $\{\hat{\mathbf{I}}_\text{fg}^{i}\}_{i=1}^{K}$ are all concatenated along the batch dimension, and then fed into the VAE encoder $\mathcal{E}_{\text{VAE}}$. This encoder~\cite{flux} downsamples the spatial dimension with a factor of 8 while obtaining a 16-channel feature dimension.
The extracted latent representations of the merged reference image $\mathbf{I}_{\text{mg}}$ and the background image $\mathbf{I}_{\text{bg}}$ are flattened into sequence of tokens:
\begin{align}
\hspace{-0.5em}
    \mathbf{z}_{\text{mg}}=\mathsf{Flatten}(\mathcal{E}_{\text{VAE}}(\mathbf{I}_\text{mg})),
    \mathbf{z}_{\text{bg}}=\mathsf{Flatten}(\mathcal{E}_{\text{VAE}}(\mathbf{I}_\text{bg})).
\end{align}
\noindent
The pre-processed foreground image layers are first subjected to a ceiling-aligned tight crop and then flattened into latent tokens with different lengths:
\begin{align}
	\mathbf{z}_{\text{fg}}^{i} = \mathsf{Flatten}(\mathsf{Crop}(\mathcal{E}_{\text{VAE}}(\hat{\mathbf{I}}_\text{fg}^{i}), \mathbf{L}_i)), \quad i=1,\cdots,K,
\end{align}
where $\mathbf{L}_i$ denotes the foreground area position of layer $\mathbf{I}_\text{fg}^{i}$. The ceiling-aligned tight crop is performed by identifying the tightest bounding box with a height and width divisible by $16$ to adapt to the VAE downsample rate of $8$ and diffusion transformer patch size $2$.
Finally, the compressed multi-layer image latent $\mathbf{z}$ is obtained by concatenating the latent of the merged reference image, the background image, and the transparent foreground layers:
\begin{align}
	\mathbf{z}   &= \mathsf{Concatenate}(\mathbf{z}_{\text{mg}}, \mathbf{z}_{\text{bg}}, \mathbf{z}_{\text{fg}}^{1}, \mathbf{z}_{\text{fg}}^{2}, \cdots, \mathbf{z}_{\text{fg}}^{K}).
\end{align}

\vspace{1mm}
\noindent\textbf{Multi-Layer Transparency Decoder.}
The detailed design of our novel multi-layer transparency decoder is illustrated on the right in \Cref{fig:framework_a}, which supports the direct decoding of a variable number of transparent layers at varying resolutions from a sequence of concatenated visual tokens in a single forward pass.
We implement the multi-layer transparent image decoder based on a standard ViT architecture.
The mathematical formulations are shown as follows:

\begin{align}
\vspace{-2mm}
	&\mathbf{v}   = \text{ViT}(\text{Linear}_{\text{in}}(\mathbf{z})), \\
	&\mathbf{t}   = \mathsf{Reshape}(\text{Linear}_{\text{out}}(\mathbf{v}), \mathbf{L}),
\end{align}
where $\text{ViT}(\cdot)$ represents the ViT model, $\text{Linear}_{\text{in}}(\cdot)$ denotes a linear projection that transforms the channel dimension of the latent representation, \ie 16, to the hidden dimension size of ViT, especially 768, $\mathbf{v}$ represents the output representation of the ViT, $\text{Linear}_{\text{out}}(\cdot)$ denotes a linear projection that transforms the output dimension from 768 to 256, where each token can be reshaped to form an RGBA patch of size $8\times8\times4$. Another key modification in our design is the replacement of the original absolute position embedding with 3D RoPE, which is explained in the following discussion.
We simply apply $\mathcal{L}_1$ loss to optimize the parameters of the multi-layer transparency decoder while freezing the parameters of the multi-layer transparency encoder.

The advantages of our multi-layer transparency decoder are twofold, including improved efficiency and enhanced transparency predictions compared to the previous single-layer transparent decoder~\cite{zhang2024transparent}. We present the qualitative comparison results in the experimental section.

\subsection{Anonymous Region Transformer} \label{sec:method:mmdit}
The Anonymous Region Transformer (ART) generates the visual tokens of a global reference image, a background image and all foreground layers simultaneously. The purpose of generating reference images is twofold: to better leverage the original capabilities of the existing text-to-image generation model and to ensure overall visual harmonization by preventing conflicts and inconsistency across layers.
Generating all layers simultaneously also avoids the need for inpainting algorithms to complete missing parts of the occluded layers. We choose the latest multimodal diffusion transformer (MMDiT), \eg, FLUX.1[dev]~\cite{flux}, to build our variable multi-layer image generation model, ART.

MMDiT is an improved variant of DiT framework~\cite{esser2024scaling} that uses two different sets of model weights to process text tokens and image tokens separately. 
The original MMDiT model, which only supports single image generation from a global prompt.
We transform it into a multi-layer generation model by modifying the input visual tokens to encode the anonymous region layout information with a novel 3D RoPE design.
We present the overall framework of ART in Figure~\ref{fig:framework} (b).
The input consists of an anonymous region layout $\mathbf{L}$ and a global prompt $\mathbf{T}$. The noisy input is computed by adding Gaussian noise to a sequence of clean multi-layer latents $\mathbf{z}$ that encodes the reference image, background image, and all the transparent layers. We extract $\mathbf{z}$ with our multi-layer transparency encoder.

\vspace{1mm}
\noindent\textbf{Layout Conditional Multi-Layer 3D RoPE.}
Rotary Position Embedding (RoPE)~\cite{su2024roformer} is a specific type of position embedding that applies a rotation operation to key and query in self-attention layers as channel-wise multiplications. The advantage of RoPE is that it allows the model to handle sequences of varying lengths, making it more flexible and efficient. The key design of our ART is to use a layout conditional multi-layer 3D RoPE to encode the accurate relative position information for all visual tokens, which is also utilized in the multi-layer transparency decoder.
We first extract the layer-wise 3D indexing for the given noisy latents according to the anonymous region layout, \ie $\mathbf{p}_n=\{p_n^x, p_n^y, p_n^l\}$ represent the width index, height index, and layer index of the $n$-th latents, respectively. Then, denoted $n$-th query and $m$-th key as $\mathbf{q}_n$ and $\mathbf{k}_m \in \mathbb{R}^{d_{\text{head}}}$, respectively, we split both query and key into 3 parts along channel dimensions, \ie $\mathbf{q}_n=\{\mathbf{q}_n^x, \mathbf{q}_n^y, \mathbf{q}_n^l\}$ and $\mathbf{k}_m=\{\mathbf{k}_m^x, \mathbf{k}_m^y, \mathbf{k}_m^l\}$. Thus, the $(n, m)$ component of the attention matrix is calculated as:
\begin{equation}
	\mathbf{A}_{(n, m)} = \sum_{c\in \{x, y, l\}}{\text{Re}[\mathbf{q}_n^c{(\mathbf{k}_m^c)}^* e^{i(p_n^c-p_m^c)\theta}]},
\end{equation}
where $\text{Re}[\cdot]$ is the real part of a complex number and $(\mathbf{k}_m^c)^*$ represents the conjugate complex number of $\mathbf{k}_m^c$. $\theta \in \mathbb{R}$ is a preset non-zero constant. The detailed implementation can be found in the supplementary material.

\subsection{Anonymous Region Layout Planner} \label{sec:method:planner}

We propose an anonymous region layout planner, which predicts a set of bounding boxes based on the text input. This planner is implemented by fine-tuning an LLM model on our layout dataset, specifically using the pre-trained LLaMa-3.1-8B~\cite{dubey2024llama}. An example of prompts as input and the corresponding predicted layouts is given below. Unlike conventional layout definitions~\cite{kong2022blt,jiang2023layoutformer++,jia2023cole,inoue2024opencole} that specify both position and content, our anonymous region layout planner avoids assigning specific semantic labels to regions. In addition, it refrains from asking users to provide explicit layout details by users, offering greater flexibility.

\begin{contentagnosticlayout}
	\textbf{Input}: The image is a vibrant Ramadan-themed ad featuring a rich blue background with Islamic art-inspired designs and three lit golden lanterns. The white text in the center announces a ``special offer Ramadan big sale", with a subtitle that states ``Discount up to $30\%$ off''.
	\textbf{Output}:  [\{
		``layer": 0,
		``x": 512,
		``y": 512,
		``width": 1024,
		``height": 1024
	\},
	\{
		``layer": 1,
		``x": 744,
		``y": 496,
		``width": 496,
		``height": 256
	\},
	\{
		``layer": 2,
		``x": 856,
		``y": 704,
		``width": 240,
		``height": 96
	\},
	\{
		``layer": 3,
		``x": 792,
		``y": 640,
		``width": 368,
		``height": 64
	\},
	\{
		``layer": 4,
		``x": 840,
		``y": 336,
		``width": 272,
		``height": 64
	\}]
\end{contentagnosticlayout}

\subsection{Multi-Layer Transparent Design Dataset}
We have collected a private, high-quality, multi-layered transparent design (MLTD) dataset that consists of approximately 1 million instances considering their high-quality alpha channels and coherent spatial arrangements. Each instance comprises multiple transparent layers with variable resolutions. The resolutions of the merged images range from $1024\times1024$ to $1500\times1500$.
The average number of layers is $11$, and $99.9\%$ of designs have fewer than $50$ layers. The average number of visual tokens is $11.38$K, which is significantly smaller than $20 \times 32 \times 32 = 20.48$K. This indicates that the area of most foregrounds is relatively small.

\begin{table}[!t]
\begin{minipage}[t]{1\linewidth}
\centering
\tablestyle{2pt}{1.1}
\resizebox{1.0\linewidth}{!}
{
\begin{tabular}{l|c|c|c|c}
\multirow{1}{*}{Dataset} & \multirow{1}{*}{\# Samples} & \multirow{1}{*}{\# Layers} & \multirow{1}{*}{Source Data} & \multirow{1}{*}{Alpha Quality} \\
\shline
MAGICK~\cite{burgert2024magick} & $\sim150$ K & $1$ & generated & good \\
Multi-layer Dataset~\cite{zhang2024transparent} & $\sim1$ M & $2$ & commercial, generated & good  \\
LAION-L\textsuperscript{2}I~\cite{zhang2023text2layer} & $\sim57$ M & $2$ & LAION & normal \\
MuLAn~\cite{tudosiu2024mulan} & $\sim44$ K & $2\sim 6$ & COCO, LAION & poor  \\
MLCID~\cite{huang2024layerdiff} & $\sim2$ M & [$2$,$3$,$4$] & LAION & poor  \\
Crello~\cite{yamaguchi2021canvasvae} & $\sim20$ K & $2\sim50$ & Graphic design website & normal  \\ \hline
MLTD (ours) & $\sim1$ M & $2\sim50$ & Graphic design website & good  \\
\end{tabular}
}
\vspace{-3mm}
\caption{
\small{Comparison with existing multi-layer datasets.}}
\label{tab:dataset_compare}
\end{minipage}
\vspace{-5mm}
\end{table}

\vspace{1mm}
\noindent\textbf{Comparison with Existing Multi-Layer Data}
Table~\ref{tab:dataset_compare} provides a comparison between previously existing multi-layer datasets and our proposed Multi-Layer-Design dataset. Our MLTD dataset is the first large-scale dataset that includes a wide range of transparent layers with high-quality alpha channels. We also verified in the experimental section that our method can achieve sufficiently good results with only $8$K high-quality data, making our method easy to replicate.
\begin{figure*}[t]
\centering
\begin{tabular}{@{}c@{\hspace{2pt}}c@{\hspace{2pt}}c@{\hspace{2pt}}c@{}}
\includegraphics[width=0.223\linewidth]{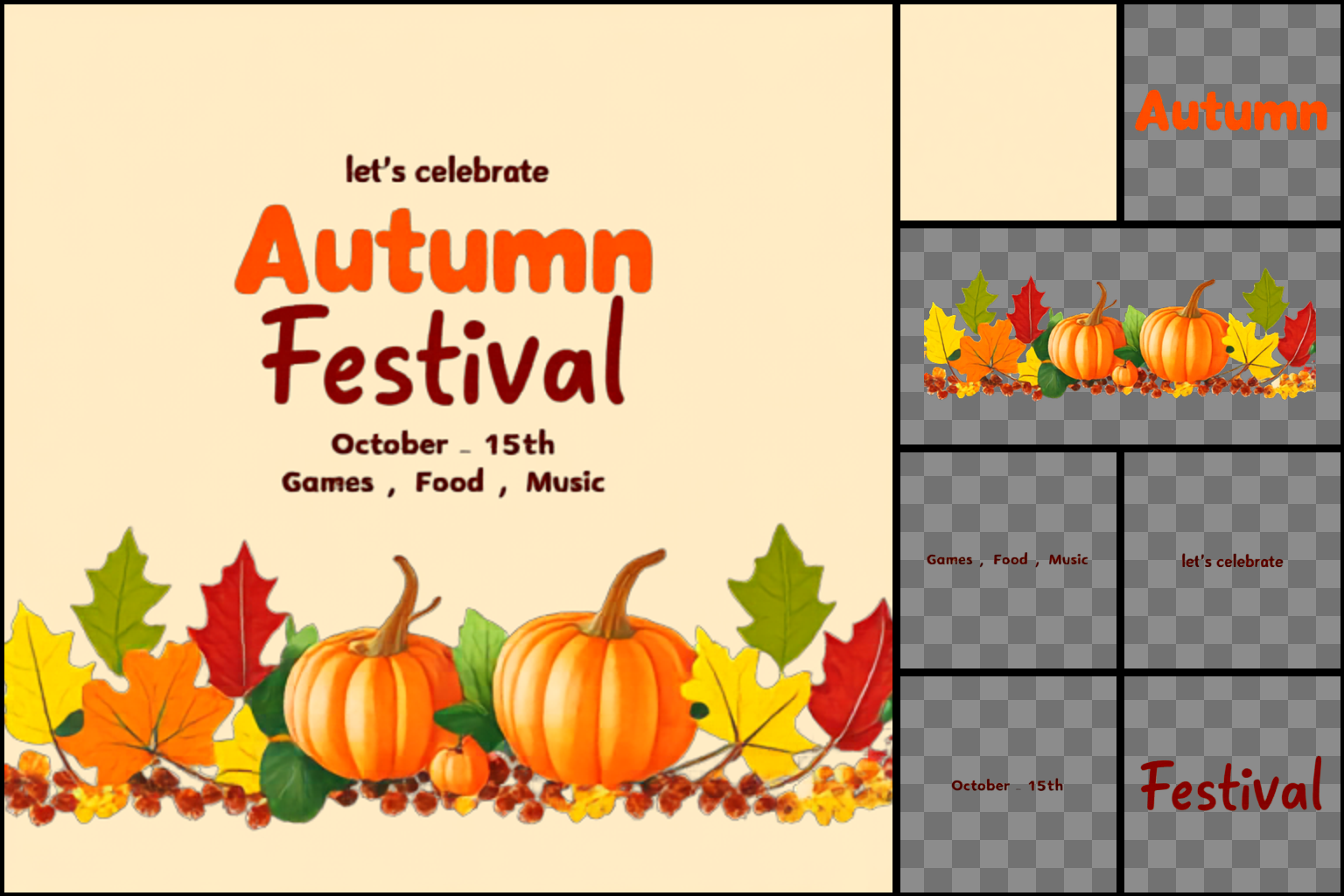} &
\includegraphics[width=0.223\linewidth]{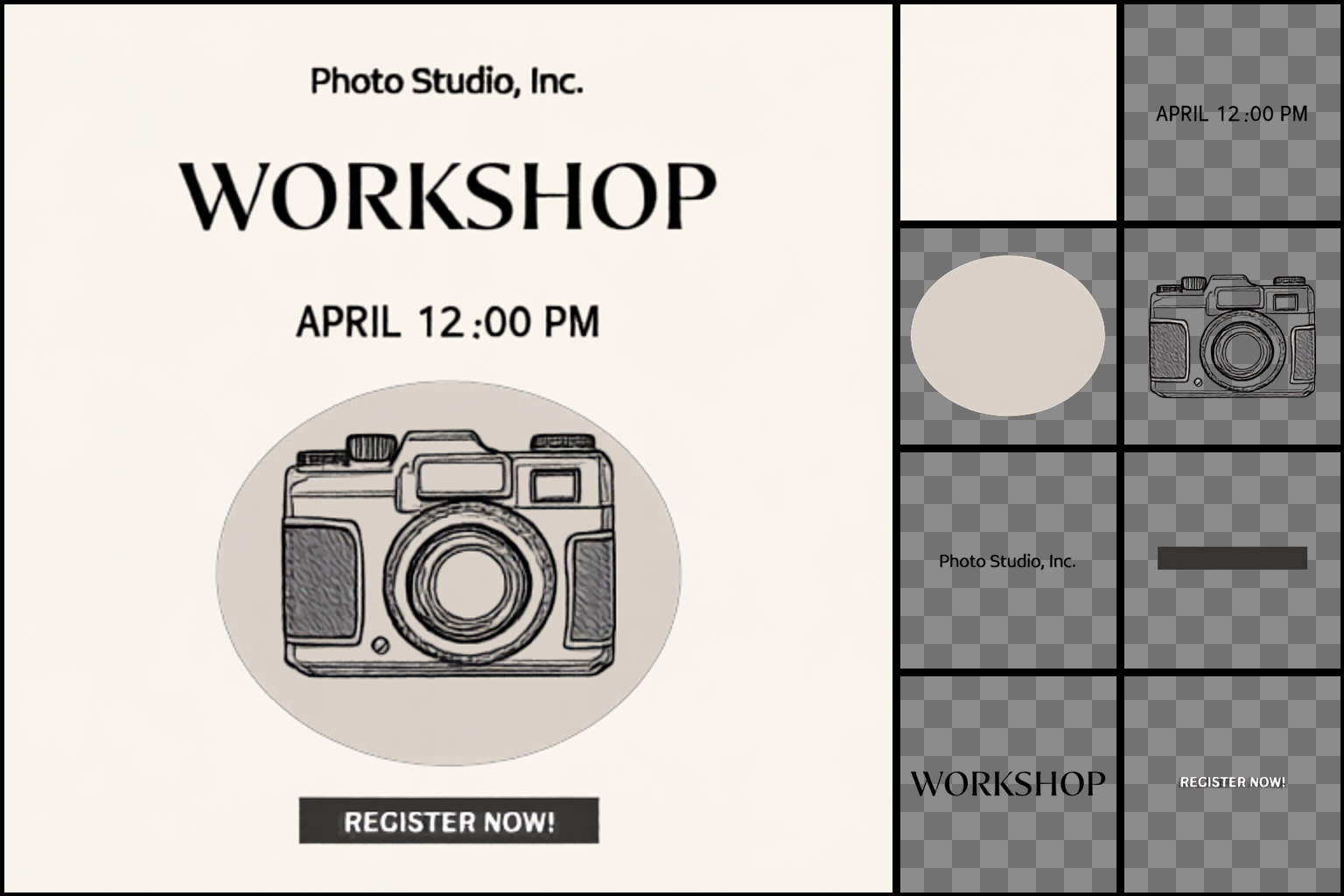} &
\includegraphics[width=0.26\linewidth]{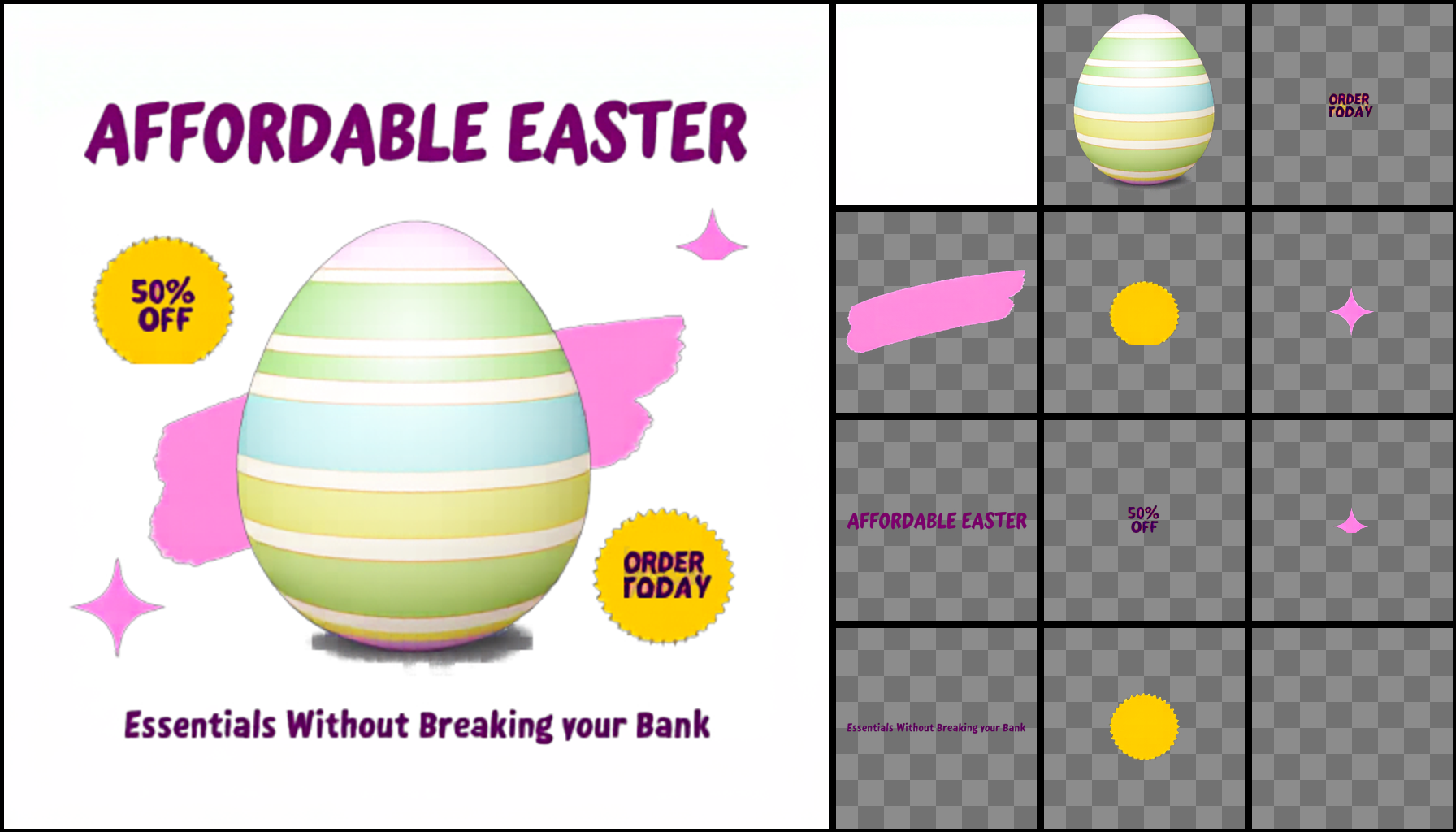} &
\includegraphics[width=0.26\linewidth]{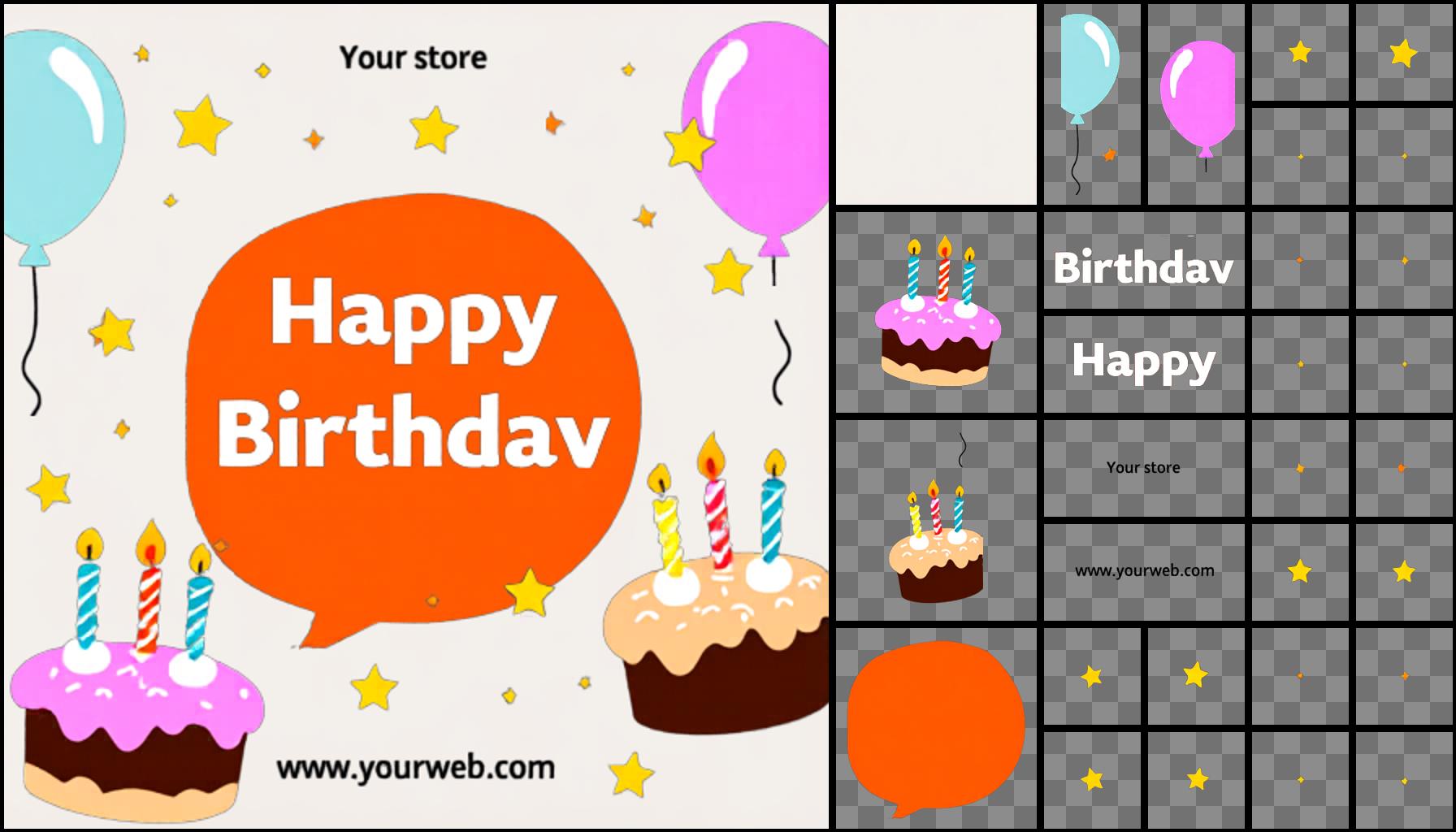} \\
\includegraphics[width=0.223\linewidth]{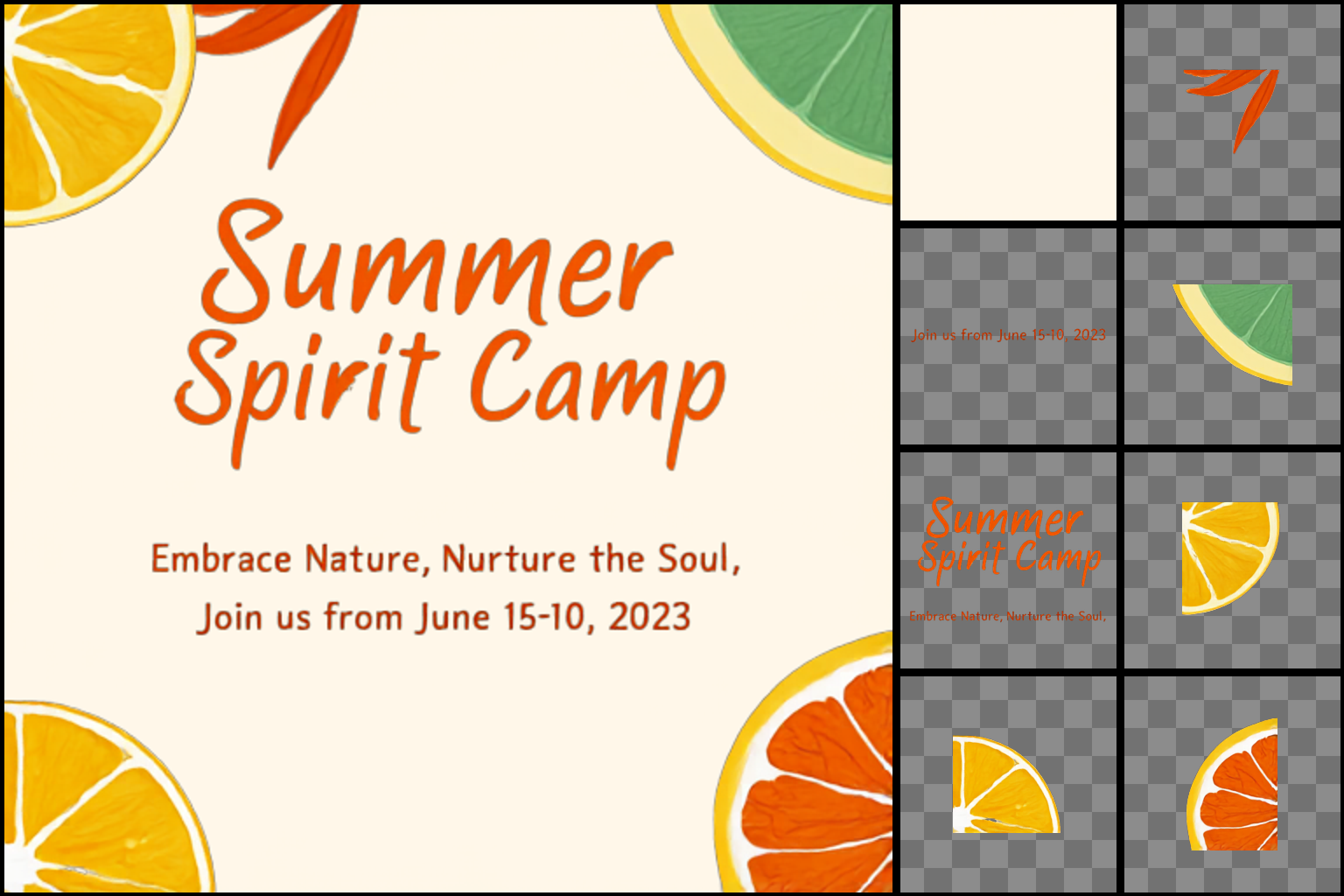} &
\includegraphics[width=0.223\linewidth]{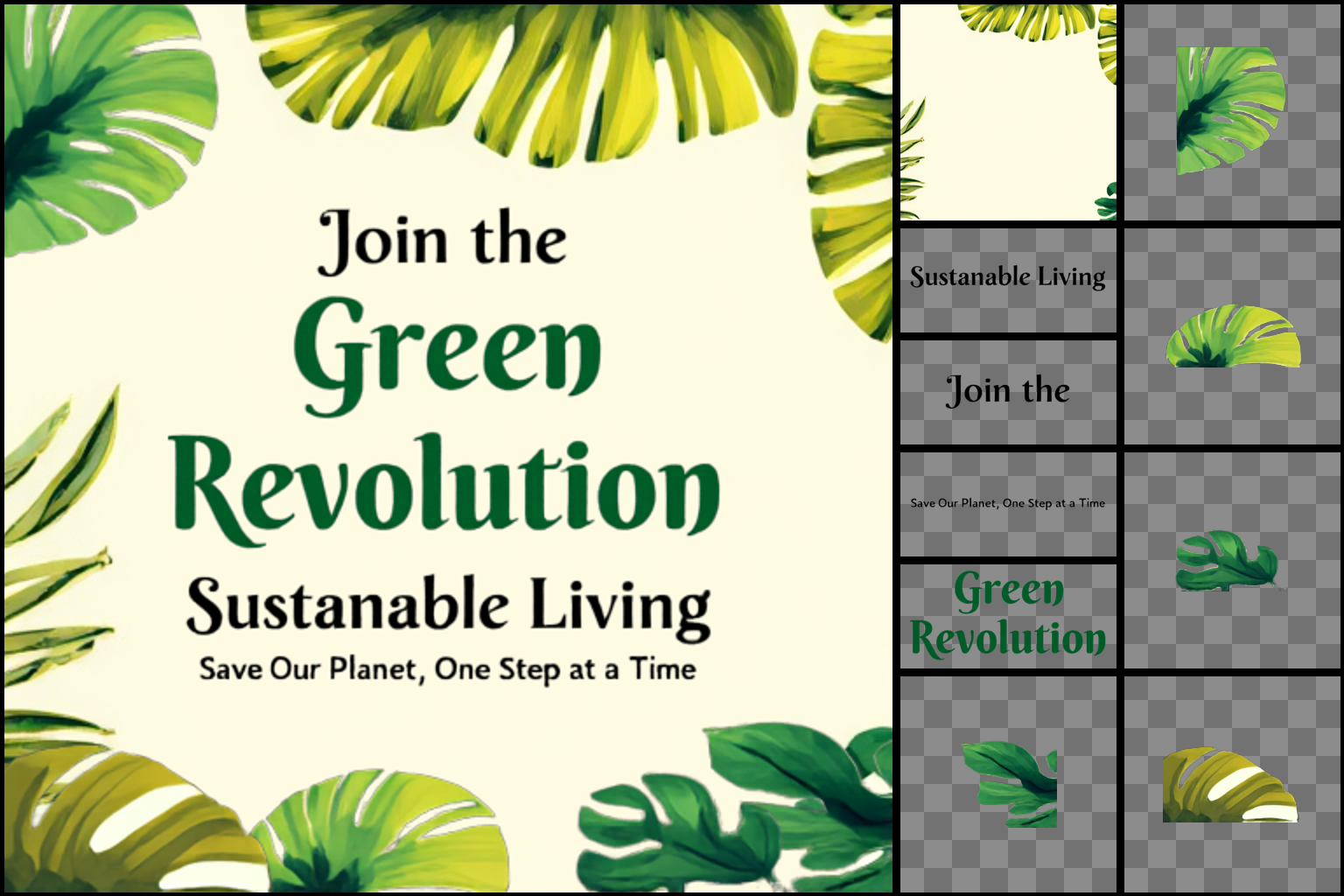} &
\includegraphics[width=0.26\linewidth]{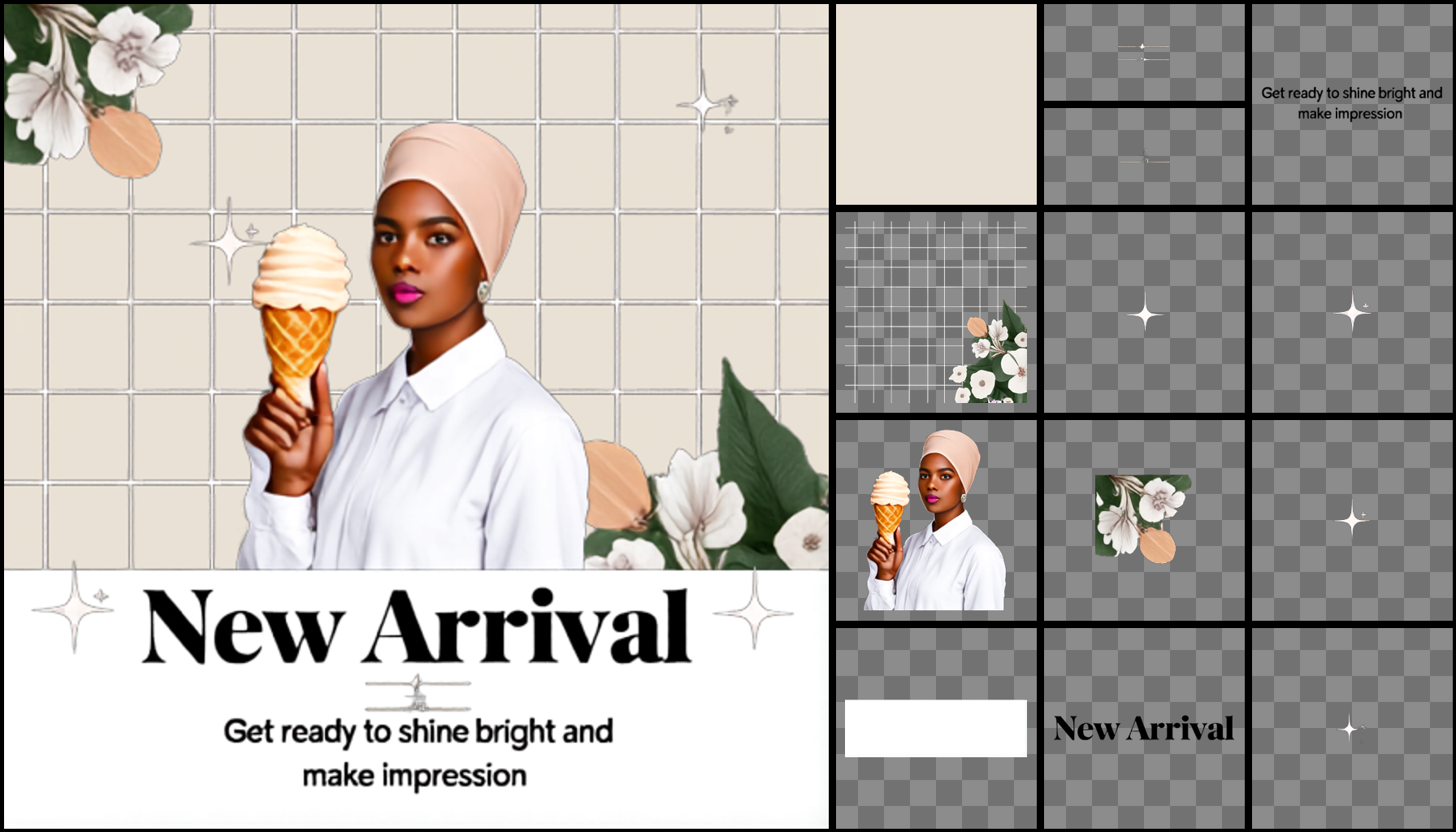} &
\includegraphics[width=0.26\linewidth]{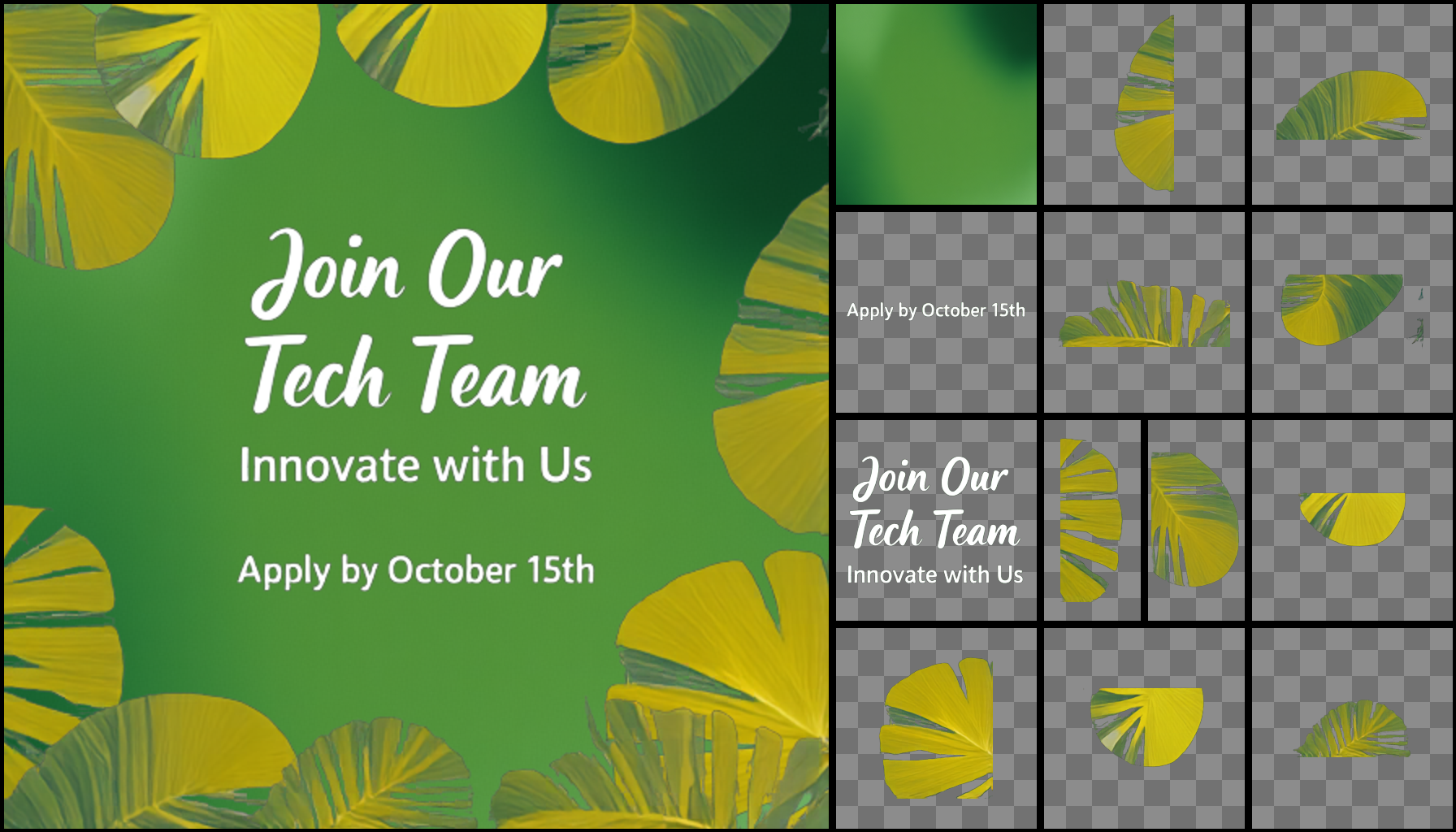}
\end{tabular}
\vspace{-12pt}
\caption{\footnotesize{
\textbf{Variable multi-layer transparent images generated with ART}. The number of transparent layers from top left to bottom right are 7, 8, 11, 30, 8, 10, 12, and 13.
}}
\label{fig:only_ours}
\vspace{-3mm}
\end{figure*}

\begin{figure}[t]
    \centering
    \hspace{-10pt}
    \begin{subfigure}[b]{0.47\linewidth}
        \centering
        \begin{tabular}{@{}c@{\hspace{1pt}}c@{}}
            \includegraphics[height=1.5cm]{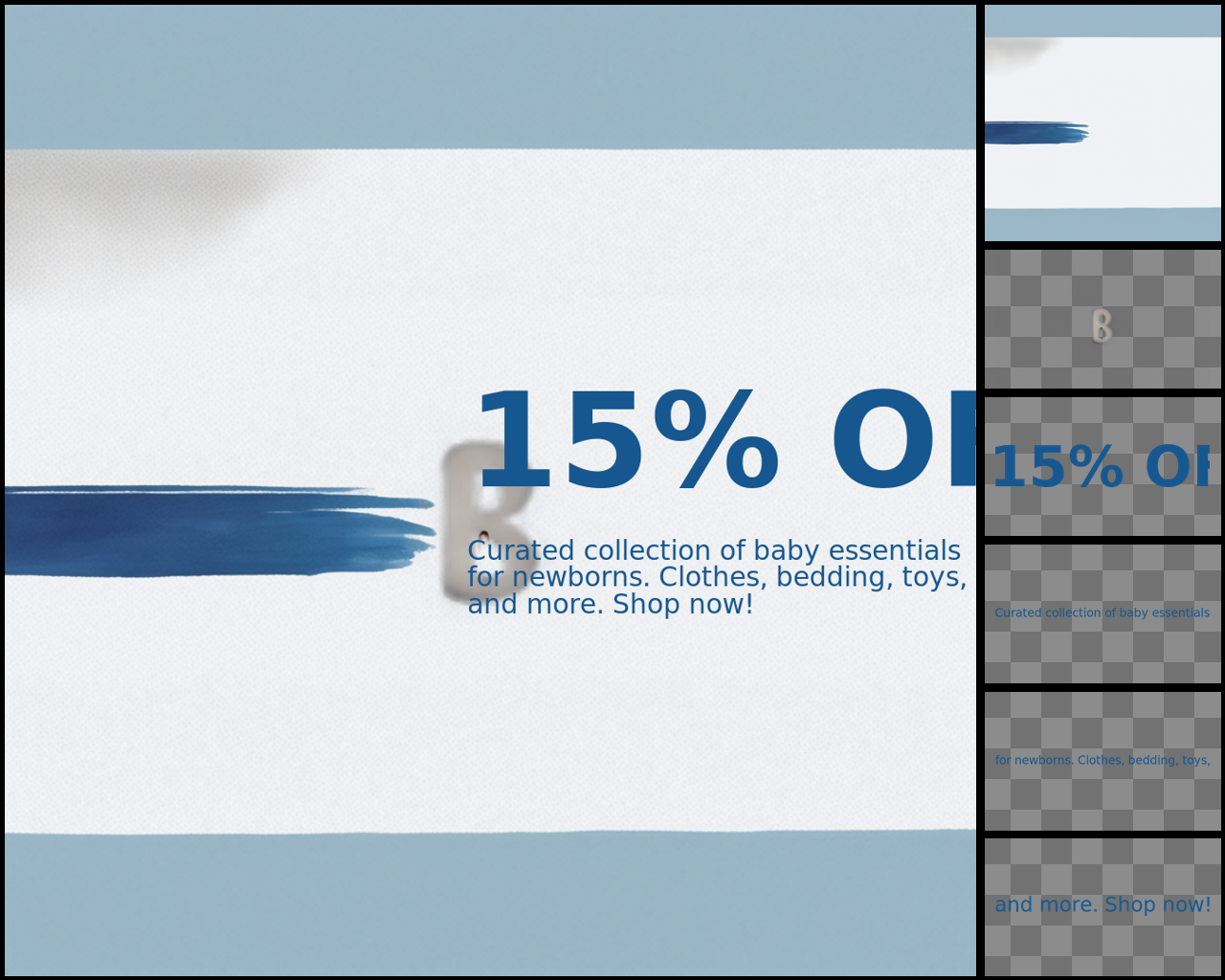} & 
            \includegraphics[height=1.5cm]{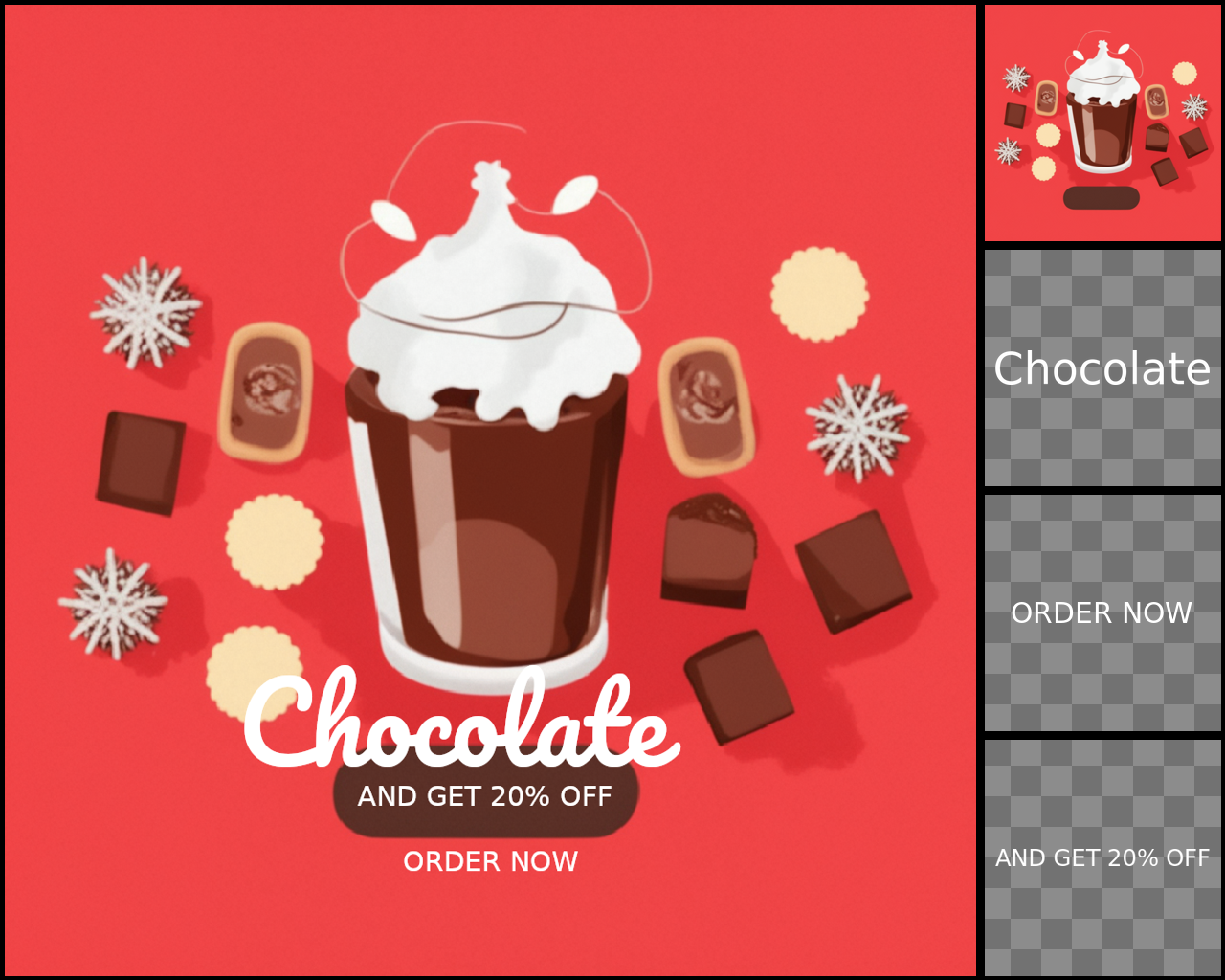} \\ [-3pt]
            \multicolumn{2}{c}{\footnotesize{$\uparrow$~COLE~\cite{jia2023cole} \vs ART~$\downarrow$ }} \\[0pt]
            \includegraphics[height=1.5cm]{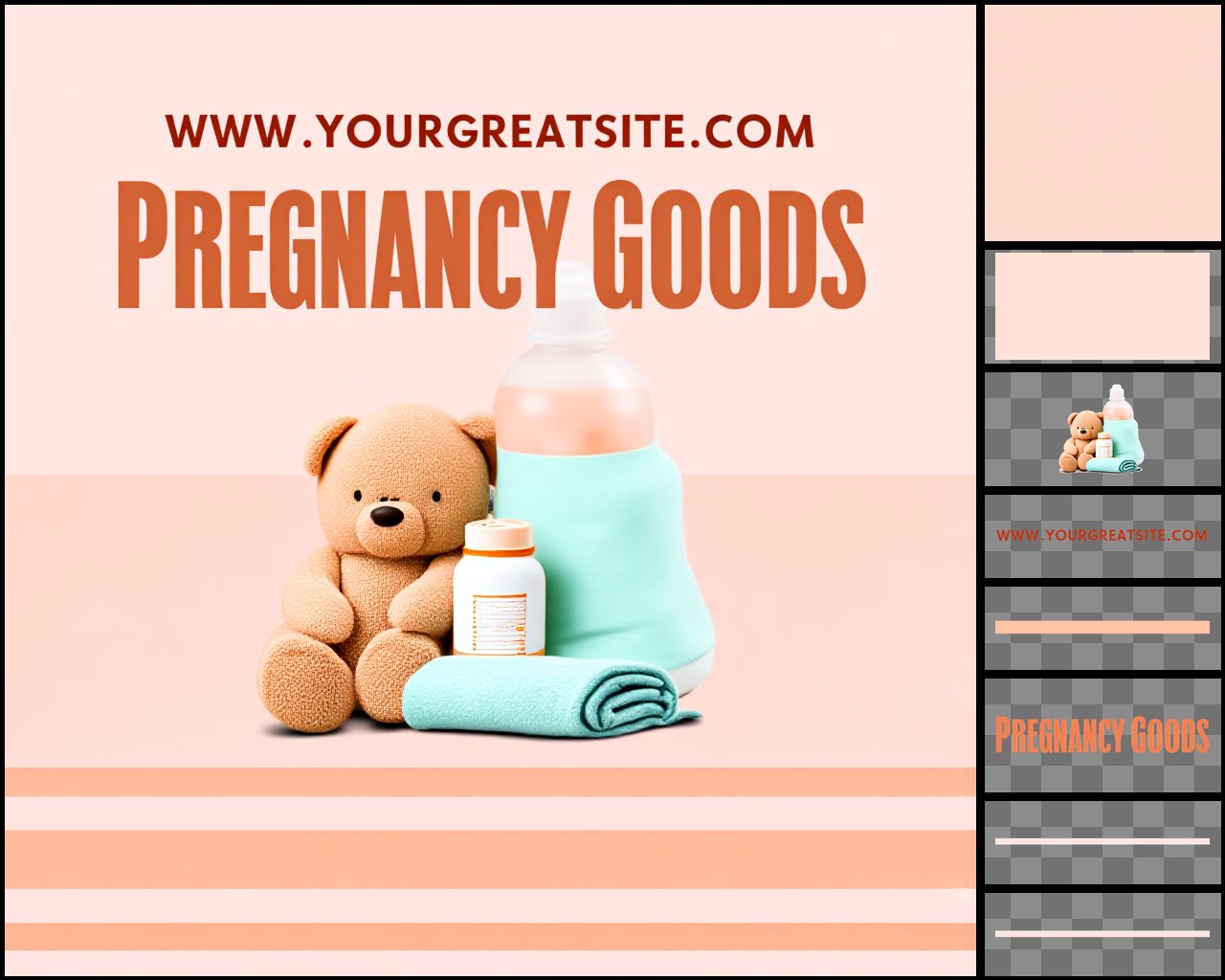} & 
            \includegraphics[height=1.5cm]{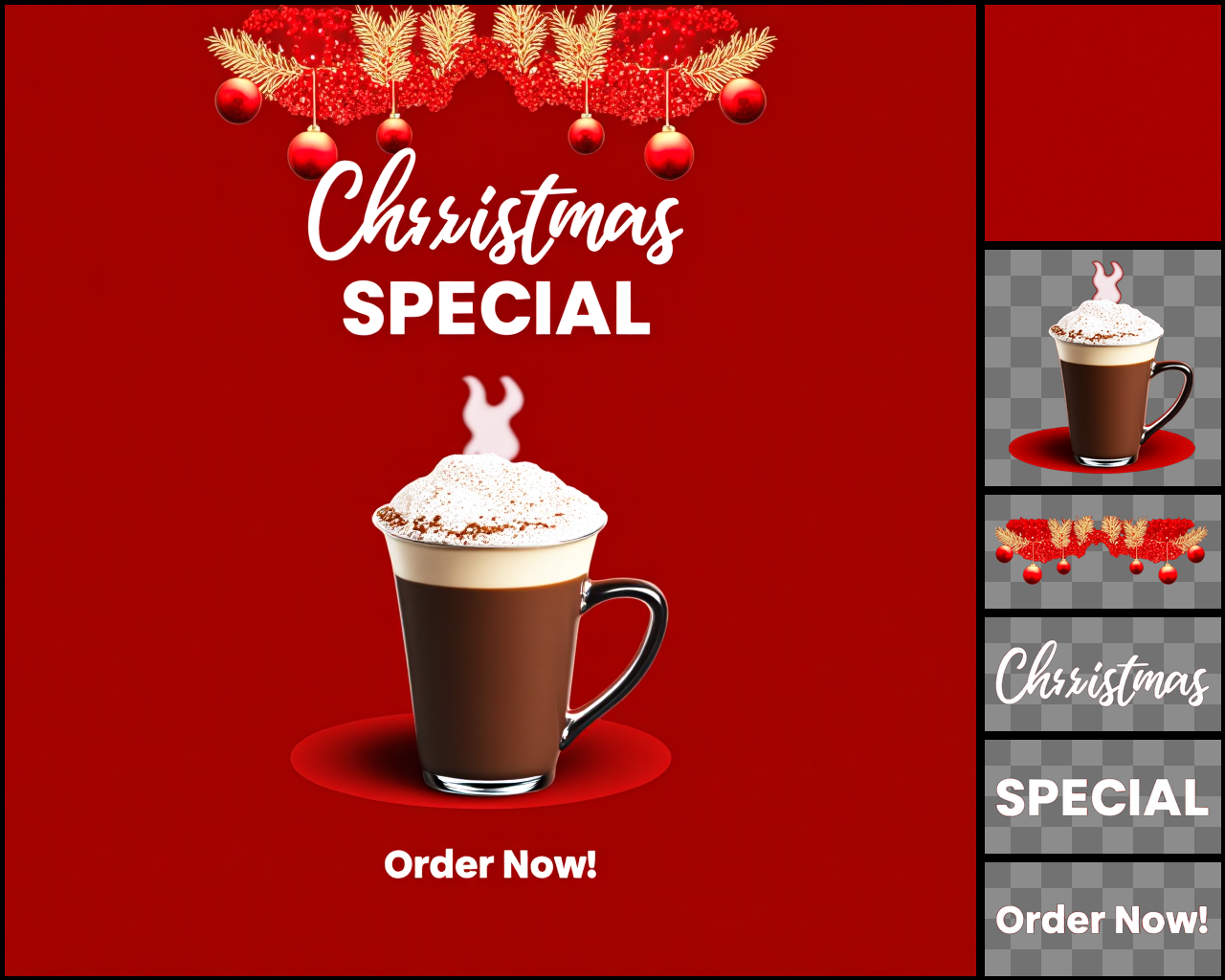} \\
        \end{tabular}
    \end{subfigure}
    \hspace{-2pt}
    \begin{subfigure}[b]{0.48\linewidth}
        \centering
        \begin{tabular}{@{}c@{\hspace{1pt}}c@{}}
            \includegraphics[height=1.5cm]{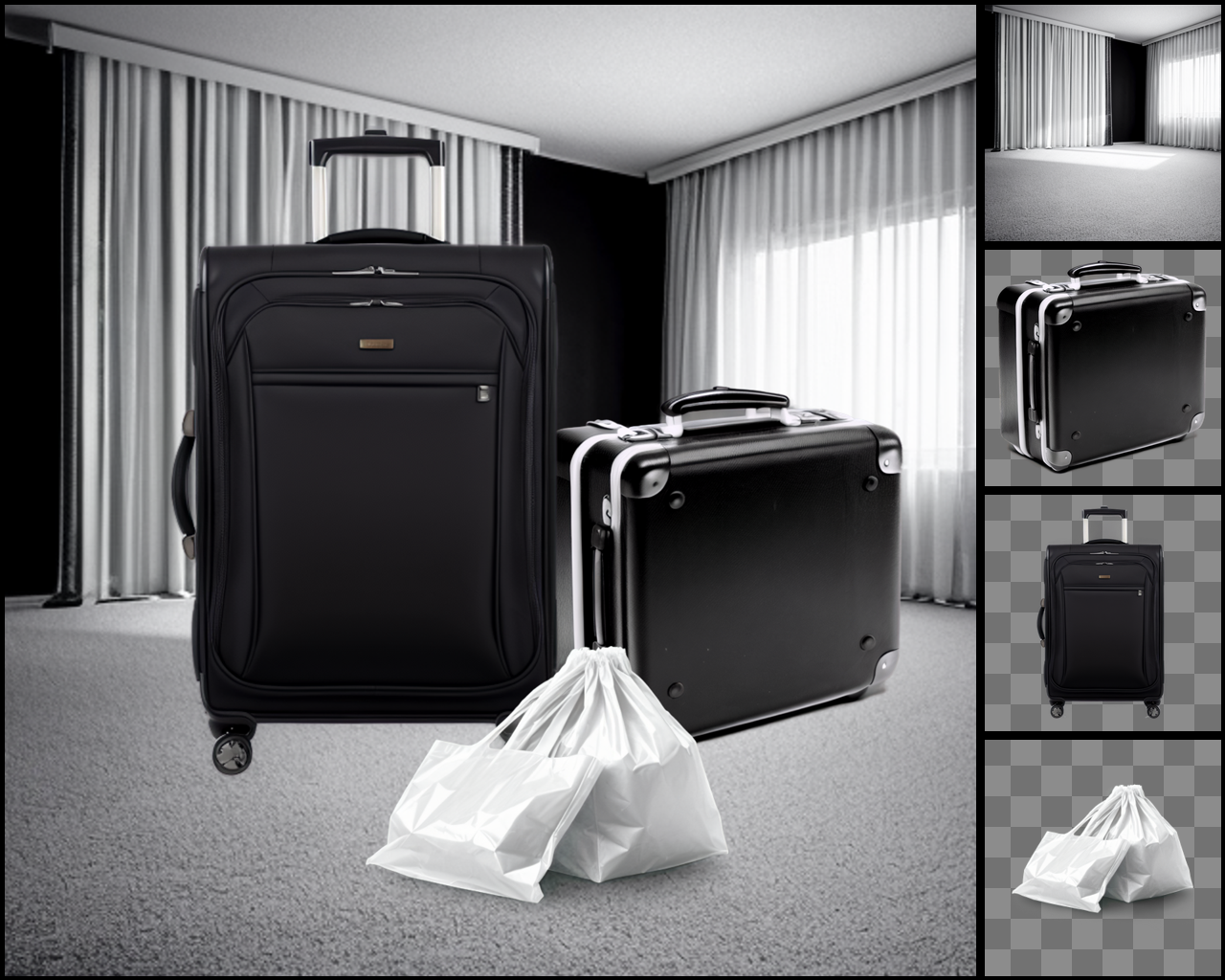} & 
            \includegraphics[height=1.5cm]{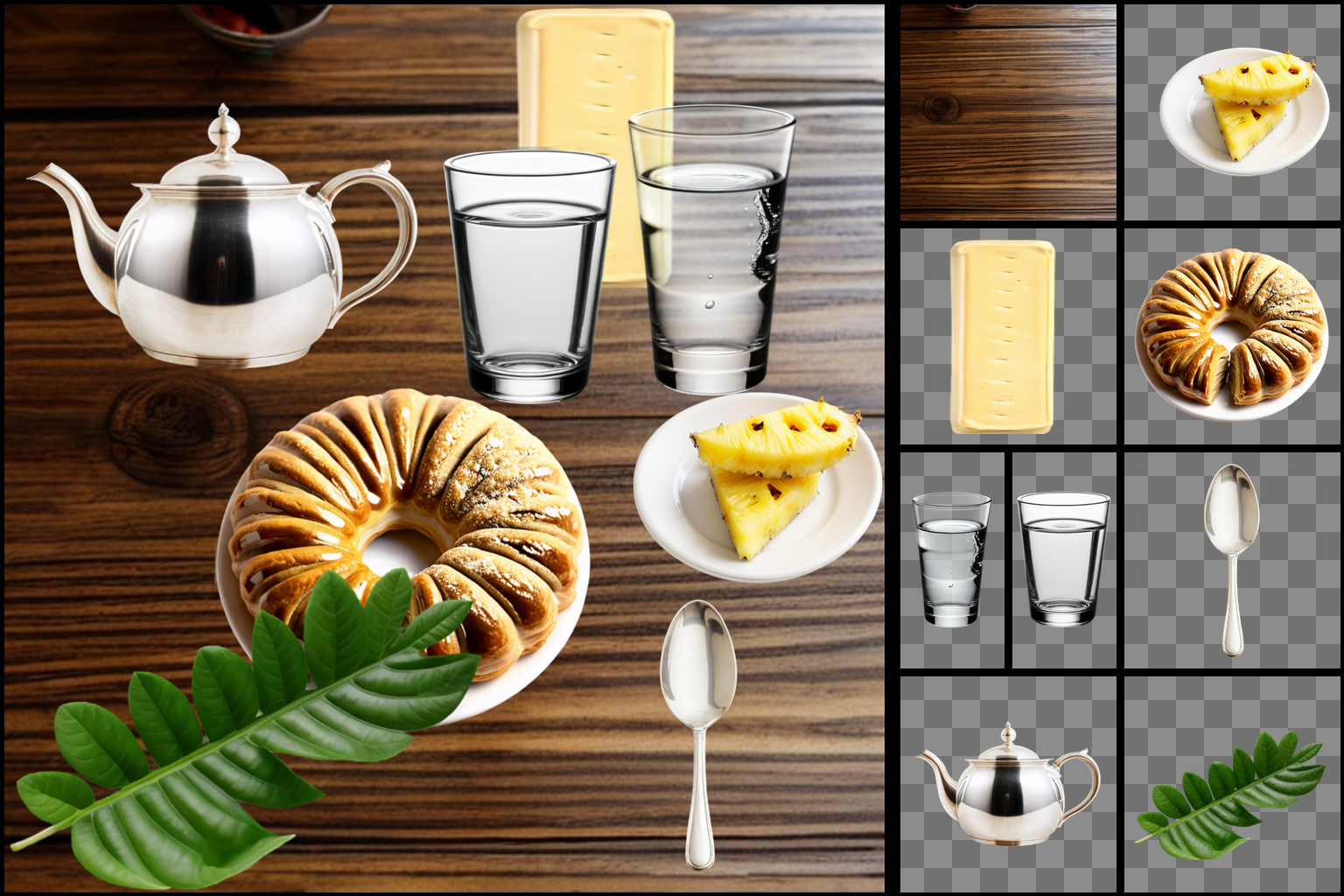} \\ [-3pt]
            \multicolumn{2}{c}{\footnotesize{$\uparrow$~LayerDiffuse~\cite{zhang2024transparent} \vs ART~$\downarrow$ }} \\[0pt]
            \includegraphics[height=1.5cm]{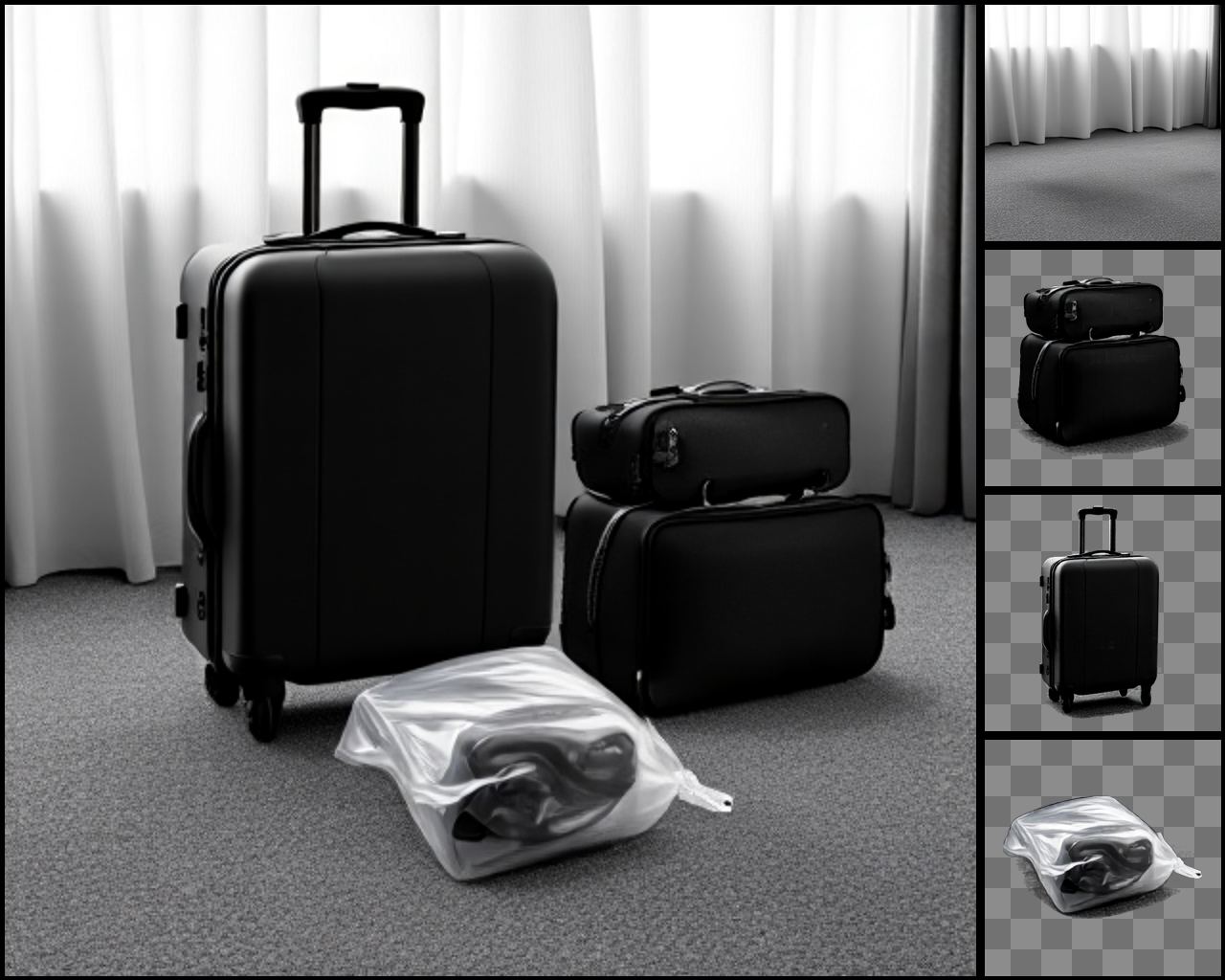} & 
            \includegraphics[height=1.5cm]{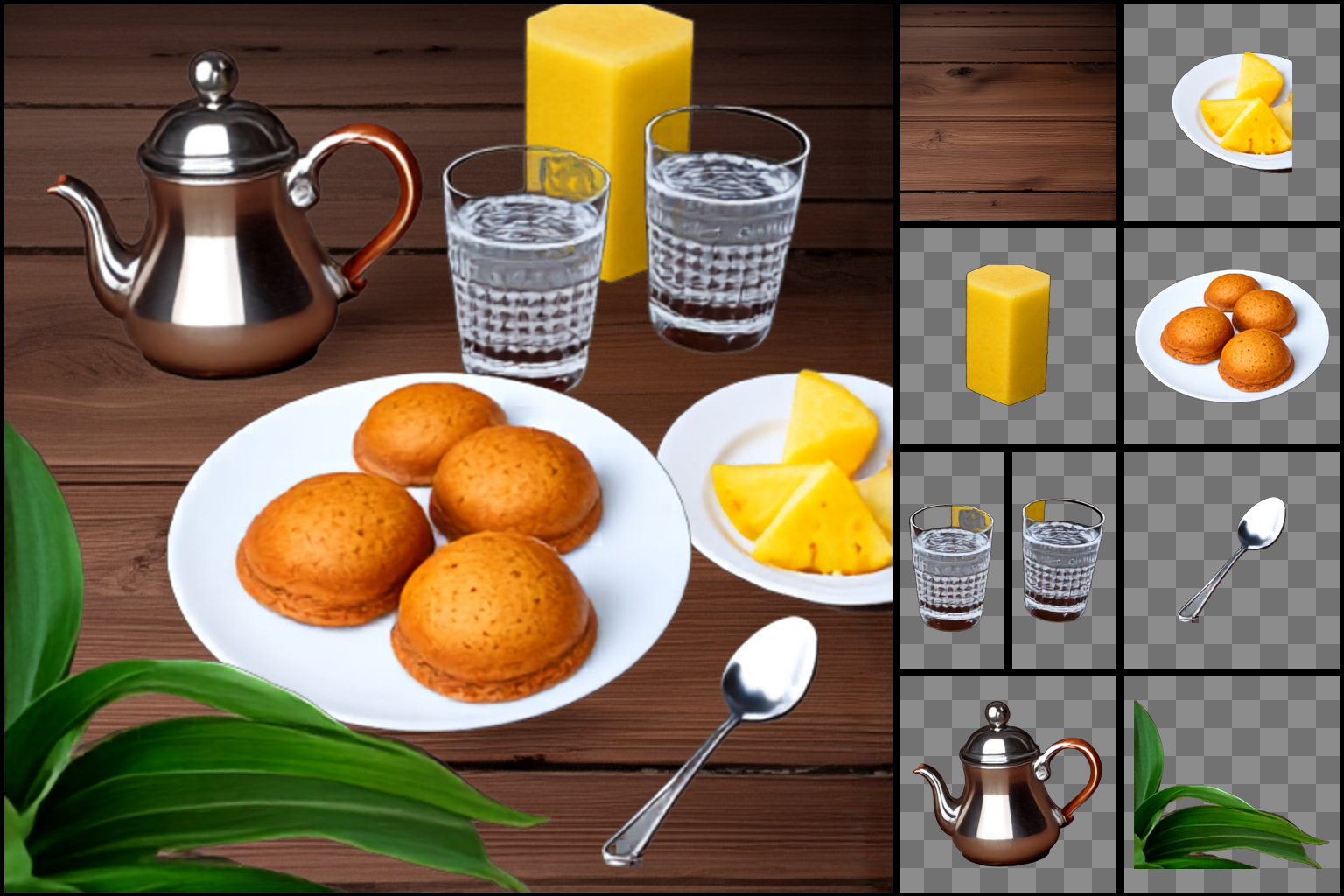} \\
        \end{tabular}
    \end{subfigure}
    \vspace{-12pt}
    \caption{\footnotesize{
    \textbf{ART v.s. COLE or LayerDiffuse}: Given the same global prompt, we display the generated multiple transparent layers to the right of their merged entire image separately. The overall aesthetics and layout of our merged image are superior.
    }}
    \label{fig:colr}
    \vspace{-5mm}
\end{figure}

\section{Experiment}

\vspace{1mm}
\noindent\textbf{Implementation details}. We conduct all the experiments using the latest FLUX.1[dev] model~\cite{flux}. For ablation studies, we train the MMDiT with LoRA for 30,000 iterations, with a global batch size of 8 and a learning rate of 1.0 using the Prodigy optimizer~\cite{mishchenko2023prodigy}. The LoRA rank is set at 64, and the image resolution is at 512$\times$512. To ensure fair comparisons during system-level experiments, we increased the number of iterations to 90,000 and the image resolution to 1024$\times$1024.
For the multi-layer transparency decoder, we selected the ViT-Base configuration~\cite{dosovitskiy2020image}. This configuration includes 12 layers, a hidden dimension size of 768, an MLP dimension size of 3072, and 12 attention heads, resulting in a total of 86 million parameters.

\vspace{1mm}
\noindent\textbf{Training set \& validation set}.
We choose $800$K multi-layer graphic design images as the training set and a set of $5$K graphic design samples to form the validation set, referred to as \designbenchmark. Additionally, we also create a set of photorealistic multi-layer image prompts chosen from the COCO dataset~\cite{lin2014microsoft}, forming \photobenchmark, to evaluate the model's performance on multi-layer real image generation.

\vspace{1mm}
\noindent\textbf{Evaluation metric}.
For the ablation studies, we report standard metrics, including FID~\cite{dowson1982frechet}, PSNR, and SSIM. 
To assess the quality of the Anonymous Region Transfomer, the FID is computed by comparing the predicted merged images to the ground truth merged images, denoted as FID${\scriptstyle\text{merged}}$. The PSNR and SSIM are calculated by comparing the merged image with the predicted reference composed image. 
To assess the quality of the multi-layer transparency image autoencoder, we report the PSNR for the RGB channels and the alpha channel separately, \ie, PSNR{$^{\textrm{layer}}_{\textrm{RGB}}$} and PSNR{$^{\textrm{layer}}_{\textrm{alpha}}$}, by comparing the reconstructed transparent layers with the ground-truth transparent layers. For the system-level comparisons, we conduct a user study to assess the quality of the composed image and transparent layers from four aspects: visual aesthetics, prompt adherence, typography, and inter-layer harmonization.

For fair comparisons, we use the layout predicted by our anonymous region layout planner model for the system-level comparison experiments, while the human-provided anonymous layout is used by default for all ablation studies, unless otherwise specified.

\subsection{System-level Comparisons}\label{exp:system_level}

We report the system-level comparisons with state-of-the-art methods in photorealistic image space (evaluated on \photobenchmark) and graphic design space (evaluated on \designbenchmark). 

\vspace{1mm}
\noindent\textbf{Comparison to LayerDiffuse}. We first compare our approach with the latest multi-layer generation method, LayerDiffuse~\cite{zhang2024transparent}, in the multi-layer real image generation benchmark, \ie, \photobenchmark. We conduct a user study involving 30 participants with diverse backgrounds in AI, graphic design, art, and marketing, each evaluating 50 pairs of multi-layer transparent images generated by our ART and LayerDiffuse across three aspects: harmonization, aesthetics, and prompt following.
The results of the user study are illustrated in \Cref{fig:user_study}. We observe that our approach significantly outperforms LayerDiffuse across all three dimensions.

\vspace{1mm}
\noindent\textbf{Comparison to COLE}. We further conduct a user study to compare our approach with the multi-layer graphic design image generation method COLE~\cite{jia2023cole}. We also ask the same 30 participants to evaluate the organization of the elements (layout), the visual appeal (aesthetics), the correctness of the text (typography), and the coherence and quality of each layer (harmonization), with each user evaluating 50 image pairs. The results in \Cref{fig:user_study} reveal that our approach achieves significantly better multi-layer image generation results in various aspects, except for typography, as the text in COLE is rendered with typography render.

\noindent\textbf{More results}. 
We present more multi-layer image generation in \Cref{fig:only_ours} (up to 30 layers), as well as qualitative comparison results with COLE and LayerDiffuse in \Cref{fig:colr}.

\subsection{Ablation Study and Analysis}
\label{sec:ablation}

\vspace{1mm}
\noindent\textbf{Anonymous Region Layout is Sufficient}. We first address the key question of whether region-specific prompts are necessary for multi-layer image generation tasks by comparing the conventional semantic layout and our anonymous region layout. For the semantic layout, we generate region-specific prompts for each layer using the LLaVA 1.6 model~\cite{liu2023improvedllava} and ensure that the visual tokens of each region mainly attend to their respective regional prompts. To ensure a fair comparison, we utilize the ground-truth layout provided by our \designbenchmark while maintaining consistency across all other experimental settings, differing only in the use of region-specific prompts.

Table~\ref{tab:ablation:anonymous_layout_vs_semantic_layout} provides a detailed comparison of the results. We find that the FID$\scriptstyle \text{merged}$ scores for both methods are comparable, while the PSNR score for the anonymous region layout is significantly higher. This suggests superior layer coherence and global harmonization in our approach. Additionally, we employ GPT-4o to evaluate both methods in terms of global harmonization, arriving at the consistent conclusion that our approach yields better layer coherence. One potential reason for the lower coherence in the semantic layout approach is the conflict between local region-specific prompts and global visual tokens. We provide a deeper analysis of these conflicts in the supplementary material.

In addition, we present a statistical analysis comparing the inferred label assignments for the anonymous regions generated by our ART model with the human-annotated region-wise prompts. Our findings reveal that over 80\% of the inferred labels align with the human annotations, suggesting that the generative models have acquired prior knowledge akin to Schema Theory. Additional details can be found in the supplementary material.

\begin{table}[t]
\begin{minipage}[t]{1\linewidth}  
\centering 
\tablestyle{6pt}{1.1}
\resizebox{0.99\linewidth}{!}
{
\begin{tabular}{l|c|cc|c}  
    method & FID$\scriptstyle \text{merged}$  & PSNR & SSIM & Harmonization Score (GPT-4o) \\
    \shline
    Semantic Layout & \underline{17.51} & 17.71 & 0.8443   & 3.67 \\
    Anonymous Region Layout  & 17.79 & \underline{22.90} & \underline{0.9021} & \underline{3.92} \\
\end{tabular}
}
\vspace{-2mm}
\caption{\footnotesize{Anonymous Region Layout \vs Semantic Layout.}}
\label{tab:ablation:anonymous_layout_vs_semantic_layout}
\vspace{2mm}
\end{minipage}
\begin{minipage}[t]{1\linewidth}  
\centering 
\tablestyle{12pt}{1.1}
\resizebox{0.99\linewidth}{!}
{
\begin{tabular}{c|c|cc|c}  
composed image pred. & FID$\scriptstyle \text{merged}$ & PSNR & SSIM & Inference speed (s)\\
\shline
\xmark & 20.44 & - & - & \underline{19.20} \\
\cmark & \underline{17.79}  & 22.90 & 0.9021 & 26.62 \\
\end{tabular}
}
\vspace{-2mm}
\caption{
\footnotesize{Composed image prediction improves the image quality.} }
\label{tab:ablation:compose_img_predict}
\vspace{2mm}
\end{minipage}
\begin{minipage}[t]{1\linewidth}  
\centering 
\tablestyle{15pt}{1.1}
\resizebox{0.99\linewidth}{!}
{
\begin{tabular}{l|c|cc}  
attention type & FID$\scriptstyle \text{merged}$ & PSNR & SSIM  \\
    \shline
    Full Att.  & 41.35 & 16.87 & 0.7738   \\
    Spatial Att.  + Temporal Att. & 167.99  & 16.92 & 0.7985  \\
    Regional Full Att.  & \underline{17.79} & \underline{22.90} & \underline{0.9021} \\
\end{tabular}
}
\vspace{-2mm}
\caption{\footnotesize{Full Att. \vs Spatial Att. + Temporal Att. \vs \textit{Regional Full Att.}}}
\label{tab:ablation:tight_crop}
\vspace{-3mm}
\end{minipage}
\end{table}

\begin{figure}[!t]
\begin{minipage}[!t]{1\linewidth}
\centering
\includegraphics[width=1\textwidth]{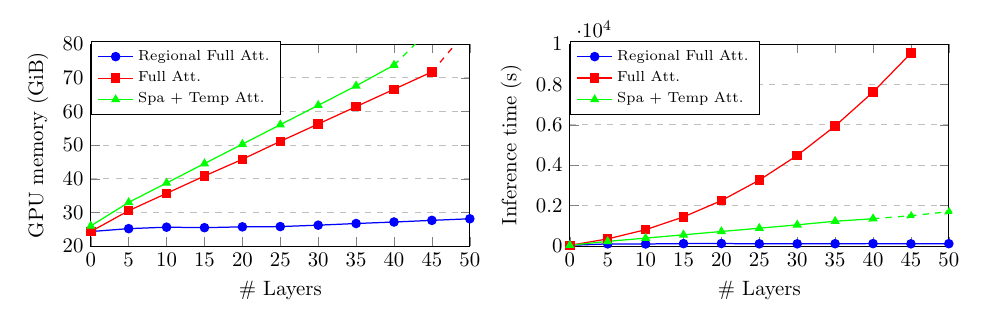}
\end{minipage}
\vspace{-2mm}
\caption{\footnotesize{
Illustrating the efficiency comparison of three different attention mechanism design: our Regional Full Attention (marked as Regional Full Att.), Full Attention (marked as Full Att.) and Spatial + Temporal Attention (marked as Spa + Temp Att.). The GPU memory consumption and inference time are evaluated and averaged over 100 samples at a resolution of 1024$\times$1024, for each given number of layers. Some data points are represented with dashed lines or are not shown due to the OOM issue.
}}
\label{fig:efficiency}
\vspace{-2mm}
\end{figure}

\vspace{1mm}
\noindent\textbf{The Benefits of Predicting the Reference Composed Image}. We introduced an additional prediction of the reference composed image for two main reasons. First, it improves coherence across multiple image layers by facilitating bidirectional information propagation between the composed image and each transparent layer. Second, it provides a mechanism to evaluate the quality and consistency of the predicted transparent layers by calculating the PSNR and SSIM scores between the reference image and the layer-merged image on the validation set. As shown in Table~\ref{tab:ablation:compose_img_predict}, our results demonstrate the significance of predicting the composed image as a reference, leading to enhanced image quality as indicated by the FID$\scriptstyle \text{merged}$ score, albeit with a slight increase in inference time.

\begin{table}[t]
\begin{minipage}[t]{1\linewidth}  
\centering 
\tablestyle{20pt}{1.1}
\resizebox{0.99\linewidth}{!}
{
\begin{tabular}{l|c|cc}  
PE method & FID$\scriptstyle \text{merged}$ & PSNR & SSIM \\
\shline
2D-RoPE &  124.3 & 11.99 & 0.4265  \\
2D-RoPE + LayerPE & 20.66  & \underline{23.23} & \underline{0.9101}  \\
3D-RoPE & \underline{17.79} & 22.90 & 0.9021 \\
\end{tabular}
}
\vspace{-2mm}
\caption{
\footnotesize{Different position embedding scheme.}}
\vspace{2mm}
\label{tab:ablation:pe_choice}
\end{minipage}
\begin{minipage}[t]{1\linewidth}  
\centering 
\tablestyle{25pt}{1.1}
\resizebox{0.99\linewidth}{!}
{
\begin{tabular}{l|c|cc}
\# samples & FID$\scriptstyle \text{merged}$ & PSNR & SSIM \\
\shline
80 & 30.38 & \underline{23.18} & 0.8893  \\
800 & 18.89  & 20.45 & 0.8609 \\
8k & 18.06  & 22.43 & 0.8882  \\
80k & 18.04  & 23.13& \underline{0.9081}  \\
800k & \underline{17.79} & 22.90 & 0.9021 \\
\end{tabular}
}
\vspace{-2mm}
\caption{
\footnotesize{Increasing the dataset scale improves performance.}}
\label{tab:ablation:data_scale}
\vspace{3mm}
\end{minipage}
\begin{minipage}[c]{1\linewidth}
\centering
\tablestyle{18pt}{1.1}
\resizebox{\linewidth}{!}{
\begin{tabular}{l|cccc}
\#layer numbers & 3$\sim$8 & 9$\sim$12 & 13$\sim$15 & 16$\sim$51 \\
\shline
FID$\scriptstyle \text{merged}$ & 49.83 & 47.19 & 44.56 & 42.40 \\
\end{tabular}}
\vspace{-2mm}
\caption{
\footnotesize{Effect of different layer numbers.}}
\vspace{3mm}
\label{tab:ablation:layer_number}
\end{minipage}
\hfill
\begin{minipage}[c]{1\linewidth}
\centering
\tablestyle{17pt}{1.1}
\resizebox{1\linewidth}{!}{
\begin{tabular}{l|cccc}
\#text tokens & 23$\sim$58 & 59$\sim$83 & 84$\sim$159 & 160$\sim$272 \\
\shline
FID$\scriptstyle \text{merged}$ & 27.70 & 26.98 & 28.66 & 28.22 \\
\end{tabular}}
\vspace{-2mm}
\caption{
\footnotesize{Effect of different caption length.}}
\label{tab:ablation:text_number}
\end{minipage}
\vspace{-5mm}
\end{table}

\vspace{1mm}
\noindent\textbf{Regional Full Attention v.s. Full Attention v.s. Spatial + Temporal Attention}. A key design element of our approach is the ceiling-aligned tight crop for each transparent layer, which removes most transparent pixels and compels the diffusion model to focus on the smallest rectangle encapsulating the non-transparent foreground regions. We refer to this as the Regional Full Attention scheme. This design is crucial for improving efficiency and explicitly constrains layer predictions to align with the positions specified by the anonymous region layout. We also evaluate two additional baselines: the Full Attention scheme, which does not apply regional cropping, and the Spatial Attention + Temporal Attention scheme, which introduces temporal attention to facilitate interactions across different layers, similar to architectural designs in video generation~\cite{blattmann2023align,guo2023animatediff}. Detailed comparison results are presented in Table~\ref{tab:ablation:tight_crop}, where our method demonstrates superior FID$\scriptstyle \text{merged}$ scores. The primary factor behind our improved performance is the use of the anonymous region layout.

Moreover, Figure~\ref{fig:efficiency} shows that our method maintains nearly constant computational costs when processing between 10 and 50 layers, whereas the Full Attention scheme, lacking regional cropping, exhibits quadratic growth in memory and inference costs.

\begin{table}[t]
	\begin{minipage}[t]{1\linewidth}  
		\centering 
		\tablestyle{2pt}{1.1}
		\resizebox{0.99\linewidth}{!}
		{
			\begin{tabular}{l|c|cc|c}  
				Method  &  FID$\scriptstyle \text{merged}$ & PSNR & SSIM & Inference speed (s) \\
				\shline
				GPT-4o & 20.72 & 22.80 & 0.9078 & - \\
				LayoutGPT~\cite{feng2024layoutgpt} & 20.92 & \underline{23.18} & \underline{0.9113} & -  \\ \hline
				Semantic Layout Planner & 21.45 & 17.69 & 0.8382 & 19.19 \\
				Semantic Layout Planner\dag & 20.63 & 22.90 & 0.9092 & 19.19 \\
				Anonymous Region Layout Planner & \underline{19.90} & 22.70 & 0.9038 & \underline{5.68} \\
			\end{tabular}
		}
		\vspace{-2mm}
		\caption{
			\footnotesize{Anonymous region layout planner v.s. semantic layout planner and other planner alternatives. \dag\, means that we remove the predicted region-specific prompts and only use the predicted bounding boxes.}}
		\label{tab:ablation:layout_compare}
	\end{minipage}
	\vspace{-2mm}
\end{table}

\begin{table}[t]
\begin{minipage}[t]{1\linewidth}  
\centering 
\tablestyle{9pt}{1.1}
\resizebox{0.99\linewidth}{!}
{
\begin{tabular}{l|ccc|c}  
PE method & PSNR{$^{\textrm{layer}}_{\textrm{rgb}}$} & PSNR{$^{\textrm{layer}}_{\textrm{alpha}}$} & PSNR & FID$\scriptstyle \text{merged}$ \\
\shline
2D-AbsPE & 26.91 & 18.42 & 26.06 & 17.04 \\
2D-AbsPE + LayerPE & 26.98 & 18.76 & 26.11 & 16.24 \\
2D-RoPE & 34.05 & 23.08 & 30.09 & 3.16 \\
2D-RoPE + LayerPE & 34.46 & 23.31 & 30.13 & 3.10 \\
3D-RoPE & \underline{34.89} & \underline{23.85} & \underline{30.48} & \underline{2.84} \\
\end{tabular}
}
\vspace{-2mm}
\caption{
\footnotesize{Position embedding scheme in multi-layer decoder.}}
\label{tab:ablation:pe_multi_layer_decoder}
\vspace{2mm}
\end{minipage}
\begin{minipage}[t]{1\linewidth}  
\centering 
\tablestyle{8pt}{1.1}
\resizebox{0.99\linewidth}{!}
{
\begin{tabular}{cc|ccc|c}  
composed image & bg image & PSNR{$^{\textrm{layer}}_{\textrm{rgb}}$} & PSNR{$^{\textrm{layer}}_{\textrm{alpha}}$} & PSNR & FID$\scriptstyle \text{merged}$ \\
\shline
\xmark & \xmark & 33.25 & 22.82 & 29.35 & 3.76 \\
\cmark & \xmark & 33.25 & 21.95 & 29.39 & 3.53 \\
\xmark & \cmark & 34.37 & 23.39 & 30.20 & 3.06 \\
\cmark & \cmark & \underline{34.89} & \underline{23.85} & \underline{30.48} & \underline{2.84} \\
\end{tabular}
}
\vspace{-2mm}
\caption{
\footnotesize{Condition choice for the multi-layer decoder.}}
\vspace{2mm}
\label{tab:ablation:condition_choice}
\end{minipage}
\begin{minipage}[t]{1\linewidth}  
\centering 
\tablestyle{6pt}{1.1}
\resizebox{0.99\linewidth}{!}
{
\begin{tabular}{lc|ccc|c}  
Method & Multi layer & PSNR{$^{\textrm{layer}}_{\textrm{rgb}}$} & PSNR{$^{\textrm{layer}}_{\textrm{alpha}}$} & PSNR & FID$\scriptstyle \text{merged}$ \\
\shline
LayerDiffuse~\cite{zhang2024transparent} & \xmark & 20.94 & 18.48 & 26.51 & 4.27 \\
Flux-RGBA decoder & \xmark & 30.25 & 20.11 & 27.74 & 5.23 \\
Ours & \cmark & \underline{34.89} & \underline{23.85} & \underline{30.48} & \underline{2.84} \\
\end{tabular}
}
\vspace{-2mm}
\caption{
\footnotesize{Comparison with existing transparency decoder.}}
\label{tab:ablation:sota_decoder_compare}
\end{minipage}
\vspace{-3mm}
\end{table}

\vspace{1mm}
\noindent\textbf{Layer-aware Position Encoding is Critical.} Encoding positional information is essential for the model to distinguish visual tokens from different transparent layers. Our empirical analysis shows that incorporating layer position information is crucial, with the proposed 3D-RoPE scheme outperforming the absolute layer position encoding method. The full comparison results are presented in Table~\ref{tab:ablation:pe_choice}.

\vspace{1mm}
\noindent\textbf{More Multi-layer Data Brings Better Performance.} Table~\ref{tab:ablation:data_scale} reports the detailed experimental results when training with datasets of varying scales. We observe that our approach benefits from a larger dataset scale. One interesting observation is that our ART already achieve strong results with just 8K training samples, demonstrating that our approach is also data efficient.

\vspace{1mm}
\noindent\textbf{Effect of number of transparent layers and the complexity of the scenarios described in the text.}
We study whether our ART performs robustly across various input complexities by partitioning the test set into different groups according to the number of transparent layers and the number of text tokens (which reflects the complexity of the scenarios) and report the quantitative comparison results on these subsets in Table~\ref{tab:ablation:layer_number} and Table~\ref{tab:ablation:text_number}. 
We can see that our ART achieves even better performance with an increasing number of transparent layers and slightly weaker performance when handling longer text tokens. We attribute this to the distributions of these factors in the training set.

\begin{figure}[!t]
\begin{minipage}[!t]{1\linewidth}
    \begin{subfigure}[b]{1\textwidth}
    \centering
\includegraphics[width=\linewidth]{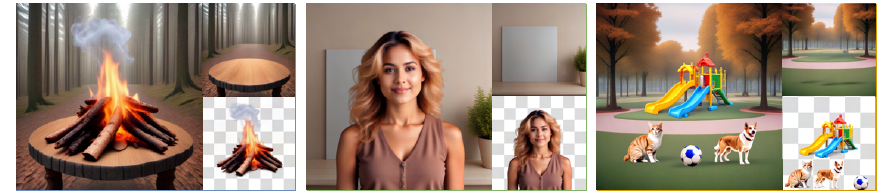}
\vspace{-2mm}
\end{subfigure}
\vspace{-6mm}
\caption{Multi-layer natural image generation results.}
\label{fig:semi_tran}
\vspace{3mm}
\end{minipage}
\begin{minipage}[!t]{1\linewidth}
\begin{minipage}[!t]{0.25\linewidth}
    \begin{subfigure}[b]{1\textwidth}
    \centering
    \includegraphics[width=0.96\textwidth]{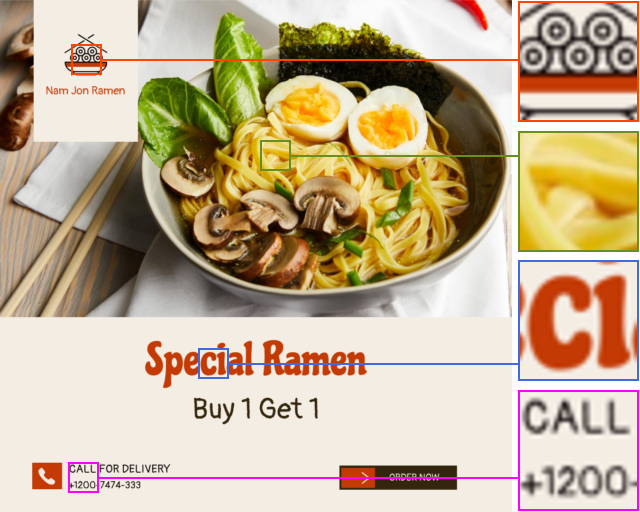}
    \caption{\footnotesize{Ground-truth}}
    \end{subfigure}
\end{minipage}%
\begin{minipage}[!t]{0.25\linewidth}
    \begin{subfigure}[b]{1\textwidth}
    \centering
    \includegraphics[width=0.96\textwidth]{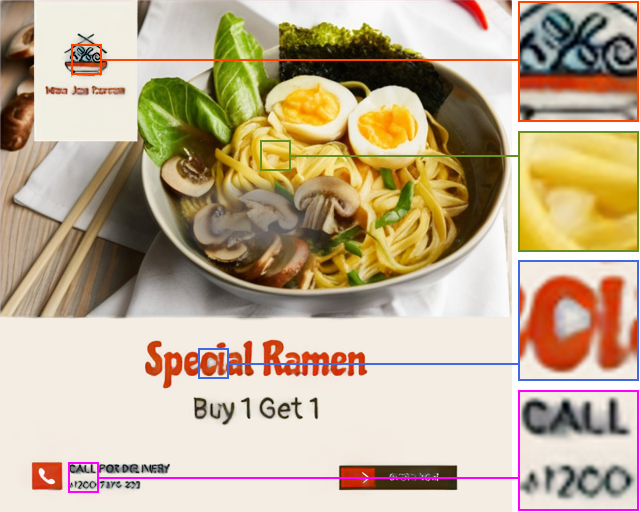}
    \caption{\footnotesize{LayerDiffuse}}
    \end{subfigure}
\end{minipage}%
\begin{minipage}[!t]{0.25\linewidth}
    \begin{subfigure}[b]{1\textwidth}
    \centering
    \includegraphics[width=0.96\textwidth]{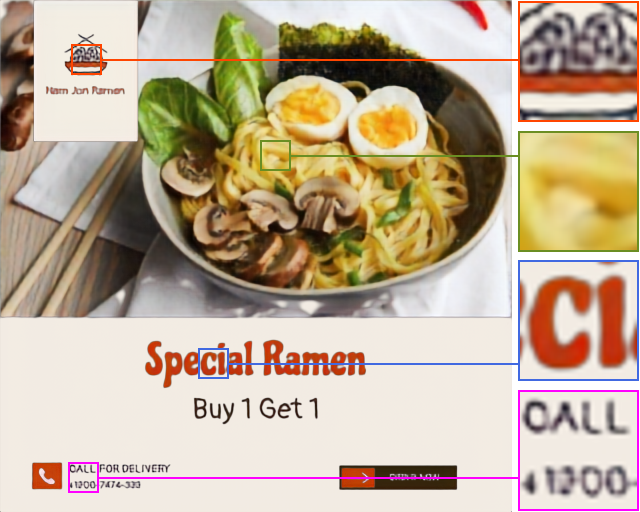}
    \caption{\footnotesize{Flux-RGBA}}
    \end{subfigure}
\end{minipage}%
\begin{minipage}[!t]{0.25\linewidth}
    \begin{subfigure}[b]{1\textwidth}
    \centering
    \includegraphics[width=0.96\textwidth]{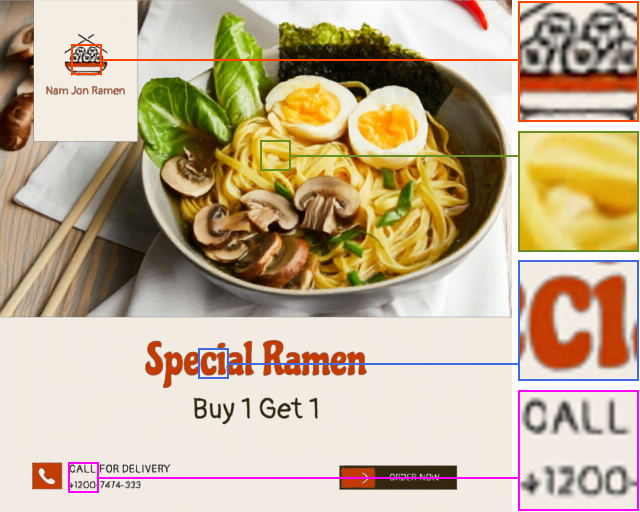}
    \caption{\footnotesize{Ours}}
    \end{subfigure}
\end{minipage}
\vspace{-3mm}
\caption{\footnotesize{Comparison with existing transparency decoder.}}
\label{fig:sota_decoder_compare}
\vspace{2mm}
\end{minipage}
\vspace{-6mm}
\end{figure}

\vspace{1mm}
\noindent\textbf{Multi-layer Natural Image Generation Results.}
Our approach can be directly applied to multi-layer natural image generation without any modifications, given access to a high-quality multi-layer natural image dataset. To this end, we show that our ART achieves strong results even when fine-tuned on only a 20 curated high-quality multi-layer natural images. Figure~\ref{fig:semi_tran} shows some qualitative results and we believe the results can continue to improve with access to more high-quality multi-layer natural images.

\vspace{1mm}
\noindent\textbf{Anonymous Region Layout Planner v.s. Semantic Layout Planner.} We fine-tune both an anonymous layout planner and a semantic layout planner using data sampled from the 800K training dataset and evaluate their performance by integrating them with our ART model. Additionally, we include two strong baselines, GPT-4o and LayoutGPT~\cite{feng2024layoutgpt}, which support transforming the global prompt into a usable layout. Detailed results are presented in Table~\ref{tab:ablation:layout_compare}. Our Anonymous Region Layout Planner not only achieves a better FID$\scriptstyle \text{merged}$ score but also operates more than 3$\times$ faster than the Semantic Layout Planner.
Interestingly, removing the region-specific prompts of the semantic layout planner can enhance overall performance by avoiding conflicts among region-wise prompts, especially regarding layer coherence, as reflected by the higher PSNR scores.

\vspace{1mm}
\noindent\textbf{RoPE is Critical for Multi-layer Decoder Quality.}
Table~\ref{tab:ablation:pe_multi_layer_decoder} summarizes the results of the comparison experiments involving different position embedding schemes for the multi-layer transparency decoder. The original ViT pre-trained on the ImageNet classification task employs absolute position encoding, which is inadequate for capturing positional information across a variable number of transparent layers. We find that simply adding an additional set of layer-wise absolute position embeddings provides minimal improvement; however, replacing the absolute position encoding with the RoPE scheme significantly enhances decoding quality. We observe that the 3D-RoPE scheme achieves the best FID$\scriptstyle \text{merged}$ score, which aligns with our findings regarding the choice of position encoding scheme for the latent features sent into MMDiT. Consequently, we adopt the 3D-RoPE scheme as default.

\vspace{1mm}
\noindent\textbf{Composed Image as Condition.} Although we only need to decode the transparency for all the foreground transparent layers, we empirically find that sending both the merged entire image and the background image as additional conditions, along with applying supervision on them, leads to even better performance, as shown in Table~\ref{tab:ablation:condition_choice}. We hypothesize that the information from the merged and background images is beneficial for the transparency layers to interact more effectively, ensuring a more coherent final composed image with these transparent layers.

\vspace{1mm}
\noindent\textbf{Comparison with Previous Transparency Decoder.}
We compare our multi-layer transparency decoder with the previous transparency decoder and two strong baselines designed for single-layer transparency decoding, as shown in Table~\ref{tab:ablation:sota_decoder_compare}. We utilize the officially released weights of the transparency decoder proposed by LayerDiffuse~\cite{zhang2024transparent}. For the Flux-RGBA decoder, we modify the output projection to support an additional alpha layer prediction and fine-tune the decoder using our dataset. Our design achieves the best FID$\scriptstyle \text{merged}$ score as shown in Table~\ref{tab:ablation:sota_decoder_compare}. The qualitative comparison results presents in Figure~\ref{fig:sota_decoder_compare}.
\vspace{-5pt}
\section{Conclusion}
\label{sec:conclusion}

In this paper, we introduce the Anonymous Region Transformer, a novel approach for generating multi-layer transparent images from an anonymous region layout. Our results and analysis reveal that our anonymous layout is sufficient for the multi-layer transparent image generation task. Our method offers several key advantages over traditional semantic layout methods, including better coherence across layers and more scalable annotation. Furthermore, our method enables the efficient generation of images with numerous distinct transparent layers, reducing computational costs and generalizing to various distinct anonymous region layouts. However, our approach does have certain limitations, including repeated layer generation and combined layer generation. The generalizability of this capability across all potential layouts requires further exploration. Future work should focus on enhancing the model's ability to autonomously identify semantic labels and improving the quality and flexibility of the generated images. Despite these challenges, our approach shows promising potential for graphic design and digital art.

\vspace{1mm}
\noindent\textbf{Future works}
We believe our work lays a solid foundation for the next generation of generative models that can produce a variable number of coherent transparent layers and support flexible image editing through layer compositing. Looking forward, we identify several promising directions for future research:
(i) \emph{Enhancing Visual Aesthetics}:
A key challenge is to improve the visual appealing of the generated transparent layers and ensure that the composite images achieve parity with those produced by state-of-the-art text-to-image models such as FLUX.
(ii) \emph{Anonymous Region Layouts}:
We anticipate that leveraging anonymous region layouts will transform conventional layout-to-image generation tasks. This approach has the potential to eliminate the need for complex regional prompt annotations and to simplify the modeling process by granting models greater control.
(iii) \emph{Human Interaction with ART}:
We also see great promise in integrating user requirements into the multi-layer image generation system. Future work could explore interactive methods for incorporating real-time user feedback, enabling dynamic refinement of generated layers and more personalized editing workflows.

{
\small
\bibliographystyle{ieeenat_fullname}
\bibliography{main}
}

\clearpage

\setcounter{page}{1}
\setcounter{section}{0}
\setcounter{figure}{0}
\setcounter{table}{0}
\setcounter{equation}{0}

\maketitlesupplementary

\vspace{1mm}
\section{Detailed List of Prompts and Anonymous Region Layouts}
\Cref{tab:generation_results_1,tab:generation_results_1_additional,tab:generation_results_2} illustrate the detailed global prompts and anonymous layouts used in Figure 5 and Figure 6 of the main paper, respectively.
In the first two rows of Table~\ref{tab:generation_results_2}, we select the global prompts based on the intentions outlined in the \textsc{Designintention} benchmark for fair comparisons.

Table~\ref{tab:user_study_prompt_1} and Table~\ref{tab:user_study_prompt_2} illustrate the detailed instructions used in our user study on the \textsc{PHOTO-MULTI-LAYER-BENCH} benchmark and \textsc{Design-MULTI-LAYER-BENCH} benchmark, respectively.

\vspace{1mm}
\section{Analyzing the Conflicts within Semantic Layouts}
As mentioned in the main paper, we observe lower coherence in the generated multi-layer images with the semantic layout approach.
First, we present some typical results in Figure~\ref{fig:conflict_case}, marking the inconsistent regions between the predicted global reference image and the merged global image. Second, we visualize the attention maps between the regional visual tokens (as Query) and the combination of the region-caption text tokens and the global visual tokens from the global reference image (as Key and Value).

We observe that the visual tokens of each region primarily attend to the region-wise prompts while relying less on the predicted reference image, resulting in less coherent outputs. The purpose of predicting the global reference image is to ensure coherence across different layers. We infer that the essential reasons behind the conflict between the global reference image and the region-wise prompts stem from the disparity between the region-wise prompts and the global prompts, as \emph{there exists a non-trivial gap between the global prompt and the region prompt associated with the same regional crop.}

\vspace{1mm}
\section{Analyzing the Inferred Label Assignments within Anonymous Region Layouts}
To measure the distance between the inferred label assignments and the human annotations provided by the anonymous region layout, we calculate the averaged layer-wise CLIP scores. These scores reflect whether the generated transparent layers in each anonymous region match the human-annotated ground-truth region-wise prompts by computing the CLIP scores between the regional visual features and the regional prompt text features.

\begin{figure}[!t]
\begin{minipage}[!t]{1\linewidth}
\begin{subfigure}[b]{1\textwidth}
\centering
\includegraphics[width=1\textwidth]{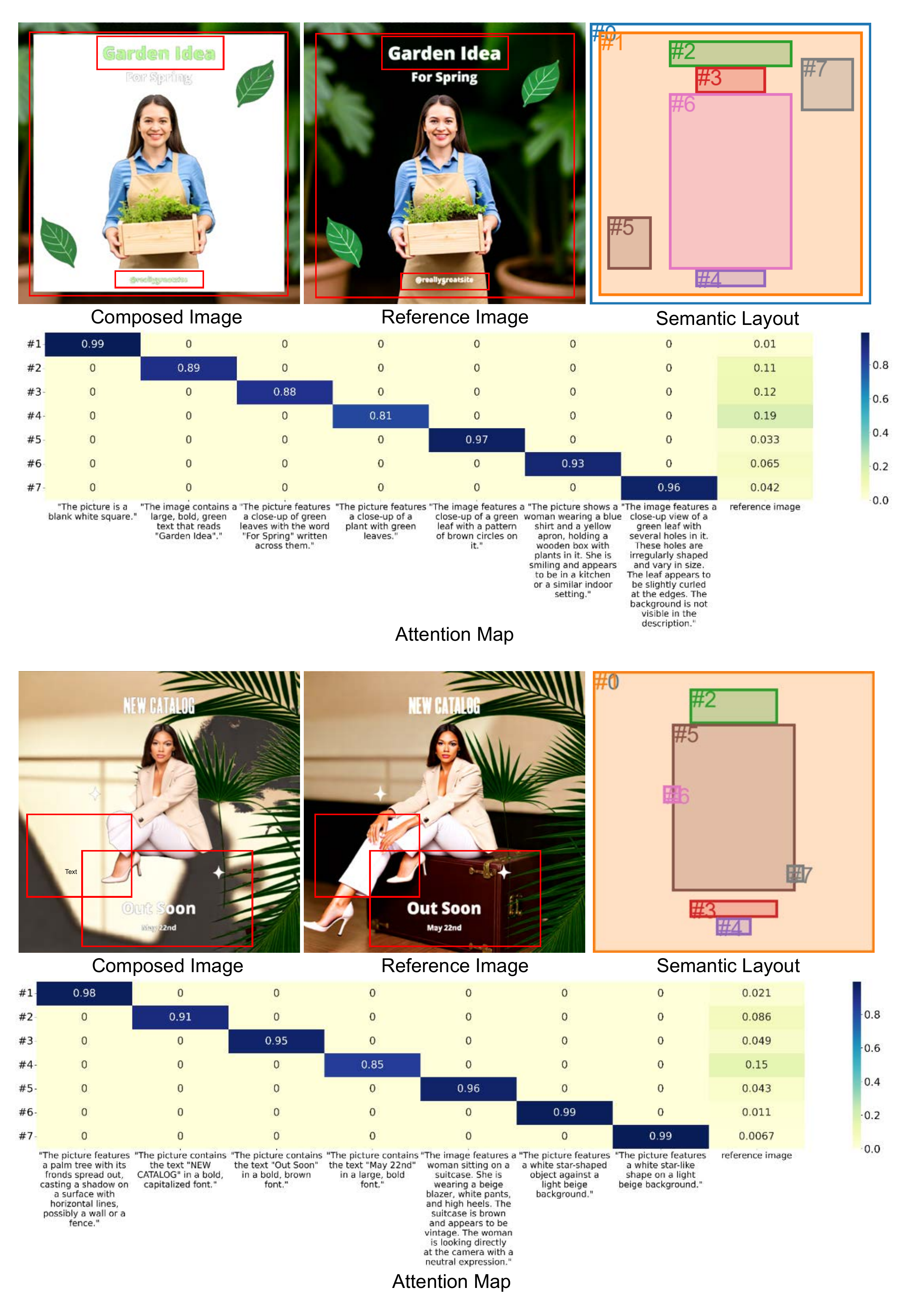}
\vspace{-3mm}
\end{subfigure}
\vspace{-5mm}
\caption{\footnotesize{
\textbf{Conflicts presented in Semantic Layout based Results}:
We display the composed entire image in the 1st column, the reference image in the 2nd column, and the semantic layout in the 3rd column. The conflicted regions are marked with red bounding boxes in both the composed entire images and the reference images. We visualize the attention maps between semantic regions, region-wise prompts, and the global reference images.
}}
\label{fig:conflict_case}
\end{minipage}
\vspace{3mm}
\begin{minipage}[!t]{1\linewidth}
\begin{subfigure}[b]{1\textwidth}
\centering
\includegraphics[width=0.8\textwidth]{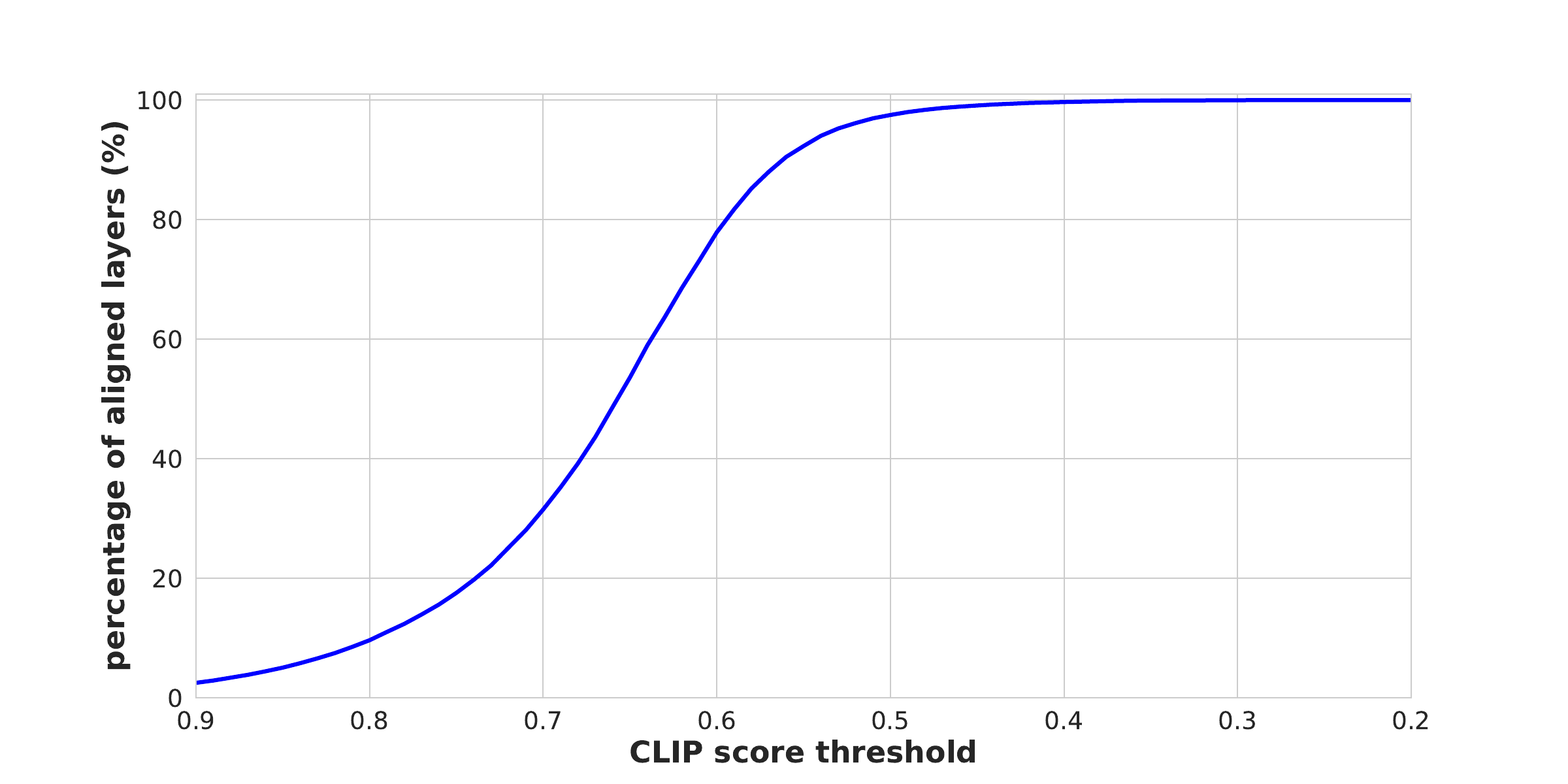}
\vspace{-3mm}
\end{subfigure}
\caption{\footnotesize{
\textbf{Percentage of Inferred Label Assignments Matching Human Annotations}
}}
\label{fig:percentage_threshold}
\end{minipage}
\end{figure}

Figure~\ref{fig:percentage_threshold} plots the curve of the percentage of aligned layers at different CLIP score thresholds, based on statistics from the test set consisting of 5,000 multi-layer transparent images. We attribute the alignment between the inferred label assignments from the generative model and the human annotations to Schema Theory.

\section{Qualitative Multi-Layer Transparent Image Generation Results with around 50 Layers}
\begin{figure*}[h]
\begin{tabular}{|p{0.35\textwidth}|p{0.55\textwidth}|}
\hline
\begin{center}
\includegraphics[height=6cm,keepaspectratio]{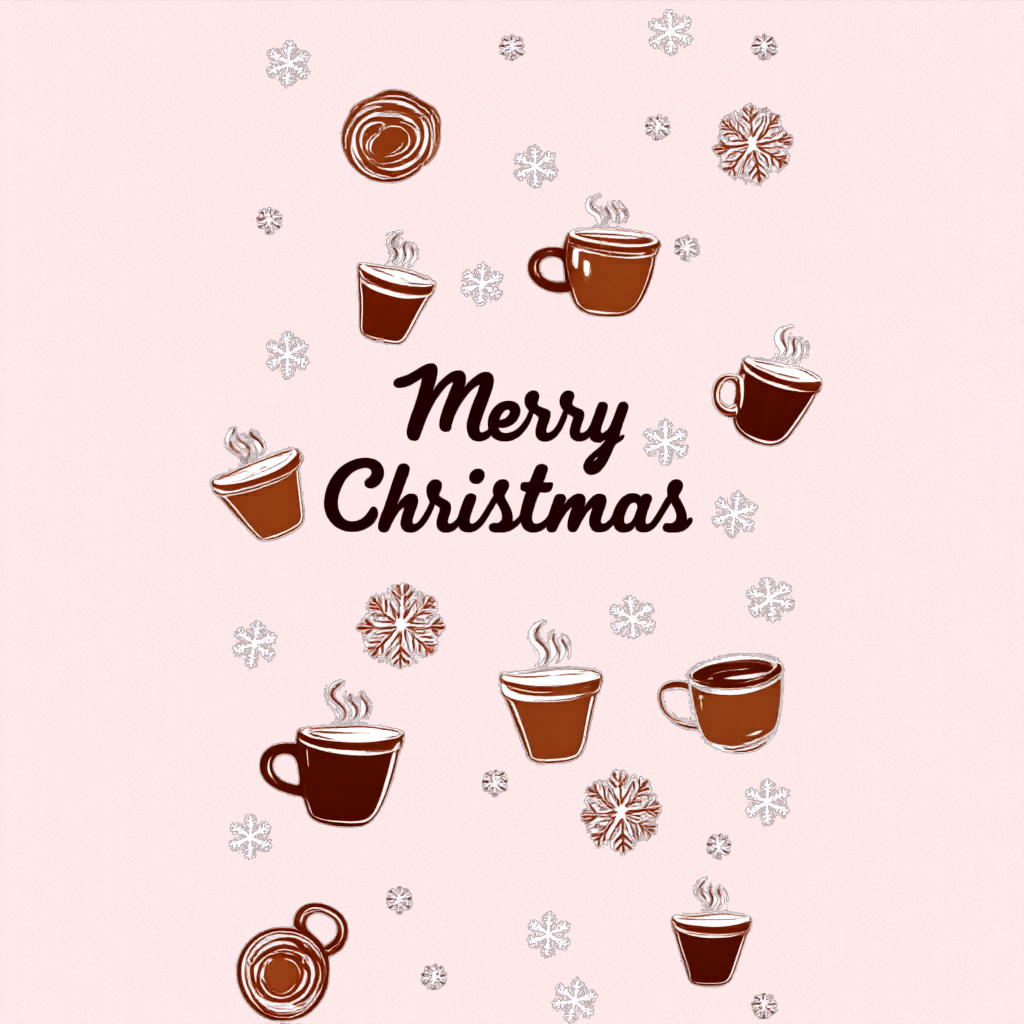}
\end{center}
& 
\begin{flushleft}

\includegraphics[height=6cm,keepaspectratio]{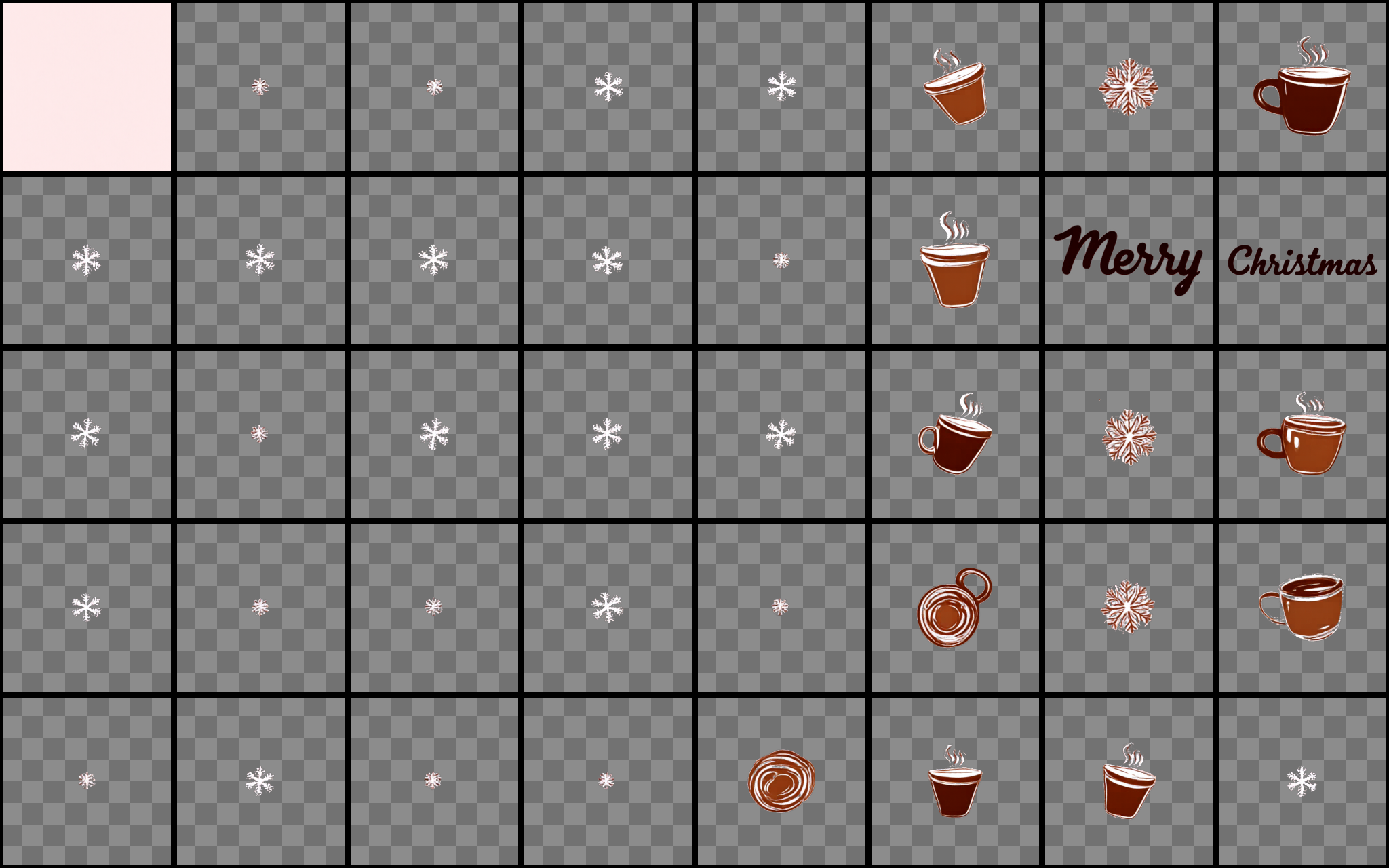}
\end{flushleft} \\
\hline
\begin{center}
\includegraphics[height=6cm,keepaspectratio]{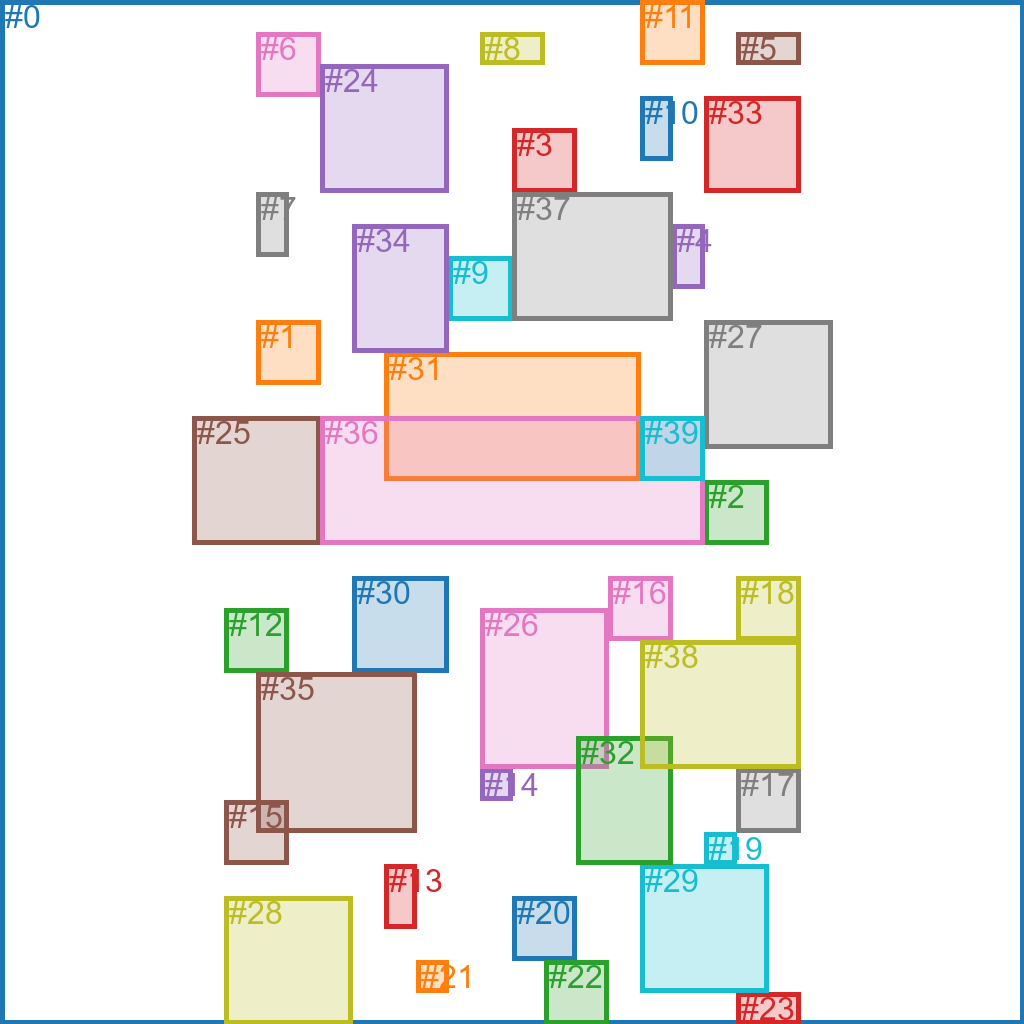}
\end{center}
& 
\begin{flushleft}
{\small The image showcases a festive and cozy Christmas-themed design. The background is a soft, pastel pink, setting a warm and inviting tone. Scattered across the design are holiday-inspired elements that evoke the magic of the season. Central to the theme are illustrations of coffee cups, each uniquely styled. Some feature intricate holiday patterns, while others have minimalist designs, all steaming with warmth, symbolizing comforting hot beverages perfect for the season. Complementing the cozy vibe are delicate snowflakes in various shapes and sizes, scattered like a gentle snowfall, adding a wintry charm to the scene. In the center, the phrase "Merry Christmas" stands out in a cursive, handwritten-style font. The darker-colored text contrasts beautifully with the soft background, giving the message a friendly and personal touch. Altogether, the design blends these elements seamlessly to create a cheerful and heartwarming Christmas greeting, embodying the joy and warmth of the holiday season.}
\end{flushleft} \\
\hline
\end{tabular}
\caption{Generated Result with 40 transparent image layers. Top-left: Generated Merged Image; Top-Right: Generated Transparent Layers; Bottom-left: Anonymous Region Layout; Bottom-right: Global Prompt.}
\label{fig:more_layer_40}
\end{figure*}

\begin{figure*}[h]
\begin{tabular}{|p{0.35\textwidth}|p{0.62\textwidth}|}
\hline
\begin{center}
\includegraphics[height=6cm,keepaspectratio]{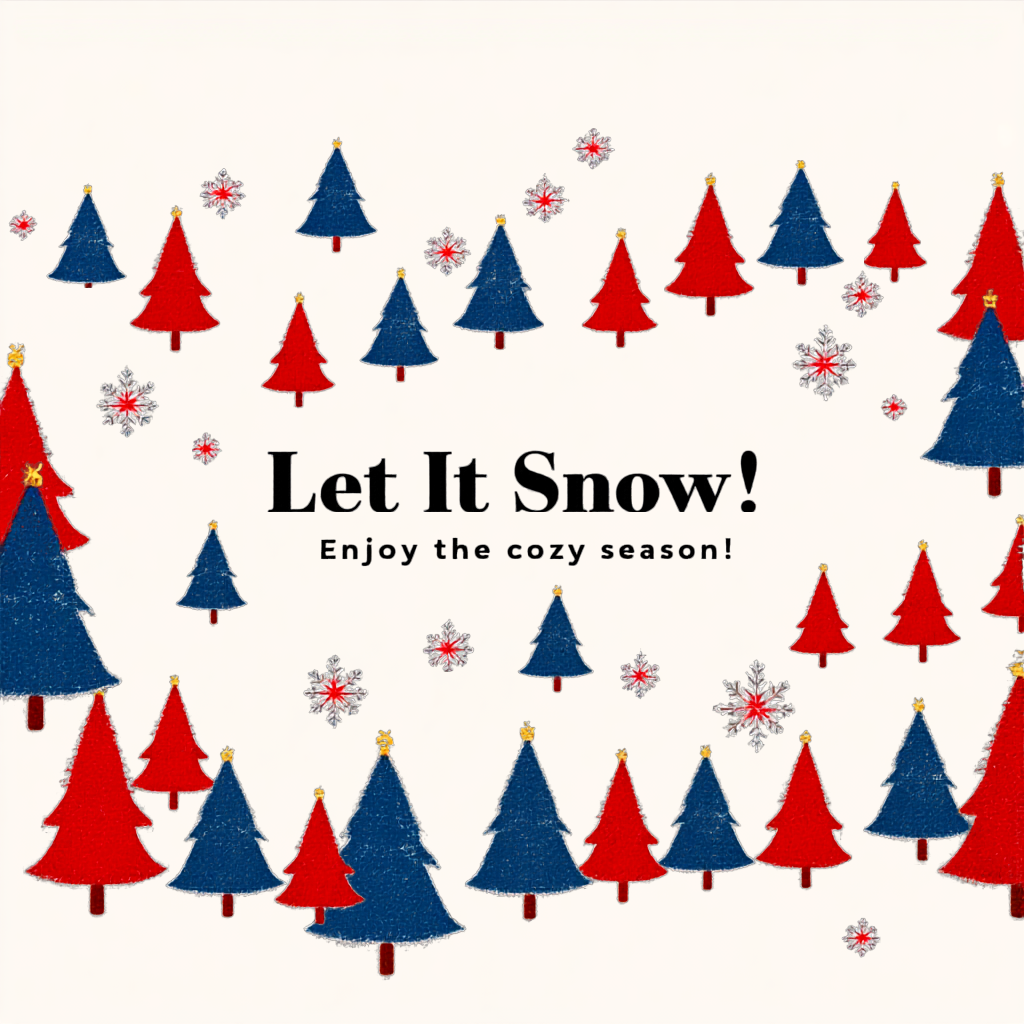}
\end{center}
& 
\begin{flushleft}

\includegraphics[height=6cm,keepaspectratio]{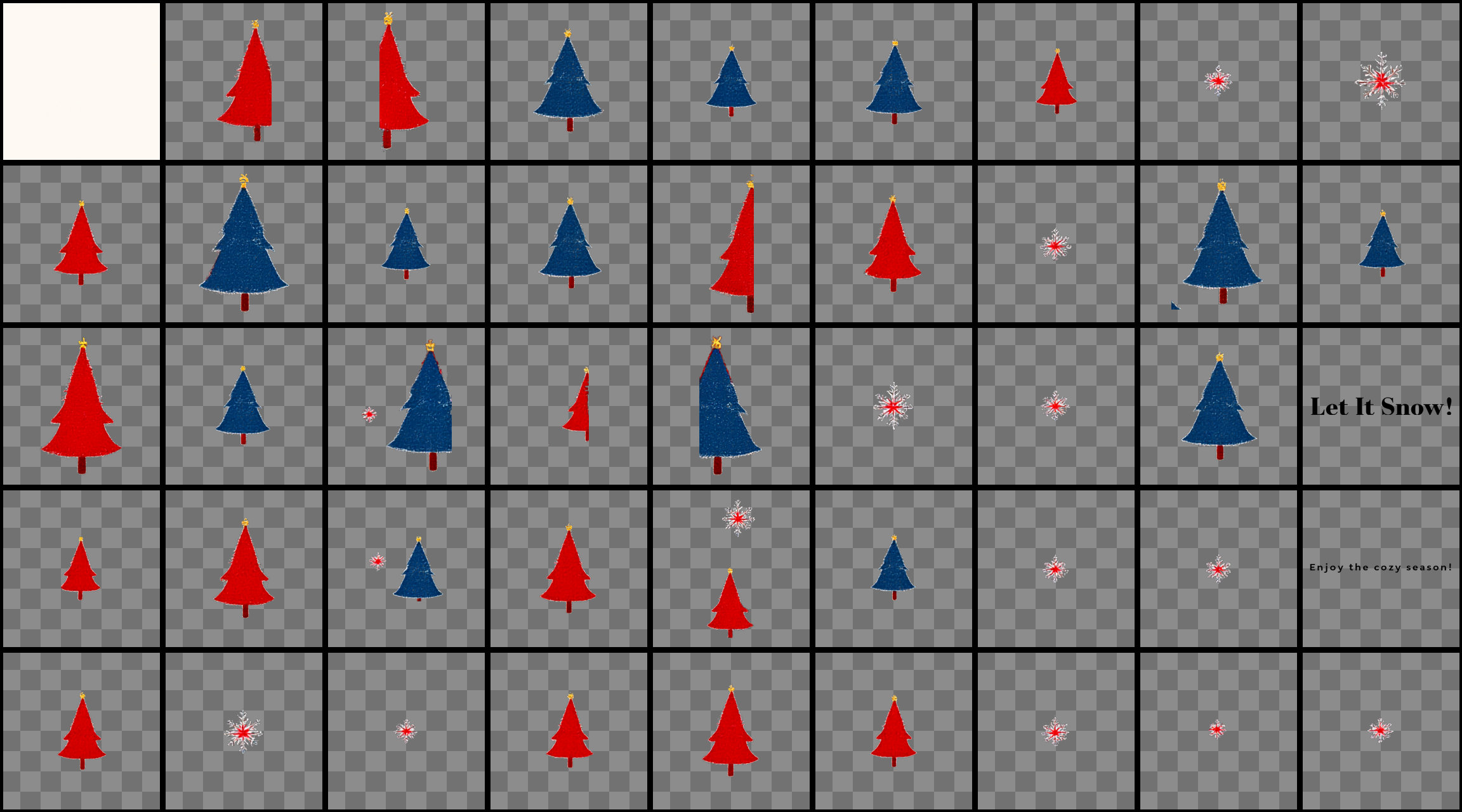}
\end{flushleft} \\
\hline
\begin{center}
\includegraphics[height=6cm,keepaspectratio]{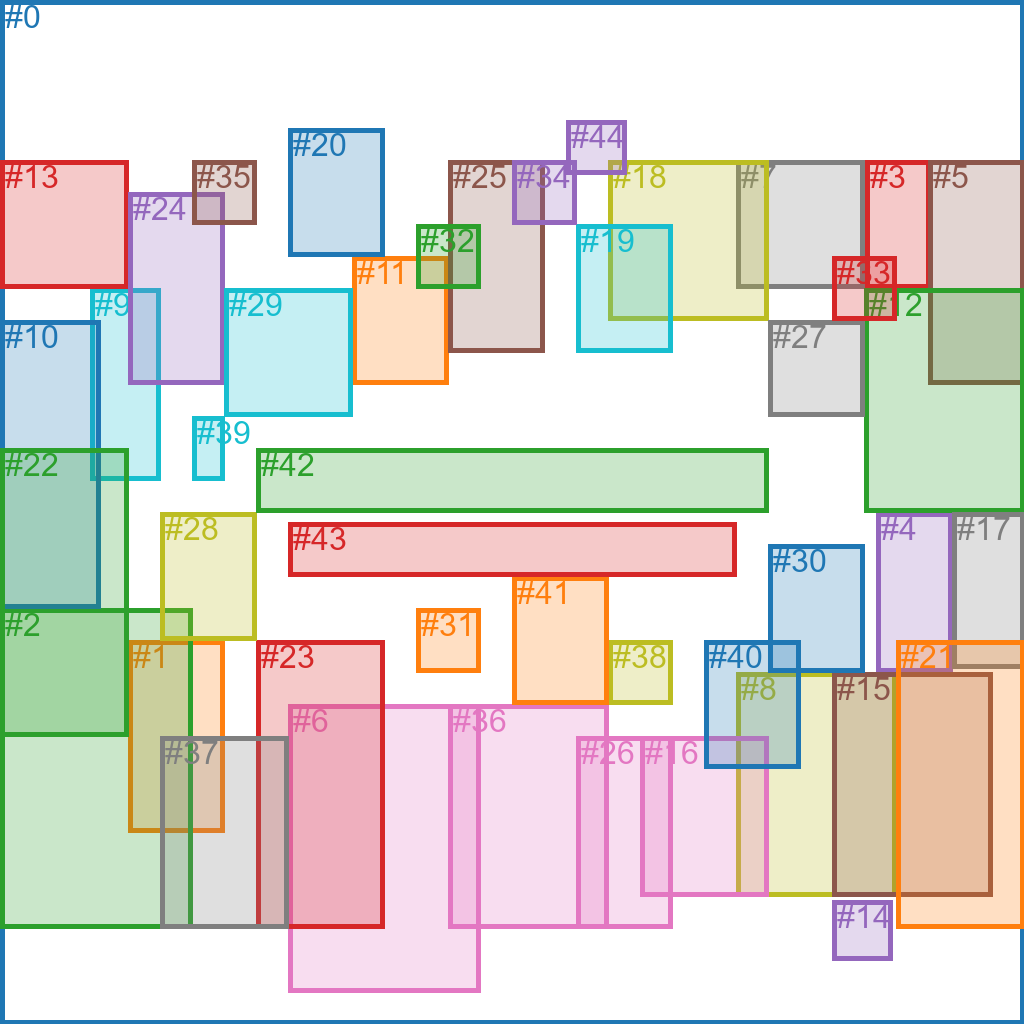}
\end{center}
& 
\begin{flushleft}
{\small The image is a vibrant digital illustration with a festive holiday theme. It showcases a collection of stylized Christmas trees in varying sizes and colors, featuring shades of blue, red, and green. The trees are scattered playfully across the design, with some adorned with snowflakes, evoking a wintry, snowy atmosphere. At the center of the image, bold and festive text reads "Let It Snow!" in a lively font, capturing the essence of the season. Just below, a smaller text offers the cheerful message "Enjoy the cozy  season!" adding a warm and inviting touch. The background is a light, neutral tone that enhances the contrast with the vibrant trees and text, making the design elements pop. The overall style is bright and cheerful, perfectly suited for a holiday greeting card or seasonal decoration.}
\end{flushleft} \\
\hline
\end{tabular}
\caption{Generated Result with 45 transparent image layers. Top-left: Generated Merged Image; Top-Right: Generated Transparent Layers; Bottom-left: Anonymous Region Layout; Bottom-right: Global Prompt.}
\label{fig:more_layer_45}
\end{figure*}

\begin{figure*}[h]
\begin{tabular}{|p{0.32\textwidth}|p{0.68\textwidth}|}
\hline
\begin{center}
\includegraphics[height=5.2cm,keepaspectratio]{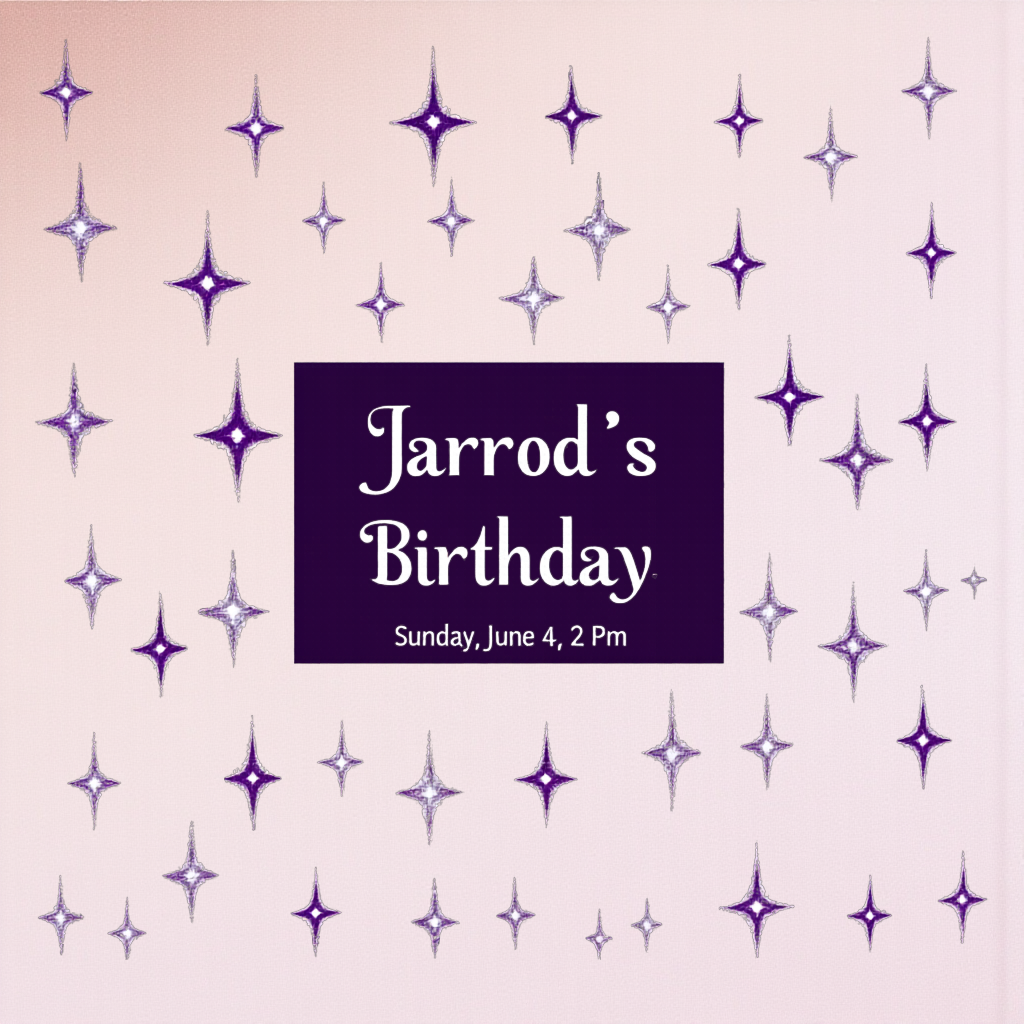}
\end{center}
& 
\begin{flushleft}

\includegraphics[height=5.2cm,keepaspectratio]{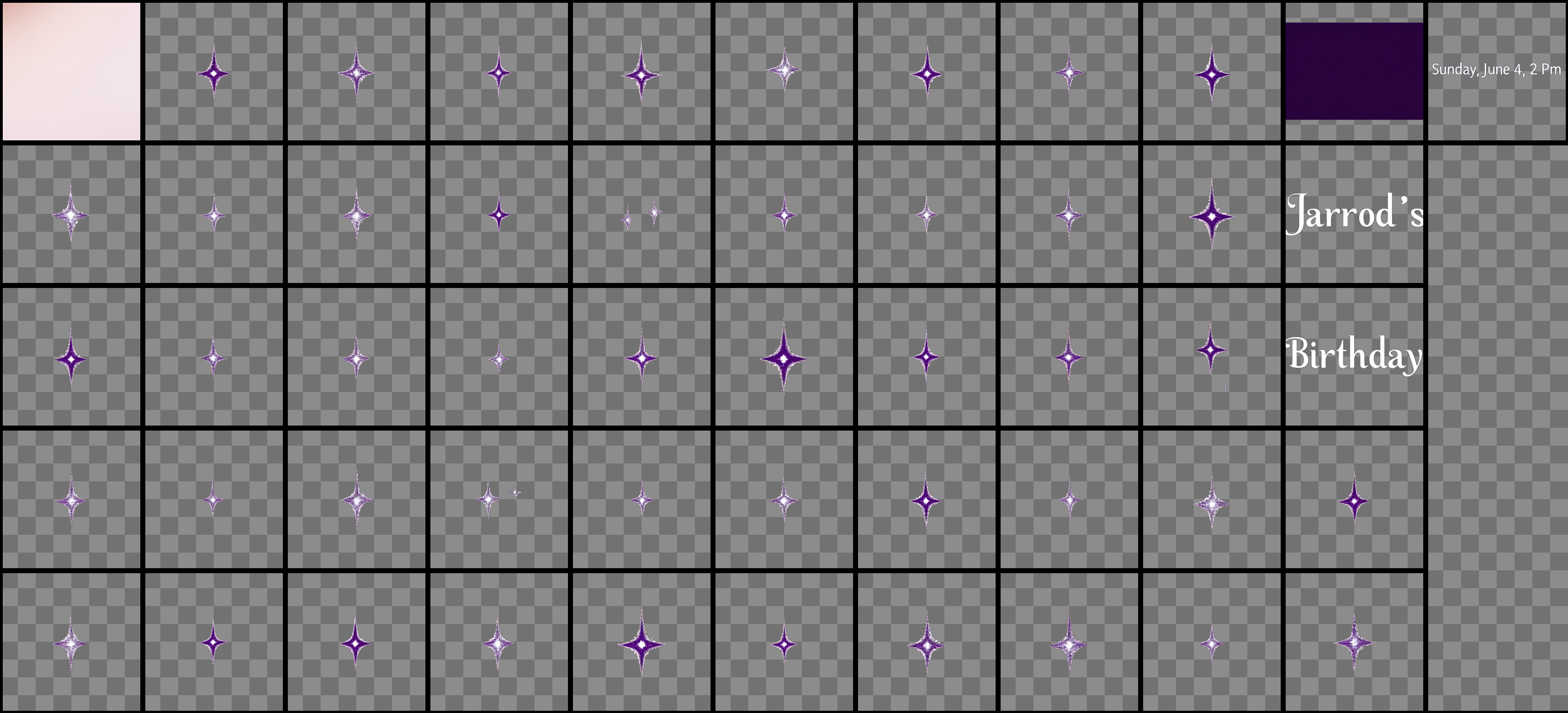}
\end{flushleft} \\
\hline
\begin{center}
\includegraphics[height=5.2cm,keepaspectratio]{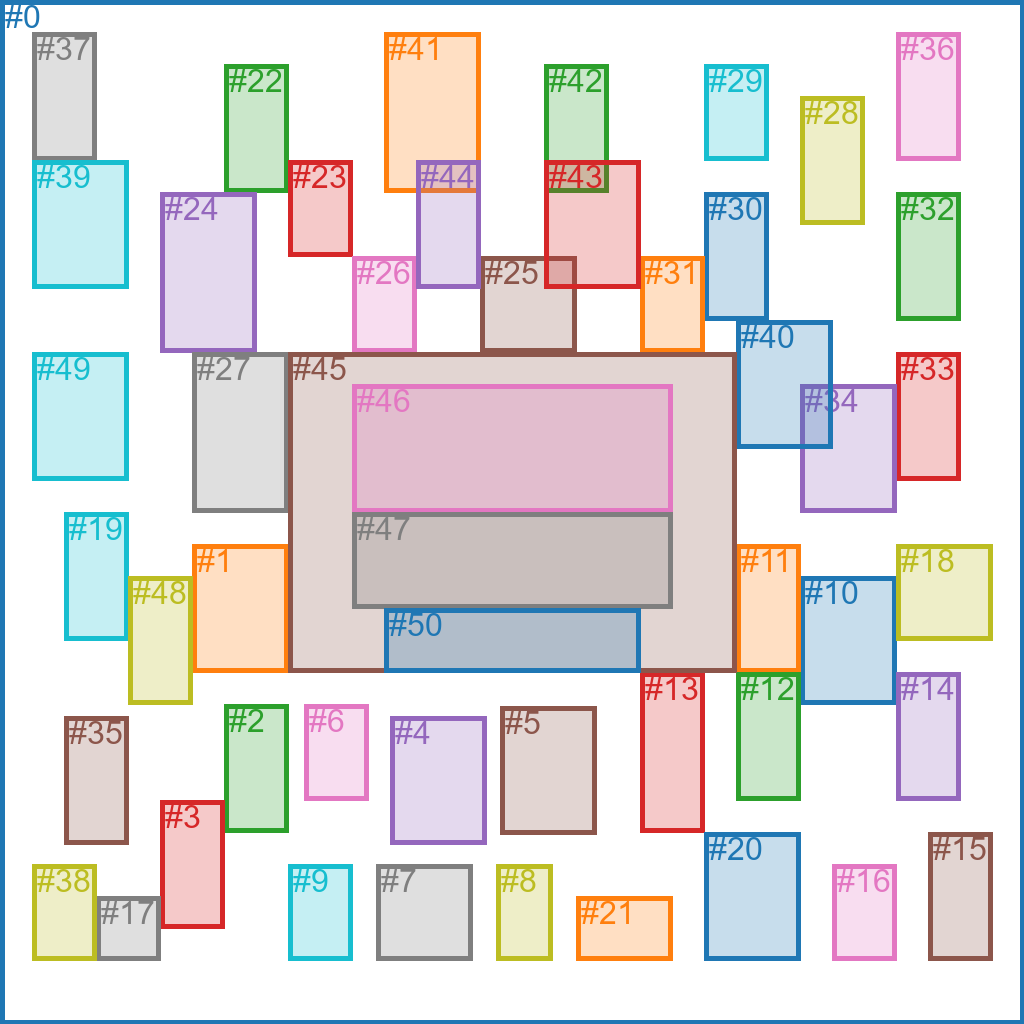}
\end{center}
& 
\begin{flushleft}
{\small The image is a beautifully designed celestial graphic, radiating a sense of wonder and elegance. It features an array of shimmering stars, soft glows, and delicate constellations in hues of silver, gold, and lavender, artistically scattered across a twilight sky backdrop. These celestial elements are depicted with intricate, flowing patterns that evoke a sense of ethereal beauty and tranquility. At the center of the design is a striking rectangular banner in a soft lavender hue, accented with a crisp white border. The banner draws attention with bold white text that reads: "JARROD'S BIRTHDAY" in an eye-catching, large font. Beneath it, in a smaller yet equally clear white font, the details continue: "SUNDAY, JUNE 4, 2 PM." While partially obscured, the date and time details are presented in a clean, standard format. The overall style of this invitation is dreamy, celebratory, and enchanting, with its celestial theme and pastel color palette evoking the feel of a magical, starlit evening. The design perfectly captures the essence of a joyful birthday celebration, making it both inviting and unforgettable.}
\end{flushleft} \\
\hline
\end{tabular}
\caption{Generated Result with 51 transparent image layers. Top-left: Generated Merged Image; Top-Right: Generated Transparent Layers; Bottom-left: Anonymous Region Layout; Bottom-right: Global Prompt.}
\label{fig:more_layer_50}
\end{figure*}

One key advantage of our approach is its ability to support the generation of tens of high-quality transparent layers from a global prompt and an ultra-dense anonymous region layout. We present the generated multi-layer image results with 40, 45, and 51 layers in \Cref{fig:more_layer_40,fig:more_layer_45,fig:more_layer_50}, respectively. These results highlight our method’s capability to generate an \emph{exceptionally high} number of layers, in contrast to previous works, which are limited to generating only a small number of layers.

\section{Implementation of Layout Conditional Multi-Layer 3D RoPE}
We present the PyTorch implementation of the proposed layout-conditional multi-layer 3D RoPE in Algorithm~\ref{algo:3drope_1} and its usage within the Attention Module in Algorithm~\ref{algo:3drope_2}.

\section{Layout Variation}
One key advantage of our approach is that our Anonymous Region Transformer generalizes to various layouts given a fixed global prompt. The ART model is capable of adaptively assigning semantic concepts to fit diverse anonymous region layouts. We illustrate some qualitative results in ~\Cref{tab:variant_layout1,tab:variant_layout2,tab:variant_layout3,tab:variant_layout4,tab:variant_layout5,tab:variant_layout6,tab:variant_layout7,tab:variant_layout8,tab:variant_layout9}.

\section{Layer-wise Editing}

\begin{figure*}
    \centering
    \includegraphics[width=0.64\linewidth]{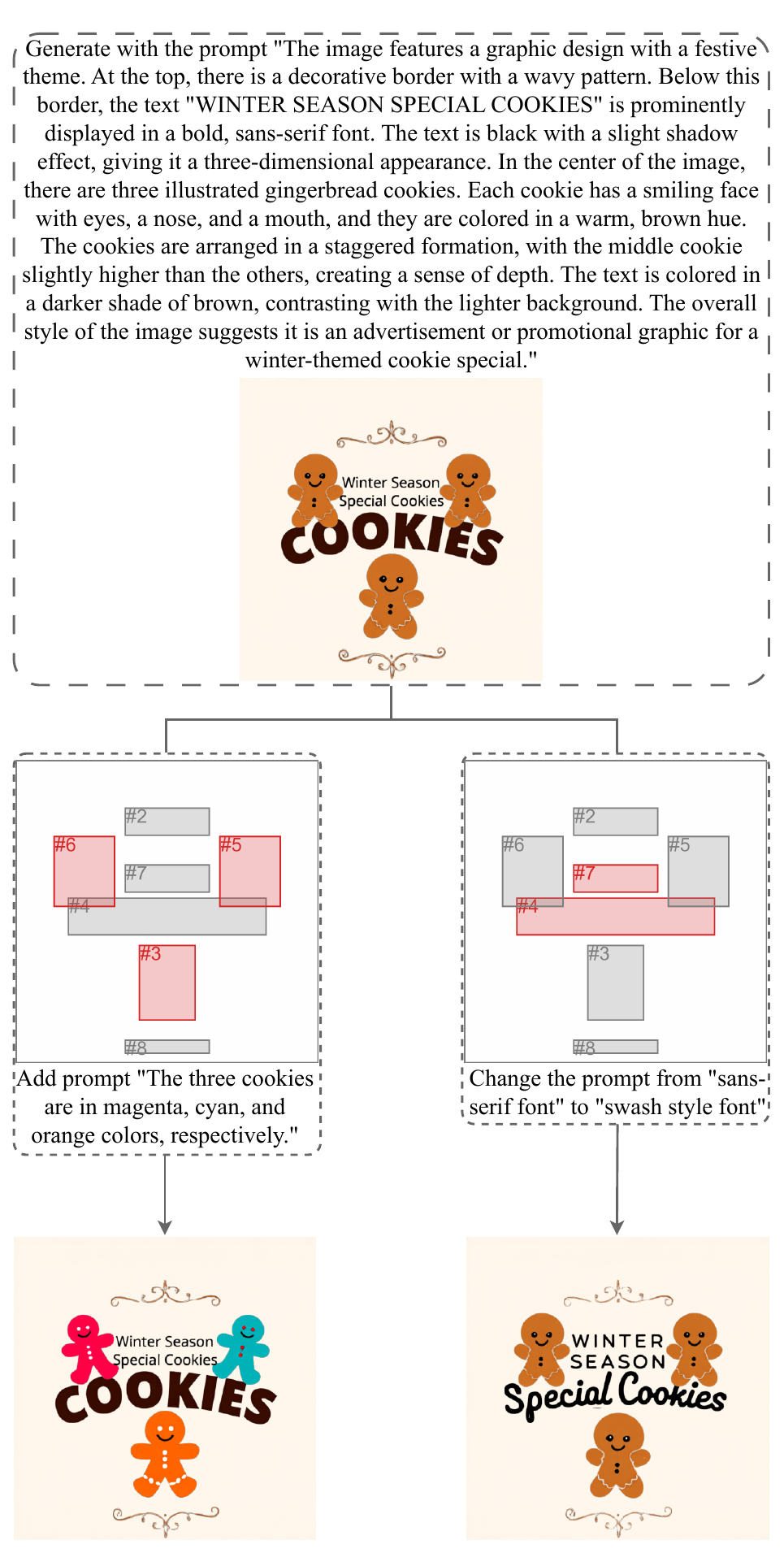}
    \caption{Layer-wise editing of the generated image.}
    \label{fig:layer_edit_1}
\end{figure*}

The purpose of the experiment is to demonstrate the effectiveness of the proposed ART method in enabling layer-wise image editing, specifically the accurate regeneration of contents on specific layers. 
The layer-wise editing pipeline consists of three steps: modifying the input prompt, regenerating the layers that need to be edited, and freezing the remaining layers. 
We have provided an editing result in Figure~\ref{fig:layer_edit_1}. As can be observed, our model can accurately regenerate specific content on the editable layers to meet the requirements from the input prompt. Moreover, the newly generated layer remains harmonious with the rest while keeping other layers unchanged, providing a feasible approach to precisely and independently control the style and contents of each layer.

\section{Details of Transparency Encoding}

Here, we provide additional details on the transparency encoding introduced in \Cref{sec:method:ml_vae}. The overall goal is to transform a 4-channel RGBA image into its 3-channel RGB counterpart, facilitating the reuse of pretrained three-channel image generation models while effectively embedding the alpha channel information into the RGB channels.

For each RGBA image $\mathbf{I}_\text{fg}^{i} \in \mathbb{R}^{H_i \times W_i \times 4}$, we first linearly normalize the three RGB channels $\mathbf{I}_{\text{fg},\text{RGB}}^{i} \in \mathbb{R}^{H_i \times W_i \times 3}$ from the range $[0, 255]$ to $[-1, 1]$, following the standard practice in Flux.1 models. Similarly, we linearly transform the alpha channel $\mathbf{I}_{\text{fg},\alpha}^{i} \in \mathbb{R}^{H_i \times W_i \times 1}$ from $[0, 255]$ to $[-1, 1]$, where $-1$ represents fully transparent pixels and $1$ represents fully opaque pixels.

To encode transparency information from the alpha channel into the RGB channels, we apply the following transformation:  
\[
\hat{\mathbf{I}}_\text{fg}^{i} = (0.5\mathbf{I}_{\text{fg},\alpha}^{i} + 0.5) \times \mathbf{I}_{\text{fg},\text{RGB}}^{i}.
\]
Here, the coefficient $(0.5\mathbf{I}_{\text{fg},\alpha}^{i} + 0.5)$ linearly maps the alpha channel from $[-1, 1]$ to $[0, 1]$. This ensures that the RGB values of fully opaque pixels remain unchanged, while fully transparent pixels are mapped to pure gray (RGB = $(0, 0, 0)$ in the $[-1, 1]$ range). Semi-transparent pixels undergo a proportional transformation based on their alpha values.

\section{Evaluation in text generation}

\begin{table}[htbp]
\vspace{-4mm}
\begin{minipage}[t]{1\linewidth}  
\centering 
\tablestyle{6pt}{1.1}
\resizebox{0.99\linewidth}{!}
{
\begin{tabular}{l|c|ccc}  
Method & PSNR{$^{\textrm{layer}}_{\textrm{rgb}}$} & PSNR{$^{\textrm{layer}}_{\textrm{alpha}}$} & PSNR & FID$\scriptstyle \text{merged}$ \\
\shline
Single-layer Autoencoder w/ CNN  & 30.10 & 20.12 & 26.88 & 5.12 \\
Single-layer Autoencoder w/ ViT  & 33.64 & 22.47 & 28.76 & 3.39 \\
Multi-layer Autoencoder w/ ViT & \underline{\textbf{34.80}} & \underline{\textbf{24.25}} & \underline{\textbf{31.37}} & \underline{\textbf{2.76}} \\
\end{tabular}
}
\vspace{-2mm}
\caption{
\footnotesize{Ablation of autoencoder (all trained with our MLTD data).}}
\label{tab:text_decoder_comparison}
\end{minipage}
\vspace{-4mm}
\end{table}

Here we provide more evaluation 
for the advantages of our multi-layer transparent image autoencoder, which has been previously illustrated in Figure~\ref{fig:sota_decoder_compare}. The images are generated by encoding and decoding the same ground-truth image, which effectively reflects the quality of the reconstructed multi-layer images. The superior performance in text generation of our method can be attributed to the following key factors: (1) the use of Vision Transformer (ViT) for visual text modeling, which outperforms CNN-based autoencoders by predicting more accurate edges. In contrast, both LayerDiffuse and Flux-RGBA rely on CNN-based autoencoders; (2) the multi-layer autoencoder architecture, which enables explicit interactions across different layers by jointly encoding and decoding them, leading to better performance compared to single-layer methods. Additionally, our results benefit from the multi-layer transparent design dataset (MLTD), which includes a larger number of visual text layers. As shown in Table~\ref{tab:text_decoder_comparison}, replacing CNN with ViT and adopting a multi-layer structure both contribute to improved performance.

\begin{table*}[t]
    \centering
    \renewcommand{\arraystretch}{1.2}
    \begin{tabular}{|m{0.25\textwidth}|m{0.22\textwidth}|m{0.5\textwidth}|}
        \hline
        \small\textbf{Multi-layer Transparent Image} & \small\textbf{Anonymous Region Layout} & \small\textbf{Global Prompt} \\
        \hline
        \includegraphics[height=2.6cm]{fig/only_ours/compose_case139.png}& \includegraphics[height=2.6cm]{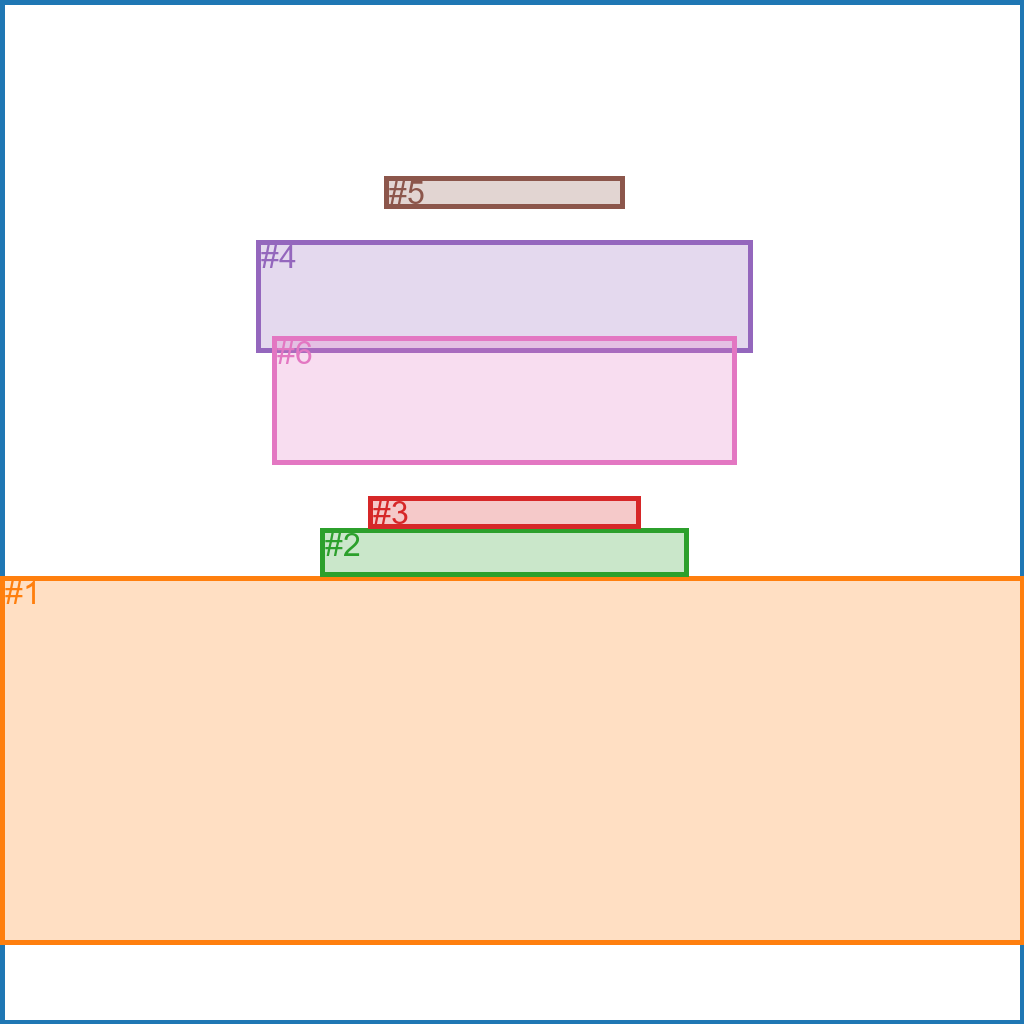} 
        & {\footnotesize The image is a poster for an Autumn Festival. The festival is scheduled to take place from October 15th to October 21st. The poster features a variety of autumn-themed elements, including pumpkins, leaves, and berries. The text on the poster is in a playful, handwritten font, and it reads "let's celebrate Autumn Festival". The poster also includes a list of activities that will be available at the festival, such as games, food, and music. The overall color scheme of the poster is warm, with shades of orange, yellow, and green, which are typical colors associated with autumn. The poster is designed to be eye-catching and inviting, encouraging people to come and enjoy the festival.} \\
        \hline
        \includegraphics[height=2.6cm]{fig/only_ours/compose_case75.png} & \includegraphics[height=2.6cm]{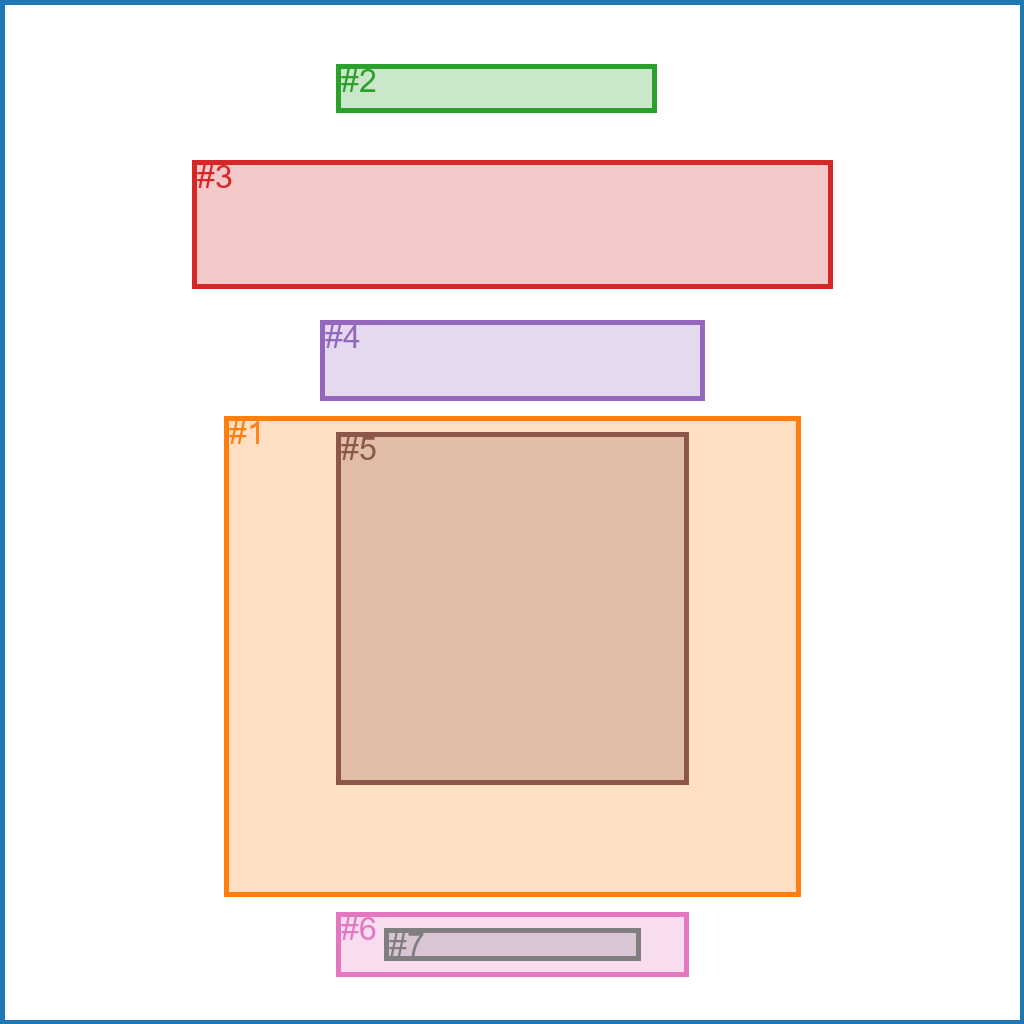} & {\footnotesize A promotional flyer for a photography workshop hosted by Photo Studio, Inc. on April 12 at 9:00 AM. It features a vintage camera illustration and a "Register Now!" button at the bottom.} \\
        \hline
        \includegraphics[height=2.6cm]{fig/only_ours/compose_case585.png}\ & \includegraphics[height=2.6cm]{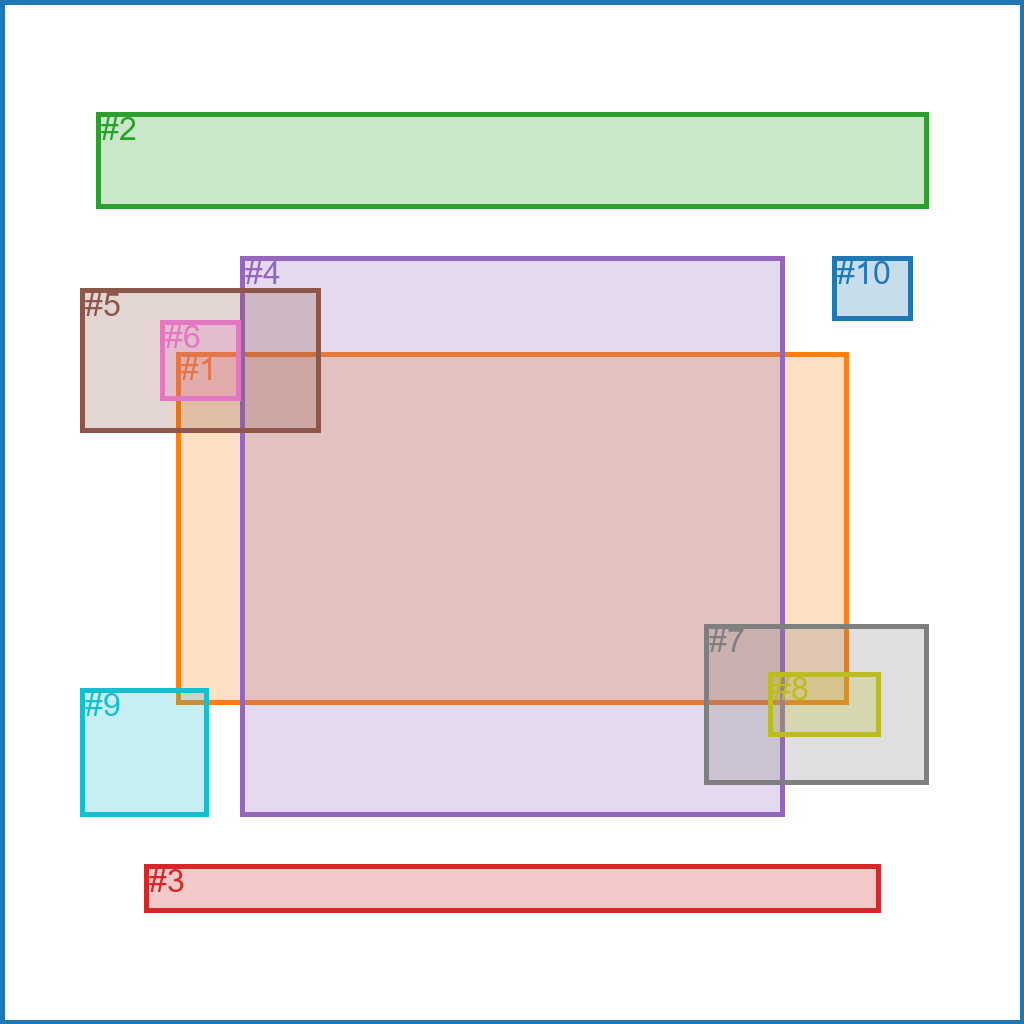} & {\footnotesize A promotional Easter-themed graphic featuring a large, colorful egg with text "AFFORDABLE EASTER" at the top. It includes discount badges stating "50\% OFF" and "ORDER TODAY" on either side of the egg, with the tagline "Essentials Without Breaking the Bank" at the bottom.} \\
        \hline
        \includegraphics[height=2.6cm]{fig/only_ours/compose_case1296.png} & \includegraphics[height=2.6cm]{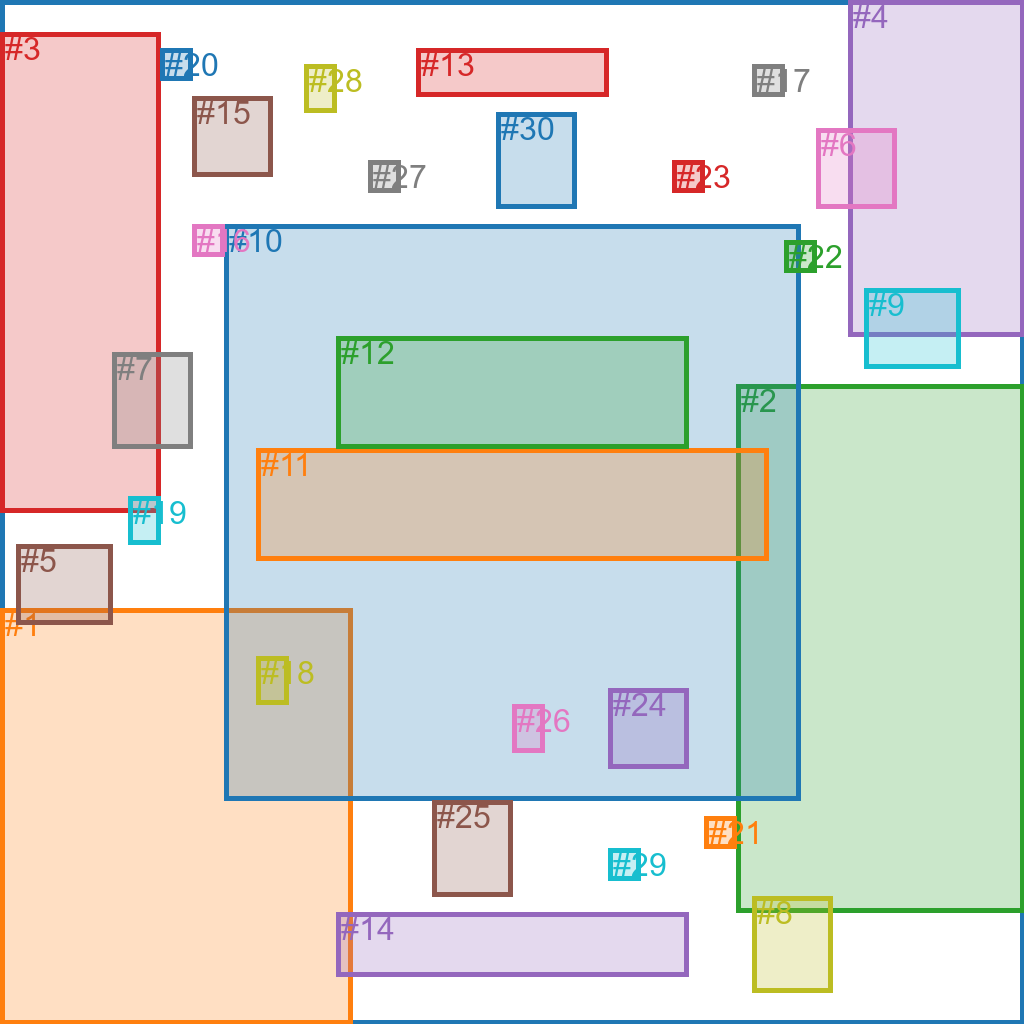} & {\footnotesize A festive birthday card design features an orange speech bubble with "Happy Birthday" text in white, surrounded by balloons, stars, and cakes with candles. The top reads "Your store" and the bottom displays "www.yourweb.com". The background is light with playful elements creating a cheerful vibe.} \\
        \hline
    \end{tabular}
    \vspace{-1mm}
    \caption{Detailed anonymous region layouts and global prompts for multi-layer image generation in Figure 5 of the main paper.}
    \label{tab:generation_results_1}
\end{table*}
\newpage

\begin{table*}[t]
    \centering
    \renewcommand{\arraystretch}{1.2}
    \begin{tabular}{|m{0.25\textwidth}|m{0.22\textwidth}|m{0.5\textwidth}|}
        \hline
        \small\textbf{Multi-layer Transparent Image} & \small\textbf{Anonymous Region Layout} & \small\textbf{Global Prompt} \\
        \hline
        \includegraphics[height=2.6cm]{fig/only_ours/compose_case7252_seed1.png} & \includegraphics[height=2.6cm]{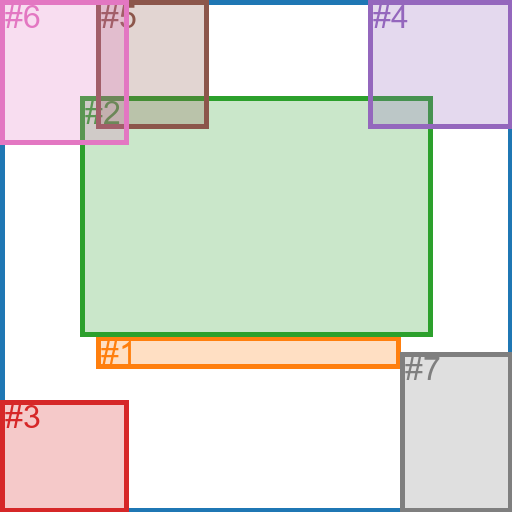} & {\tiny The image is a romantic and spiritual graphic design, likely intended for a summer camp brochure. The overall design showcases a citrus-inspired palette, featuring vibrant oranges, yellows, and soft greens, which enhances its sophisticated and refreshing atmosphere. Styled in ornamental calligraphy, the design features seamless patterns that evoke a sense of harmony and continuity, appealing to fashion-forward thinkers who appreciate intricate details. The title, "Summer Spirit Camp," is written in a Brush font, with a size of 95px. Positioned at the top of the image, it is bold and immediately captures the viewer's attention, setting a tone of elegance and anticipation. Below the title, a secondary text reads "Embrace Nature, Nurture the Soul," sized at 100px. This text complements the main title by highlighting the camp's core values, inviting viewers to explore a deeper connection with nature and spirituality. At the bottom, another piece of text states "Join us from June 10-15, 2023," written in a smaller 100px font. This serves as supporting information, providing essential logistics such as the date, ensuring clarity and accessibility for potential attendees. The text content in this design is specific and directly contributes to the overall purpose of the graphic, effectively conveying the essence of the summer camp experience. This design captures an artistic and creative spirit, making it both visually striking and emotionally resonant. The seamless integration of text and imagery creates a cohesive narrative that resonates with the intended audience, inviting them to embark on a transformative journey.} \\
        \hline
        \includegraphics[height=2.6cm]{fig/only_ours/compose_case6623_seed1.png} & \includegraphics[height=2.6cm]{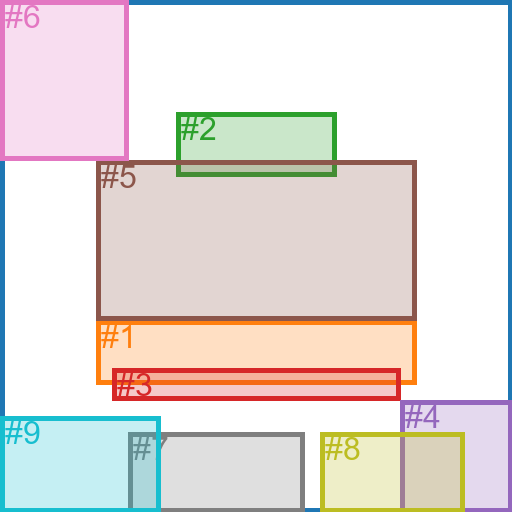} & {\tiny The image is a romantic and polished graphic design, likely intended for an environmental campaign. The overall design showcases a tropical greens and yellows color scheme that enhances its ornamental atmosphere. Styled in hand-drawn doodles, the design features whimsical hand-lettering, adding to its appeal for luxury consumers. This design captures a thoughtful and balanced aesthetic, making it both visually striking and emotionally resonant. Text elements play a crucial role in conveying the message. The title reads "Join the Green Revolution," written in a Condensed serif font, with a size of 88px. Positioned at the top of the image, it is bold and immediately captures the viewer's attention. Below the title, a secondary text reads "Sustainable Living," providing additional information and complementing the main title. This text is sized at 24px, maintaining a harmonious balance with the title. At the bottom, another piece of text states "Save Our Planet, One Step at a Time," written in a smaller 18px font. This serves as supporting information, encouraging action and engagement. The text content in this design is specific and directly contributes to the overall purpose of the graphic. The title announces the campaign's mission, the secondary text highlights the theme, and the footer provides a motivational call to action. The combination of tropical colors, hand-drawn elements, and carefully chosen typography creates a design that is both visually appealing and emotionally impactful, resonating with an audience passionate about environmental sustainability.} \\
        \hline
        \includegraphics[height=2.6cm]{fig/only_ours/compose_case62b32ca8079dcd9363c3e0ab_seed3.png} & \includegraphics[height=2.6cm]{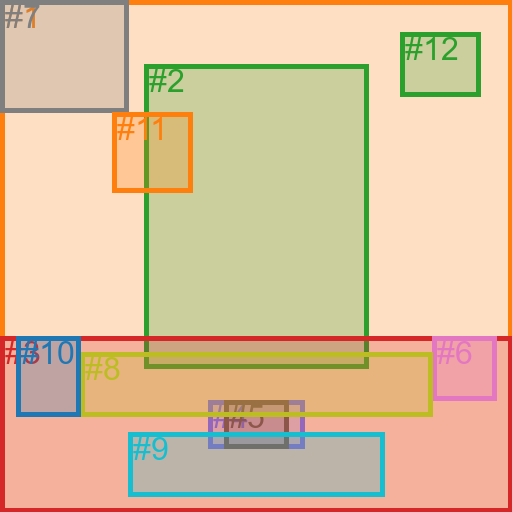} & {\footnotesize A promotional graphic features a person in a white outfit and headscarf holding an ice cream cone, with the text "New Arrival" and "Get ready to shine bright and make impression" on a stylish, modern background with grid and floral elements.} \\
        \hline
        \includegraphics[height=2.6cm]{fig/only_ours/compose_case6684_seed1.png} & \includegraphics[height=2.6cm]{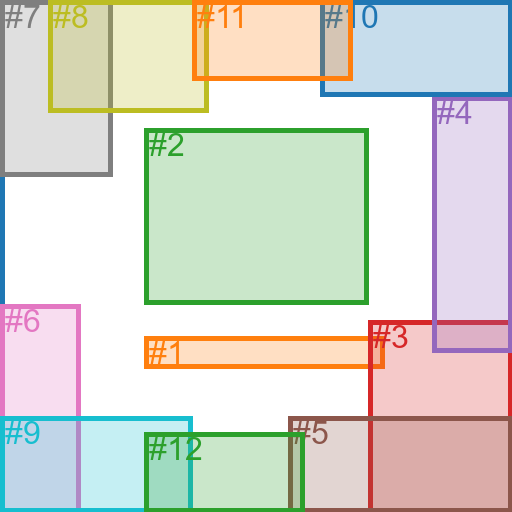} & {\tiny The image is a painterly and soft graphic design, likely intended for a job recruitment poster. The overall design showcases a tropical greens and yellows color scheme, enhancing its structured yet inviting atmosphere. Styled with motion blur visuals, the design features repeating motifs that add a dynamic appeal, making it particularly attractive for tech companies. This design captures a timeless yet modern aesthetic, making it both visually striking and emotionally resonant. Text elements play a crucial role in conveying the message. The title reads "Join Our Tech Team," written in an Italic Serif font, with a size of 20px. Positioned at the top of the image, it is bold and immediately captures the viewer's attention. Below the title, a secondary text reads "Innovate with Us," sized at 72px, providing additional information and complementing the main title. At the bottom, another piece of text states "Apply by October 15th," written in a smaller 36px font. This serves as supporting information, specifying logistics like a deadline. The text content in this design is specific and directly contributes to the overall purpose of the graphic. The title announces the recruitment opportunity, the secondary text highlights the company's mission, and the footer provides essential application details. This cohesive blend of design elements and text creates a compelling invitation for potential candidates, evoking a sense of excitement and opportunity.} \\
        \hline
    \end{tabular}
    \vspace{-1mm}
    \caption{Detailed anonymous region layouts and global prompts for multi-layer image generation in Figure 5 of the main paper.}
    \label{tab:generation_results_1_additional}
\end{table*}
\newpage

\begin{table*}[t]
    \centering
    \renewcommand{\arraystretch}{1.2}
    \begin{tabular}{|m{0.25\textwidth}|m{0.22\textwidth}|m{0.5\textwidth}|}\hline
    \small\textbf{Multi-layer Transparent Image} & \small\textbf{Anonymous Region Layout} & \small\textbf{Global Prompt} \\ \hline
        \includegraphics[height=2.6cm]{fig/cole_benchmark_with_layers/ART_MarketingMaterials_100.png}& \includegraphics[height=2.6cm]{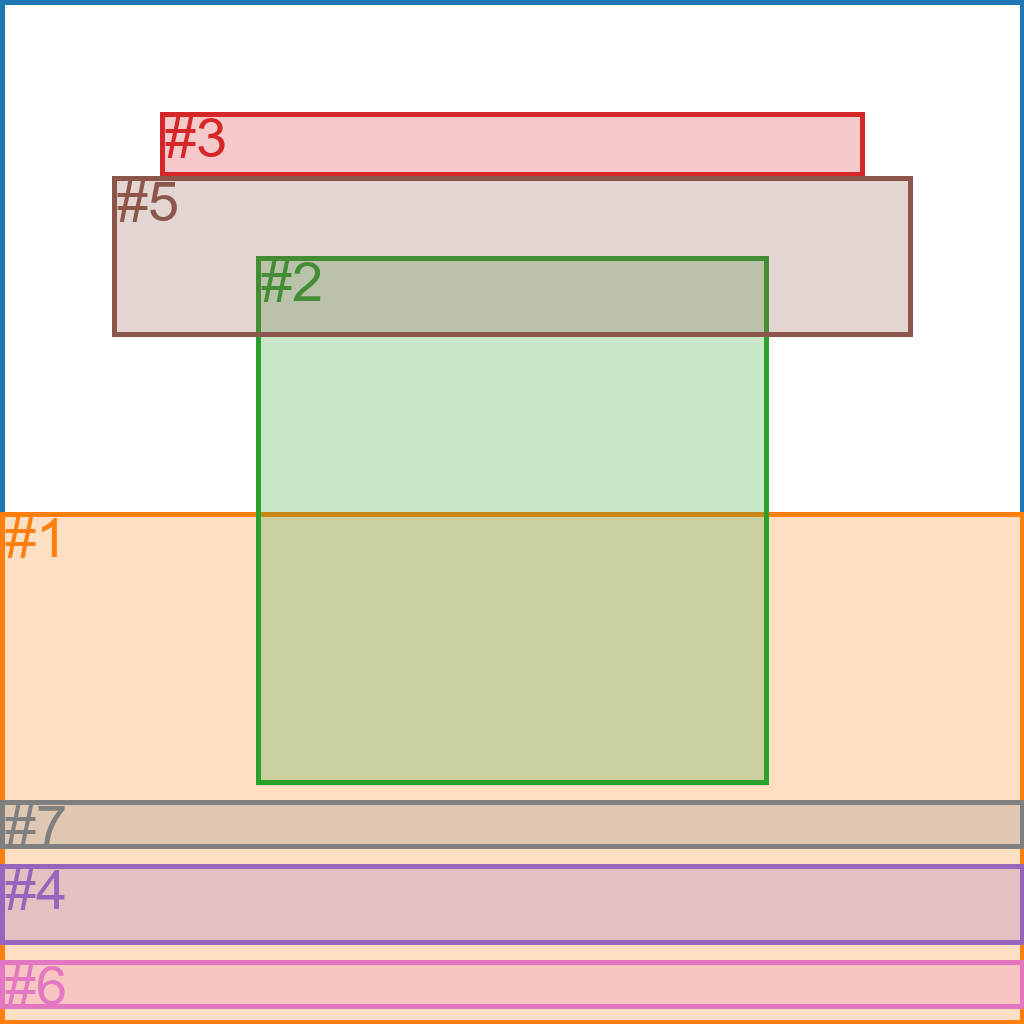} & {\footnotesize The image is designed as a Facebook cover for a website that specializes in selling pregnancy goods. The theme is warm and inviting, geared towards expecting parents. The background features a soft pastel palette, predominantly in shades of baby pink and light blue, which are colors commonly associated with babies and pregnancy. Arranged throughout the image are a selection of baby essentials, which may include items like a plush teddy bear, a set of pastel-colored baby clothes, a small stack of diapers, a baby bottle, and a swaddle blanket. These items are artistically placed to create an impression of luxury and care, suggesting that the website offers a premium selection of products. Prominently displayed within the design is a bold, attractive advertisement for a 15\% off discount. This text is strategically positioned to catch the viewer's attention without overshadowing the curated display of goods. The text is written in soft, rounded font to maintain a gentle and friendly aesthetic. In one of the bottom corners, the website URL, 'www.yourgreatsite.com', is included in a clear font for easy readability. The overall effect of the design is comforting and welcoming, aiming to attract expecting parents to explore the website's offerings further. This Facebook cover is effectively tailored to appeal to the needs and desires of its target audience.} \\
        \hline
        \includegraphics[height=2.6cm]{fig/cole_benchmark_with_layers/ART_Posts_16.png}& \includegraphics[height=2.6cm]{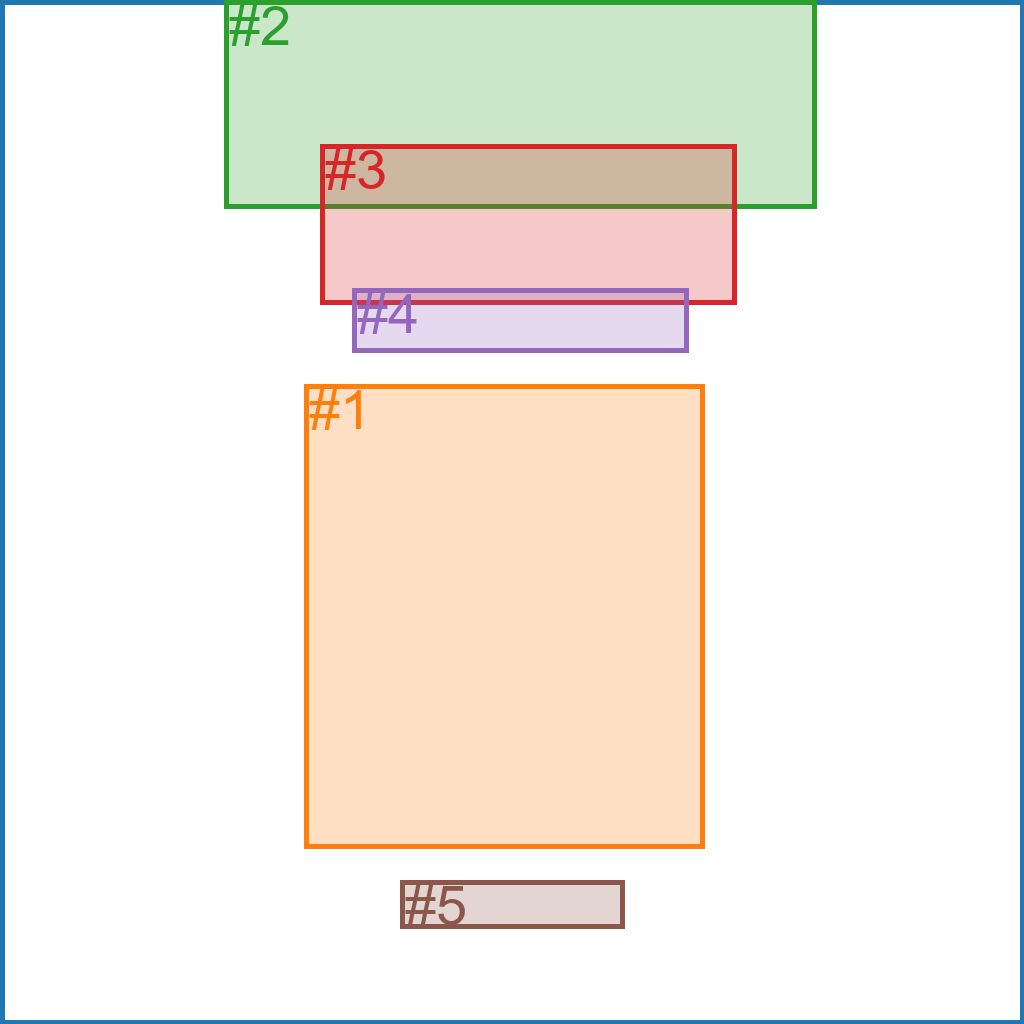} & {\footnotesize The image is designed as an Instagram story promoting a special Christmas offer for a chocolate drink. The background of the story features a cozy, festive theme with a warm and inviting color scheme, primarily consisting of rich browns and deep reds, reminiscent of hot chocolate and Christmas decor. Centered in the image is a steaming cup of chocolate drink, garnished with a sprinkle of cocoa powder and a cinnamon stick, suggesting warmth and indulgence. To enhance the festive atmosphere, there are elements such as small evergreen branches, a scattering of red berries, and a few decorative golden bells placed around the cup. The text on the story is bold and eye-catching, starting with 'Christmas Special' in elegant white script at the top. Below this, the details of the offer are highlighted in bright red, stating '20\% OFF' to capture attention. Further down, the call to action 'Order Now!' is displayed in bold white letters, encouraging viewers to take immediate advantage of the offer. The overall style of the image is cozy and appealing, designed to evoke a sense of the holiday spirit and entice customers to enjoy a delicious chocolate drink during the Christmas season. The aesthetic is suited to engage viewers on social media, making the offer both attractive and memorable.} \\
        \hline
        \includegraphics[height=2.6cm]{fig/real_benchmark/curtain_art.png}& \includegraphics[height=2.6cm]{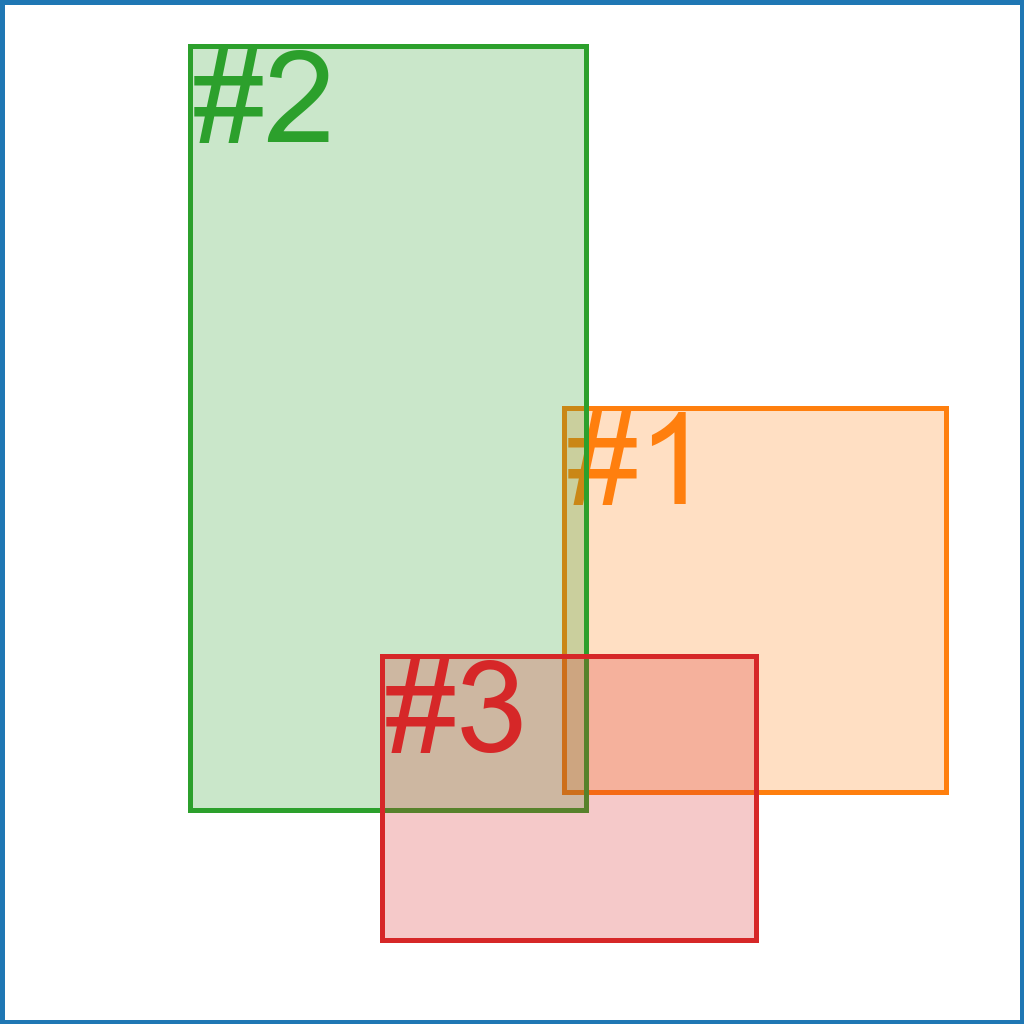} & { \footnotesize The image shows a collection of luggage items on a carpeted floor. There are three main pieces of luggage: a large suitcase, a smaller suitcase, and a duffel bag. The large suitcase is positioned in the center, with the smaller suitcase to its left and the duffel bag to its right. The luggage appears to be packed and ready for travel. In the foreground, there is a plastic bag containing what looks like a pair of shoes. The background features a white curtain, suggesting that the setting might be indoors, possibly a hotel room or a similar temporary accommodation. The image is in black and white, which gives it a timeless or classic feel.}  \\
        \hline
        \includegraphics[height=2.6cm]{fig/real_benchmark/table_art.png}& \includegraphics[height=2.6cm]{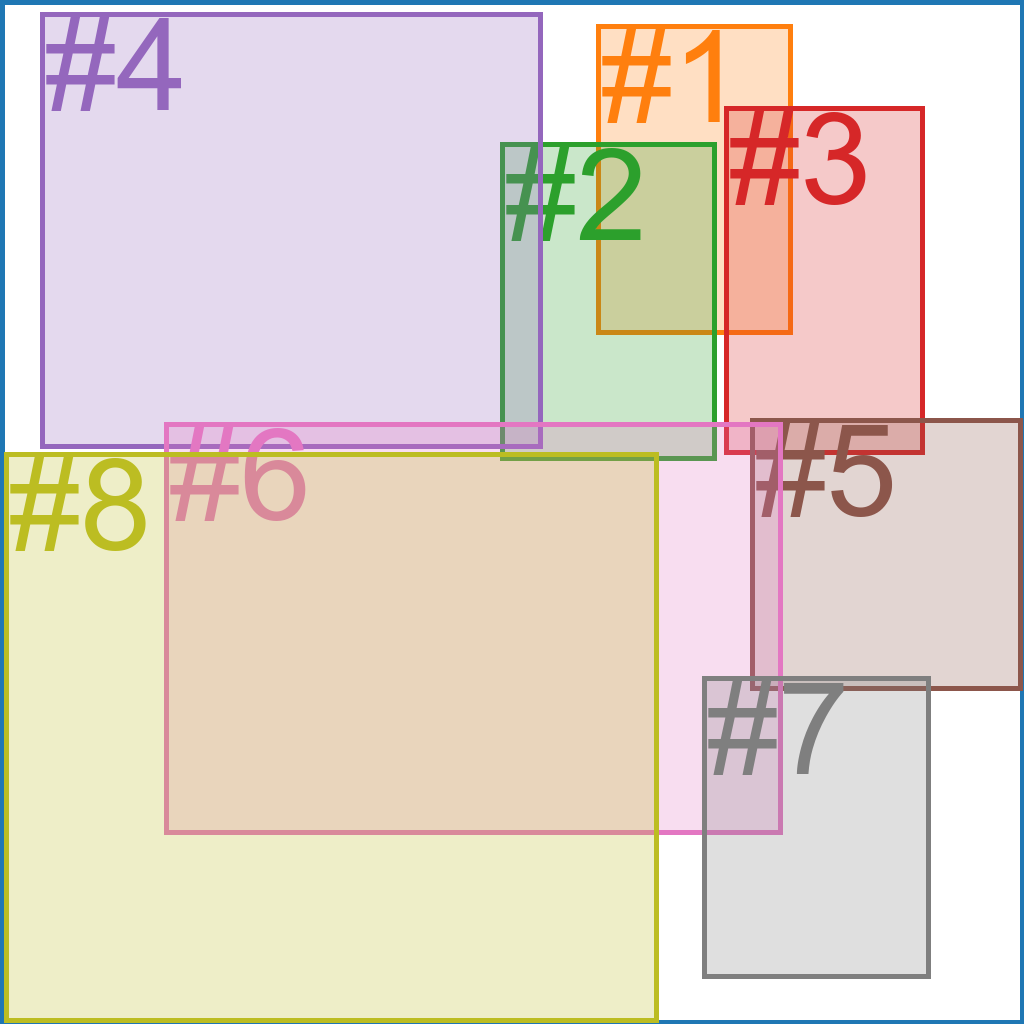} & {\footnotesize The image shows a rustic wooden table setting with a variety of items. On the table, there is a plate with six golden-brown, round, hollow pastries, which appear to be madeleines. To the left of the plate, there is a silver teapot with a wooden handle and spout. Next to the teapot, there are three glasses with different designs, filled with a clear liquid, possibly water. To the right of the plate, there is a small white plate with slices of yellow fruit, which could be pineapple. In the foreground, there is a green plant with broad leaves, and a silver spoon is placed on the table. The overall setting suggests a cozy, inviting atmosphere, possibly for a tea or dessert time.} \\ \hline
    \end{tabular}
    \vspace{-1mm}
    \caption{Detailed anonymous region layouts and global prompts for multi-layer image generation in Figure 6 of the main paper.}
    \label{tab:generation_results_2}
\end{table*}
\newpage

\begin{table*}[t]
\centering
\begin{minipage}{\textwidth}
    \centering
    \small
    \begin{tabular}{p{0.2\textwidth}|p{0.7\textwidth}}
    \textbf{Metrics} & \textbf{Detailed Instruction} \\
    \hline
    Aesthetics & Please evaluate the overall visual appeal of the images. Consider which method produces more visually pleasing and attractive results. Focus on the artistic quality, color harmony, and whether the style matches design aesthetics. \\
    \hline
    Typography & Please assess the text quality in the generated images. Check if the text is clear, readable and accurately rendered without distortions. Evaluate whether the font style, size and spacing are appropriate, and if the text matches the intended content. \\
    \hline
    Harmonization & Please examine the harmony of layers around the merged image. Consider whether the transitions between layers are smooth and natural, and if the layer effects enhance the overall visual quality without looking artificial. \\
    \hline
    Layout & Please evaluate the overall composition and arrangement. Check if text and graphic elements are well-balanced and properly aligned. Consider whether the spacing is appropriate, elements are organized logically, and if there are any awkward overlaps or conflicts between components. \\
    \end{tabular}
    \vspace{-1mm}
    \caption{Detailed Instructions for the User Study on the \textsc{DESIGN-MULTI-LAYER-BENCH}}
    \label{tab:user_study_prompt_1}
\end{minipage}

\vspace{5em}

\begin{minipage}{\textwidth}
    \centering
    \small
    \begin{tabular}{p{0.2\textwidth}|p{0.7\textwidth}}
    \textbf{Metrics} & \textbf{Description} \\
    \hline
    Aesthetics & Please evaluate the visual appeal of the generated images. Consider which result looks more visually pleasing and artistically satisfying. Focus on the overall aesthetic quality and visual attractiveness of the designs. \\
    \hline
    PromptFollow & Please assess how well each generated image matches the given text prompt. Compare the results and determine which method better captures and reflects the requirements specified in the prompt text. \\
    \hline
    Harmonization & Please examine the visual consistency and smoothness between different layers, particularly focusing on the transitions at the right and bottom edges. Consider whether the layer blending appears natural and well-integrated. \\
    \end{tabular}
    \vspace{-1mm}
    \caption{Detailed Instructions for the User Study on the \textsc{PHOTO-MULTI-LAYER-BENCH}}
    \label{tab:user_study_prompt_2}
\end{minipage}
\end{table*}

\onecolumn

\begin{minipage}[b]{1\columnwidth}
\begin{algorithm}[H]
\SetAlgoLined
\caption{Layout Conditional Multi-Layer 3D-RoPE}
\label{algo:3drope_1}
\tiny
\begin{lstlisting}[language=Python]
import torch

def get_1d_rotary_pos_embed(dim, pos, theta=10000.0):
    # dim: Dimension of the frequency tensor.
    # pos: Position indices for the frequency tensor. Shape: [S]
    # theta: Scaling factor for frequency computation.

    freqs = 1.0 / (theta ** (torch.arange(0, dim, 2)[:(dim // 2)] / dim))
    freqs = torch.outer(pos, freqs)
    freqs_cos = freqs.cos().repeat_interleave(2, dim=1)
      freqs_sin = freqs.sin().repeat_interleave(2, dim=1)
    
    return freqs_cos, freqs_sin

def get_3d_rotary_pos_embed(ids, axes_dim = (16, 56, 56)):
    # ids: 3D position indices of visual tokens. Shape: [S, 3]
    # axes_dim: RoPE dimensions for each axis.

    cos_out = []
    sin_out = []
    for i in range(3):
        cos, sin = get_1d_rotary_pos_embed(axes_dim[i], ids[:, i])
        cos_out.append(cos)
        sin_out.append(sin)
    freqs_cos = torch.cat(cos_out, dim=-1)
    freqs_sin = torch.cat(sin_out, dim=-1)

    return freqs_cos, freqs_sin
    
def prepare_latent_image_ids(height, width, list_layer_box):
    # height: The height of the image latent.
    # width: The width of the image latent.
    # list_layer_box: List of bounding boxes in each layer. 

    ids_list = []
    for layer_idx, layer_box in enumerate(list_layer_box):
        ids = torch.zeros(height//2, width//2, 3)
        ids[..., 0] = layer_idx # use the first axis to distinguish layers
        ids[..., 1] = ids[..., 1] + torch.arange(height//2)[:, None]
        ids[..., 2] = ids[..., 2] + torch.arange(width//2)[None, :]

        x1, y1, x2, y2 = layer_box
        ids = ids[y1:y2, x1:x2, :]
        ids = ids.reshape(-1, ids.shape[-1])
        ids_list.append(ids)
    latent_image_ids = torch.cat(ids_list, dim=0)

    return flatent_image_ids

\end{lstlisting}
\end{algorithm}
\end{minipage}

\twocolumn

\onecolumn

\begin{minipage}[b]{1\columnwidth}
\begin{algorithm}[H]
\SetAlgoLined
\caption{Layout Conditional Multi-Layer 3D-RoPE within Attention Module}
\label{algo:3drope_2}
\tiny
\begin{lstlisting}[language=Python]
import torch
import torch.nn as nn
import torch.nn.functional as F

def apply_rotary_pos_embed(x, freqs_cis):
    # x: Query or key tensor to apply rotary embeddings. Shape: [B, H, S, Dh]
    # freqs_cis: Precomputed frequency tensor for complex exponentials. Shape: [S, Dh]
    
    cos, sin = freqs_cis
    cos = cos[None, None]
    sin = sin[None, None]
    x_real, x_imag = x.reshape(*x.shape[:-1], -1, 2).unbind(-1)
    x_rotated = torch.stack([-x_imag, x_real], dim=-1).flatten(3)
    out = x.float() * cos + x_rotated.float() * sin

    return out

class AttentionProcessor(nn.module):
    to_q: nn.Linear
    to_k: nn.Linear
    to_v: nn.Linear
    to_out: nn.Linear
    def __call__(self, hidden_states, image_rotary_emb):
        # hidden_states: Input hidden states of the block. # [B, S, D]
        # image_rotary_emb: Precomputed 3D-RoPE frequency tensor. # [S, Dh]
        
        query = self.to_q(hidden_states)
        key = self.to_k(hidden_states)
        value = self.to_v(hidden_states)

        ...

        query = apply_rotary_pos_embed(query, image_rotary_emb)
        key = apply_rotary_pos_embed(key, image_rotary_emb)
        hidden_states = F.scaled_dot_product_attention(query, key, value, is_causal=False)
        hidden_state = self.to_out(hidden_states)
        
        ...

        return hidden_states
        

\end{lstlisting}
\end{algorithm}
\end{minipage}

\begin{table}[htbp]
    \centering
    \begin{tabular}{p{\textwidth}}
    \toprule
    \textbf{Prompt:} \small{The image is a graphic design with a celebratory theme. At the top, there is a banner with the text "Happy Anniversary" in a bold, sans-serif font. Below this banner, there is a circular frame containing a photograph of a couple. The man has short, dark hair and is wearing a light-colored sweater, while the woman has long blonde hair and is also wearing a light-colored sweater. They are both smiling and appear to be embracing each other. Surrounding the circular frame are decorative elements such as pink flowers and green leaves, which add a festive touch to the design. Below the circular frame, there is a text that reads "Isabel \& Morgan" in a cursive, elegant font, suggesting that the couple's names are Isabel and Morgan. At the bottom of the image, there is a banner with a message that says "Happy Anniversary! Cheers to another year of love, laughter, and cherished memories together." This text is in a smaller, sans-serif font and is placed against a solid background, providing a clear message of celebration and well-wishes for the couple. The overall style of the image is warm and celebratory, with a color scheme that includes shades of pink, green, and white, which contribute to a joyful and romantic atmosphere.}
    \vspace{1em}
    \end{tabular}
    \begin{tabular}{ccc}
        \includegraphics[width=0.25\textwidth]{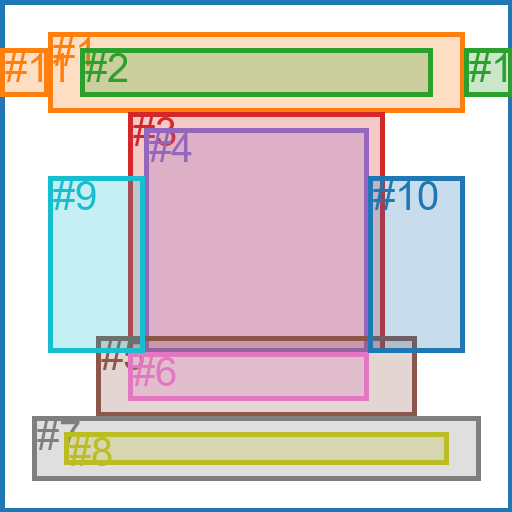} &
        \includegraphics[width=0.25\textwidth]{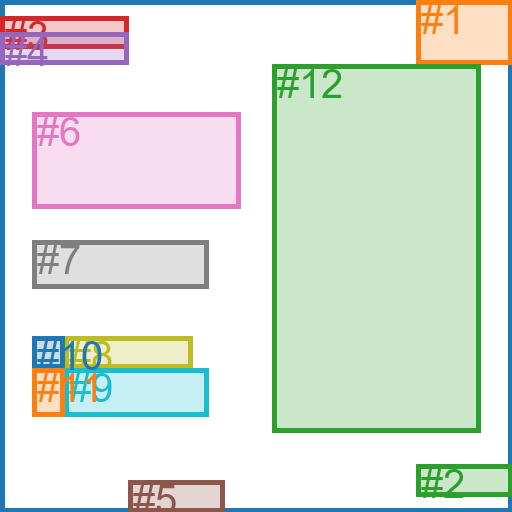} &
        \includegraphics[width=0.25\textwidth]{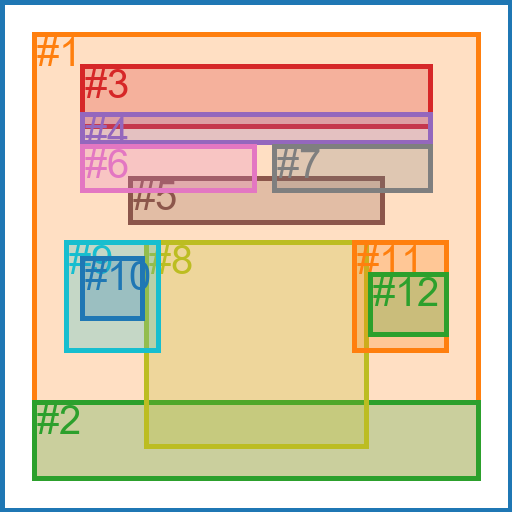} \\
        \small{Layout A} & \small{Layout B} & \small{Layout C} \\[1em]
        \includegraphics[width=0.3\textwidth]{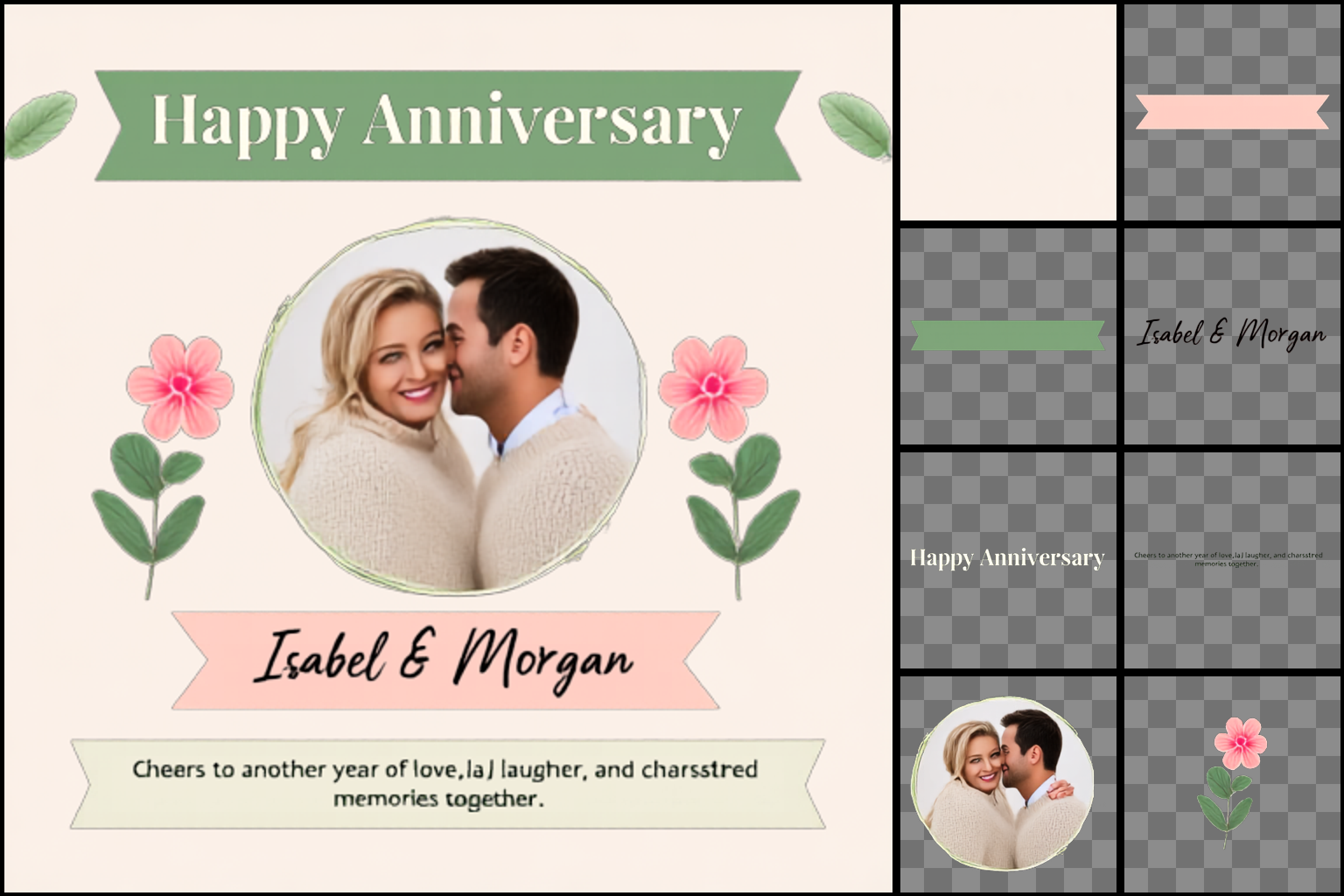} &
        \includegraphics[width=0.3\textwidth]{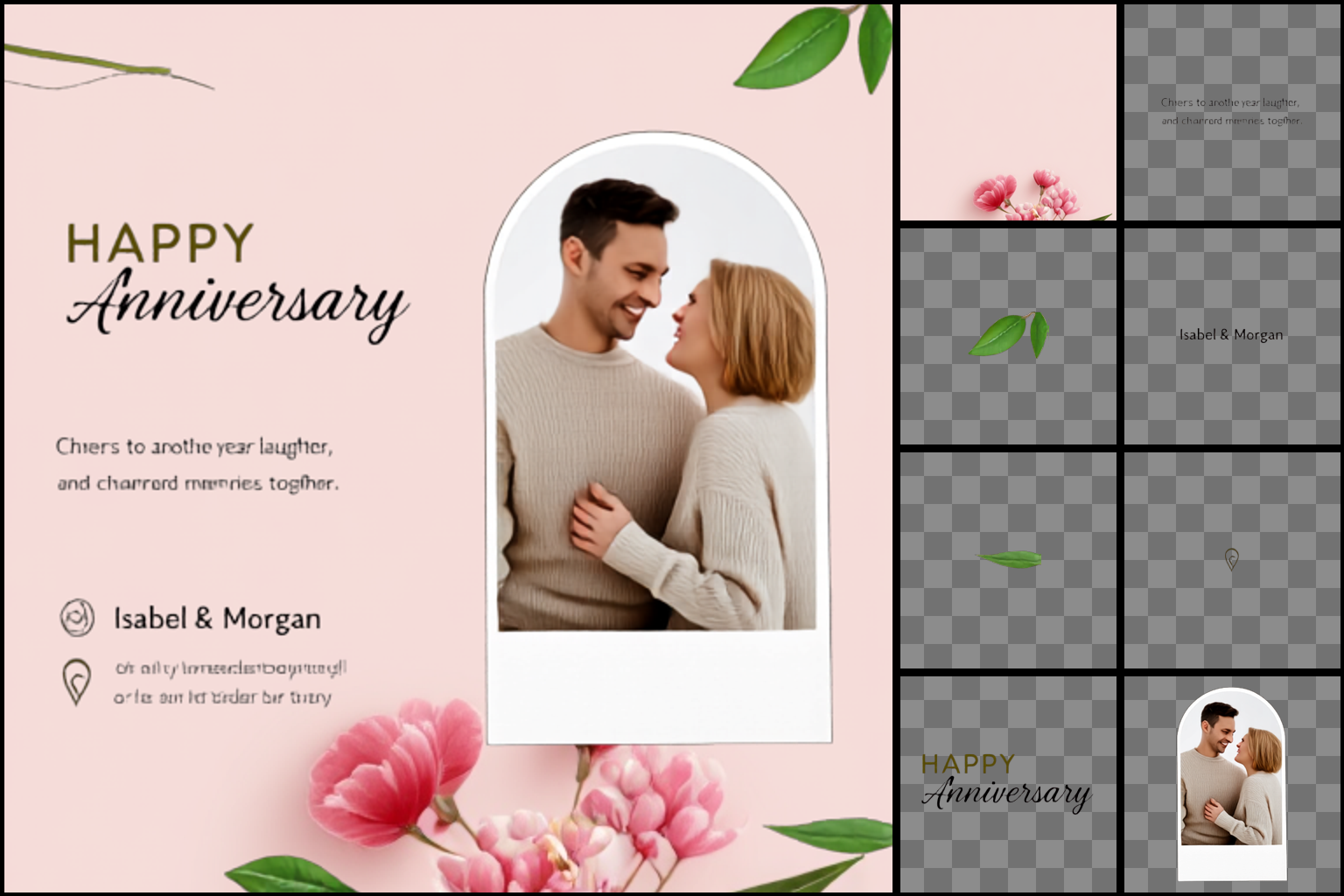} &
        \includegraphics[width=0.3\textwidth]{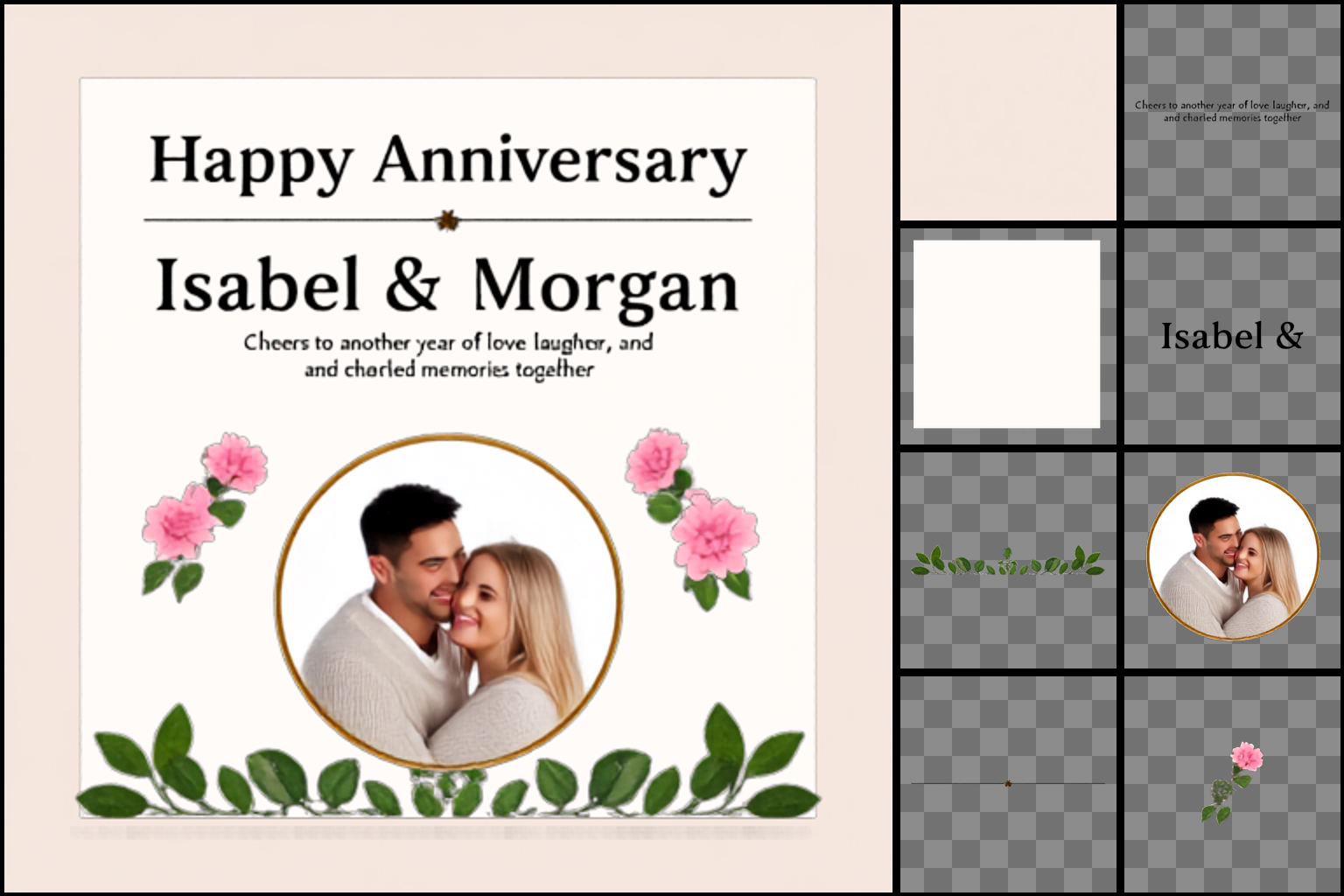} \\
        \small{Generated A} & \small{Generated B} & \small{Generated C} \\
        \bottomrule
    \end{tabular}

    \caption{Generated results conditioned on the same prompt and variant layouts. We show the prompt at the first row, three different layouts (the background index `\#0' is omitted) at the second row and the generated results at the last row. (Case 1)}
    \label{tab:variant_layout1}
\end{table}

\begin{table}[htbp]
    \centering
    \begin{tabular}{p{\textwidth}}
    \midrule
    \textbf{Prompt:} \small{The image is a promotional graphic for a new collection that is coming soon in February 20xx. The central focus of the image is a collection of items that suggest a theme of natural beauty and freshness. There are two bottles of what appears to be a yellow-colored liquid, possibly a fragrance or essential oil, given their shape and the context. The bottles are placed on a white, oval-shaped surface that resembles a soap or a decorative plate. Surrounding the bottles are slices of lemon, which are scattered around the surface, adding a citrus element to the composition. There are also green leaves, possibly basil, which are placed near the lemon slices, contributing to the natural and fresh theme. The background is a solid, warm yellow color that complements the overall color scheme of the image. At the top of the image, there is text that reads "Our new collection is COMING SOON FEBRUARY 20xx," indicating the time frame for the release of the new collection. At the bottom, the text "Lime Basil" is visible, which likely refers to the scent or flavor of the items in the collection. The overall style of the image is clean, modern, and designe5d to evoke a sense of anticipation for the new collection.}
    \vspace{1em}
    \end{tabular}
    \begin{tabular}{ccc}
        \includegraphics[width=0.25\textwidth]{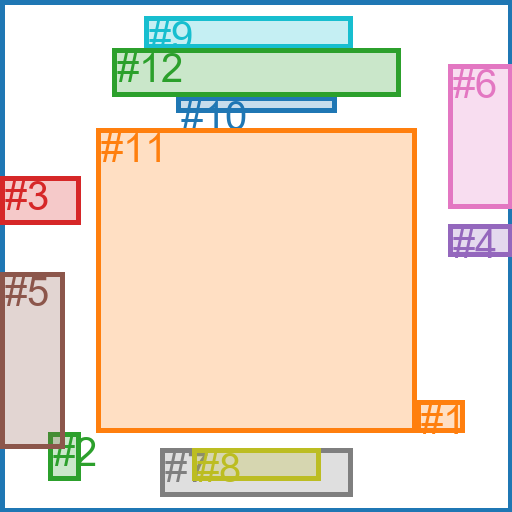} &
        \includegraphics[width=0.25\textwidth]{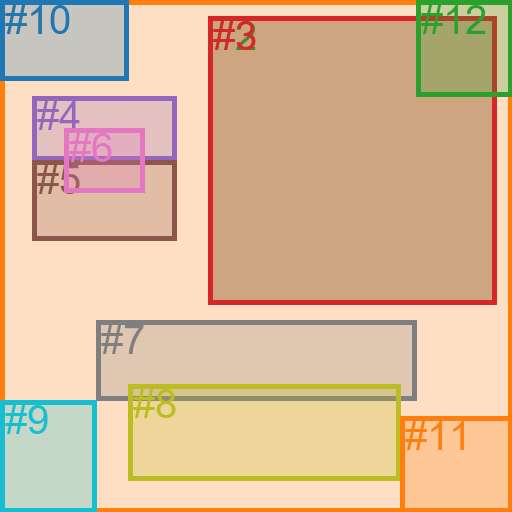} &
        \includegraphics[width=0.25\textwidth]{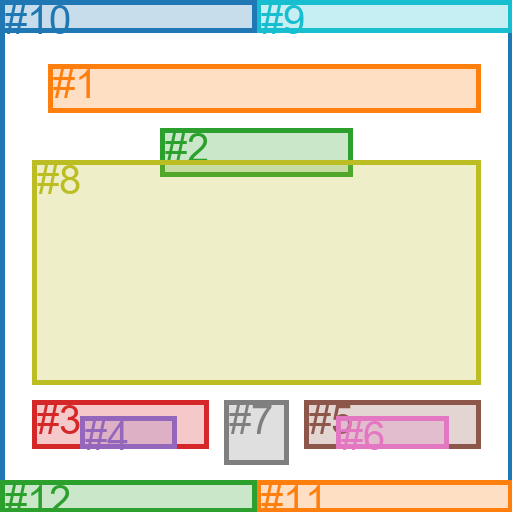} \\
        \small{Layout A} & \small{Layout B} & \small{Layout C} \\[1em]
        \includegraphics[width=0.3\textwidth]{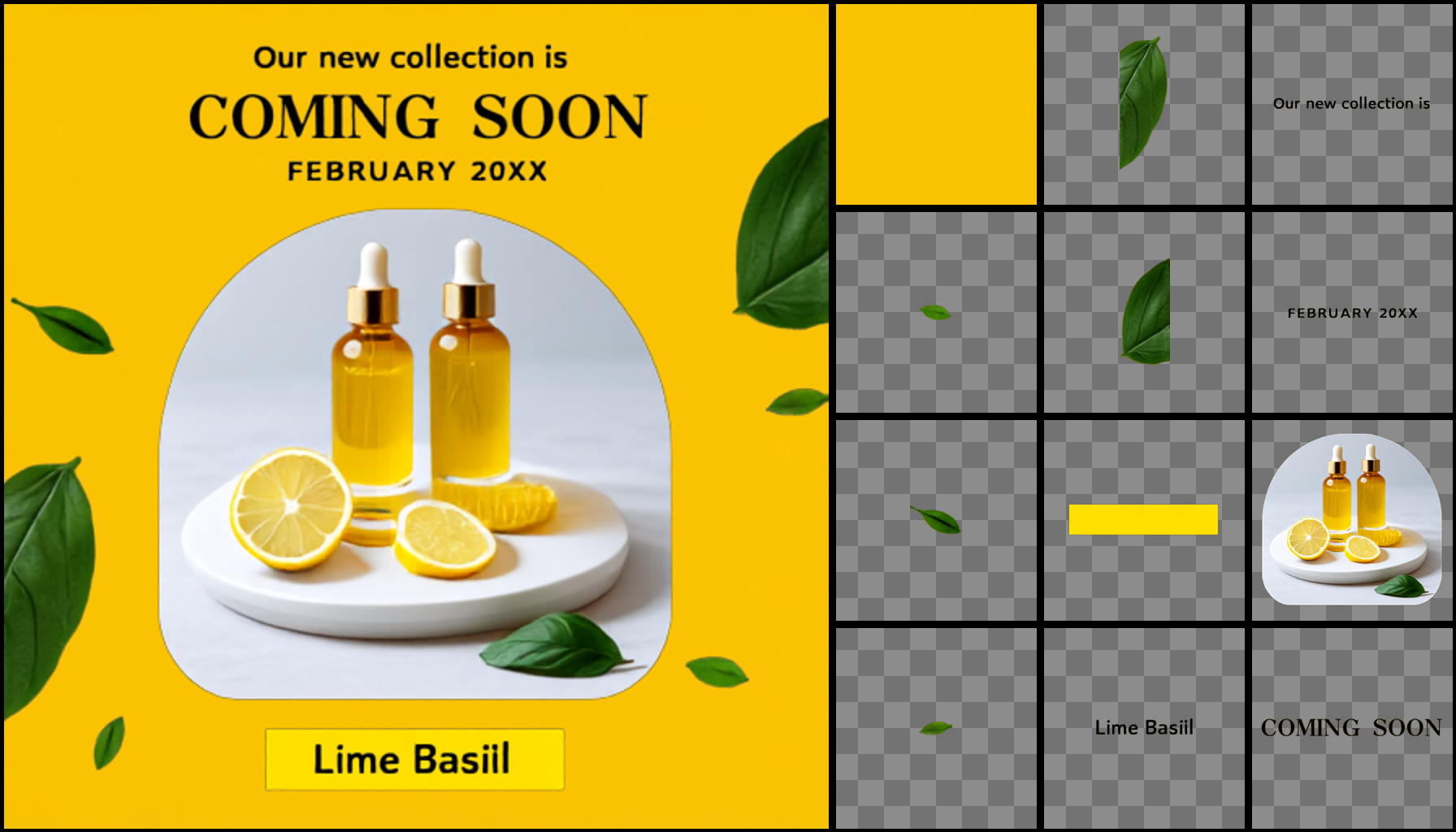} &
        \includegraphics[width=0.3\textwidth]{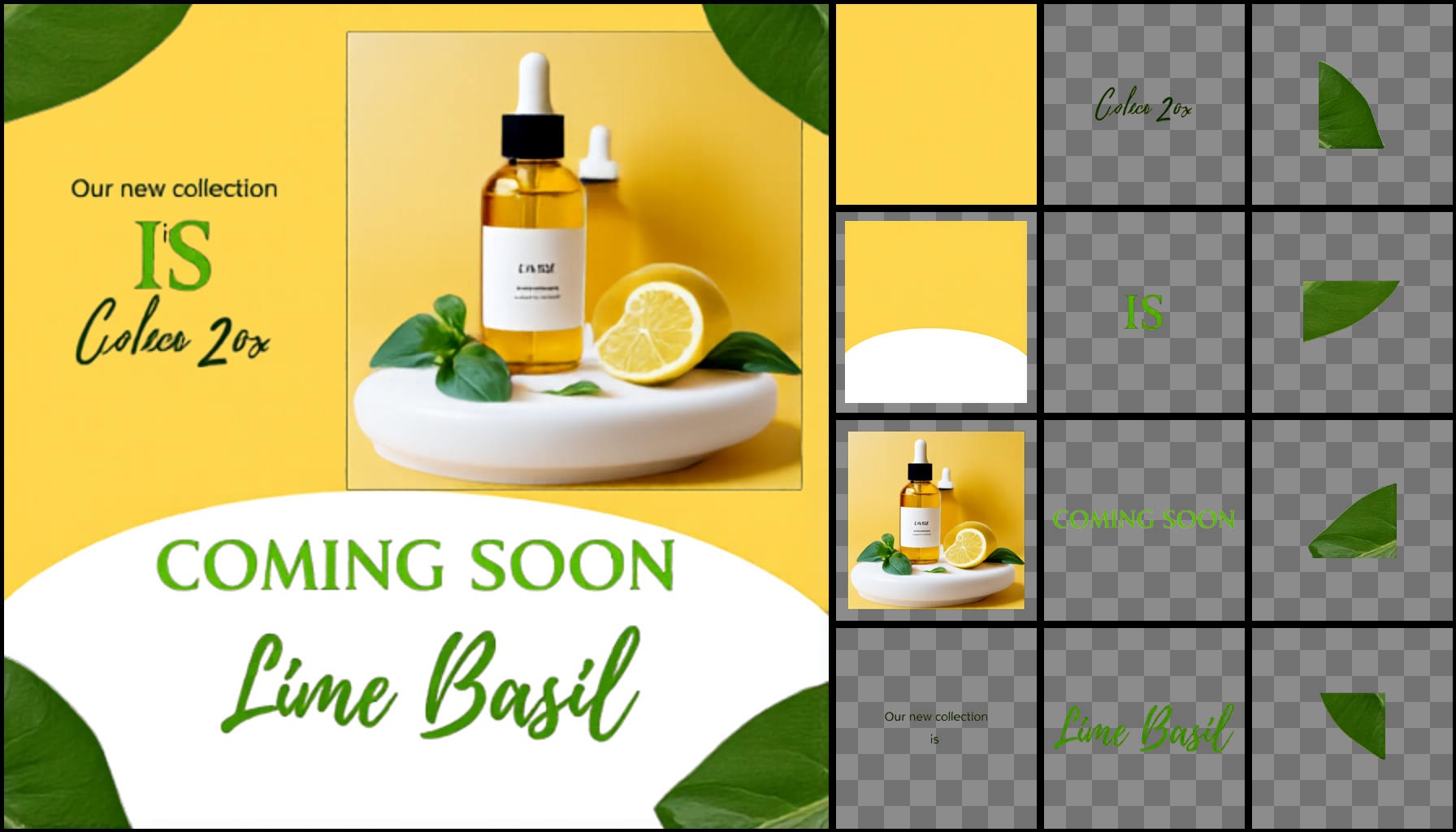} &
        \includegraphics[width=0.3\textwidth]{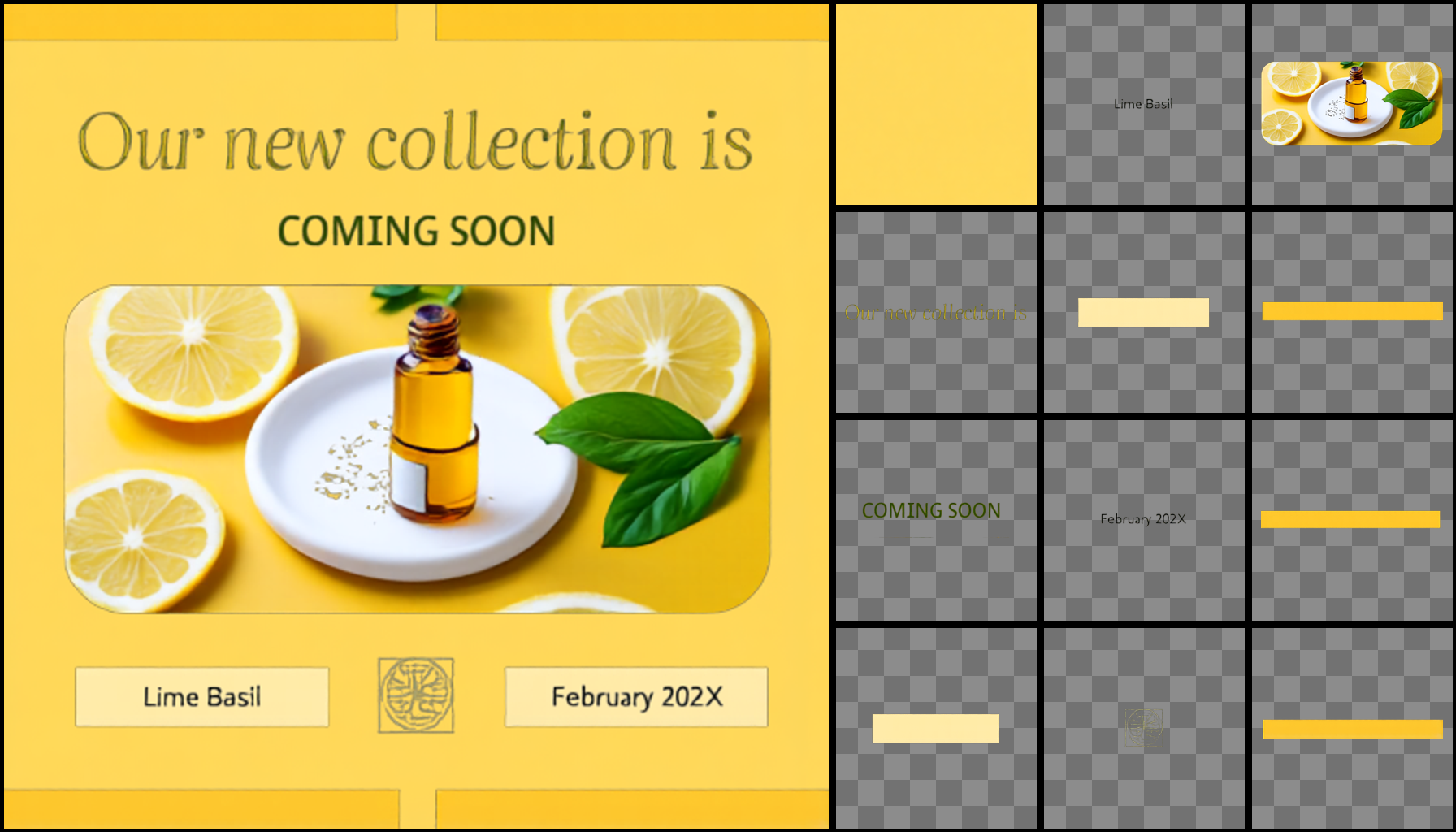} \\
        \small{Generated A} & \small{Generated B} & \small{Generated C} \\
        \bottomrule
    \end{tabular}
    
    \caption{Generated results conditioned on the same prompt and variant layouts. We show the prompt at the first row, three different layouts (the background index `\#0' is omitted) at the second row and the generated results at the last row. (Case 2)}
    \label{tab:variant_layout2}
\end{table}

\begin{table}[htbp]
    \centering
    \begin{tabular}{p{\textwidth}}
    \midrule
    \textbf{Prompt:} \small{The image features a stylized graphic of a carpentry home project. At the center, there is a three-dimensional illustration of a wooden house with a visible interior. The house is filled with various carpentry tools and materials, such as a ladder, a hammer, a saw, a measuring tape, a paint roller, and a paint tray. These items are arranged to suggest that they are being used for a home renovation or construction project. The background of the image is a dark green color, and there are two yellow diamonds on either side of the house, each containing the text "50\% OFF." This suggests that there is a discount offer associated with the carpentry home project. At the bottom of the image, there is a bold text that reads "CARPENTRY HOME PROJECT" in capital letters, indicating the theme of the image. Below this main title, there is a tagline that says "Dreams into reality with our expert guides," which implies that the image is likely an advertisement or promotional material for a service or product related to carpentry and home projects. The overall style of the image is clean and modern, with a clear focus on the carpentry theme and the promotional offer. The use of bright colors and bold text is designed to attract attention and convey the message of the advertisement effectively.}
    \vspace{1em}
    \end{tabular}
    \begin{tabular}{ccc}
        \includegraphics[width=0.25\textwidth]{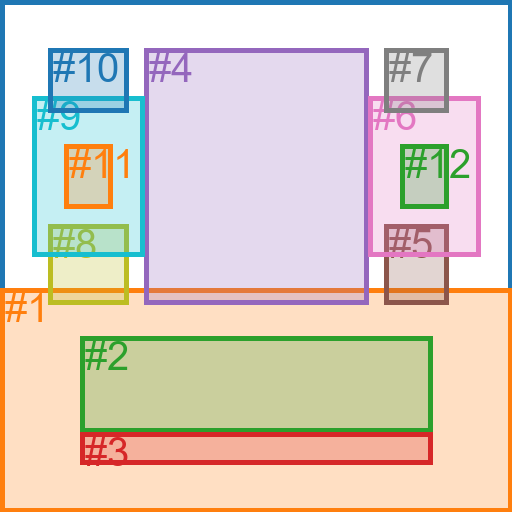} &
        \includegraphics[width=0.25\textwidth]{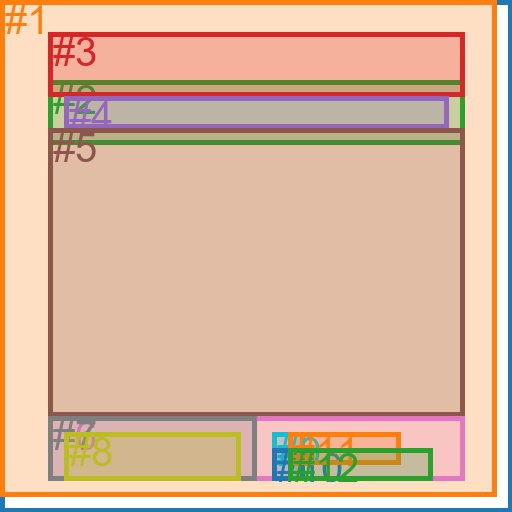} &
        \includegraphics[width=0.25\textwidth]{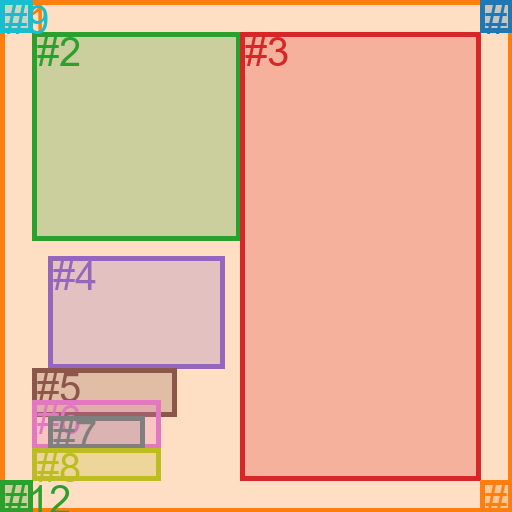} \\
        \small{Layout A} & \small{Layout B} & \small{Layout C} \\[1em]
        \includegraphics[width=0.3\textwidth]{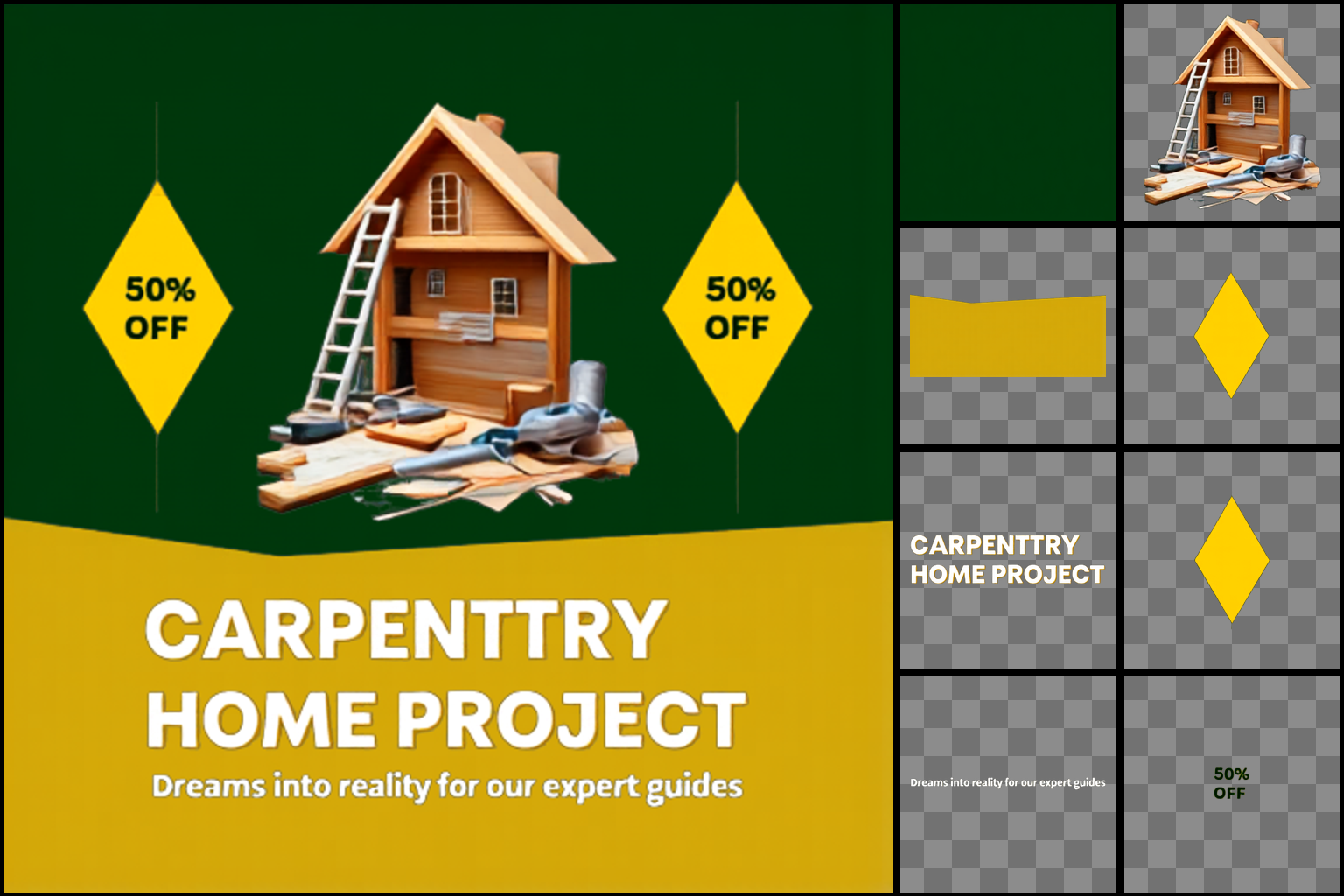} &
        \includegraphics[width=0.3\textwidth]{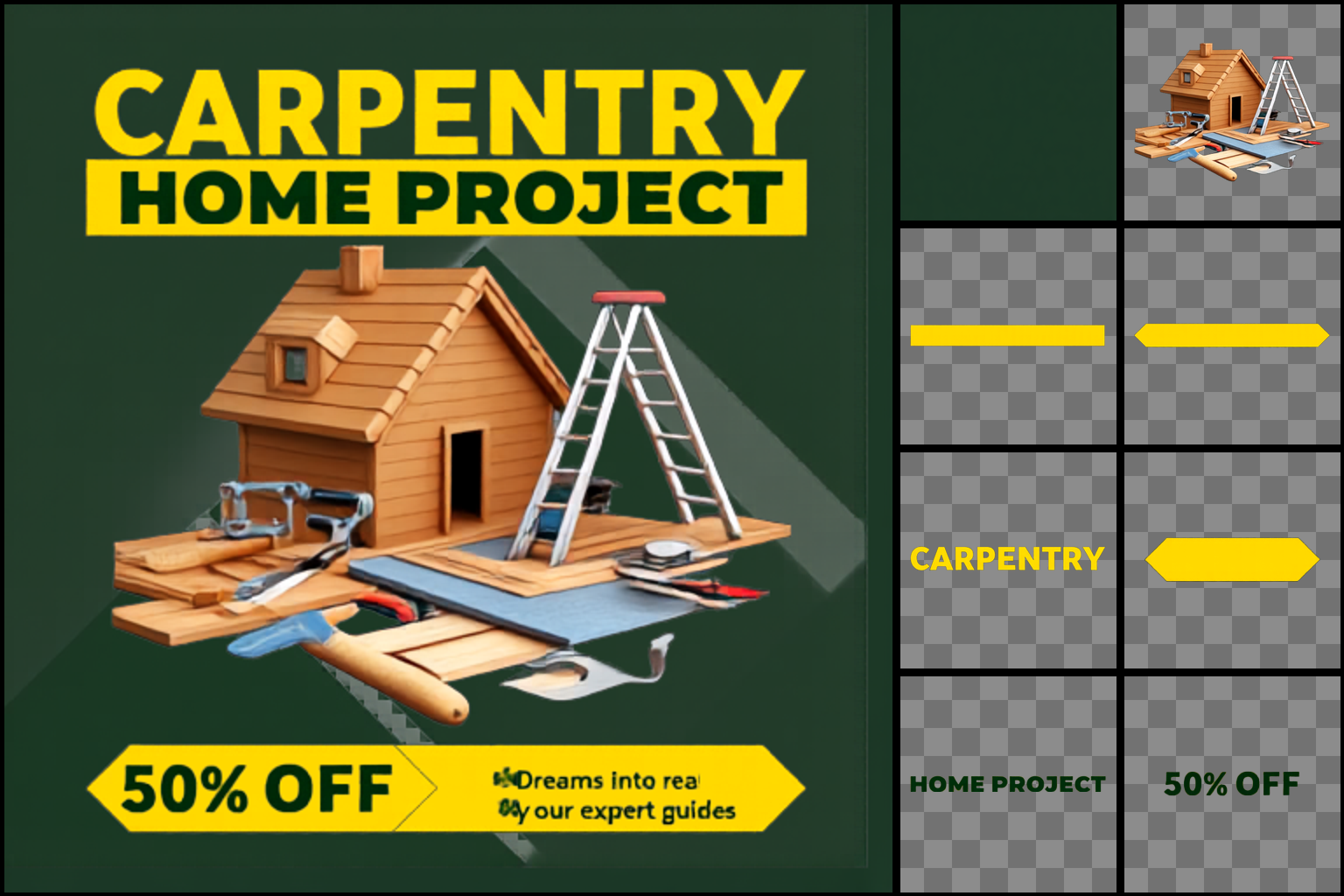} &
        \includegraphics[width=0.3\textwidth]{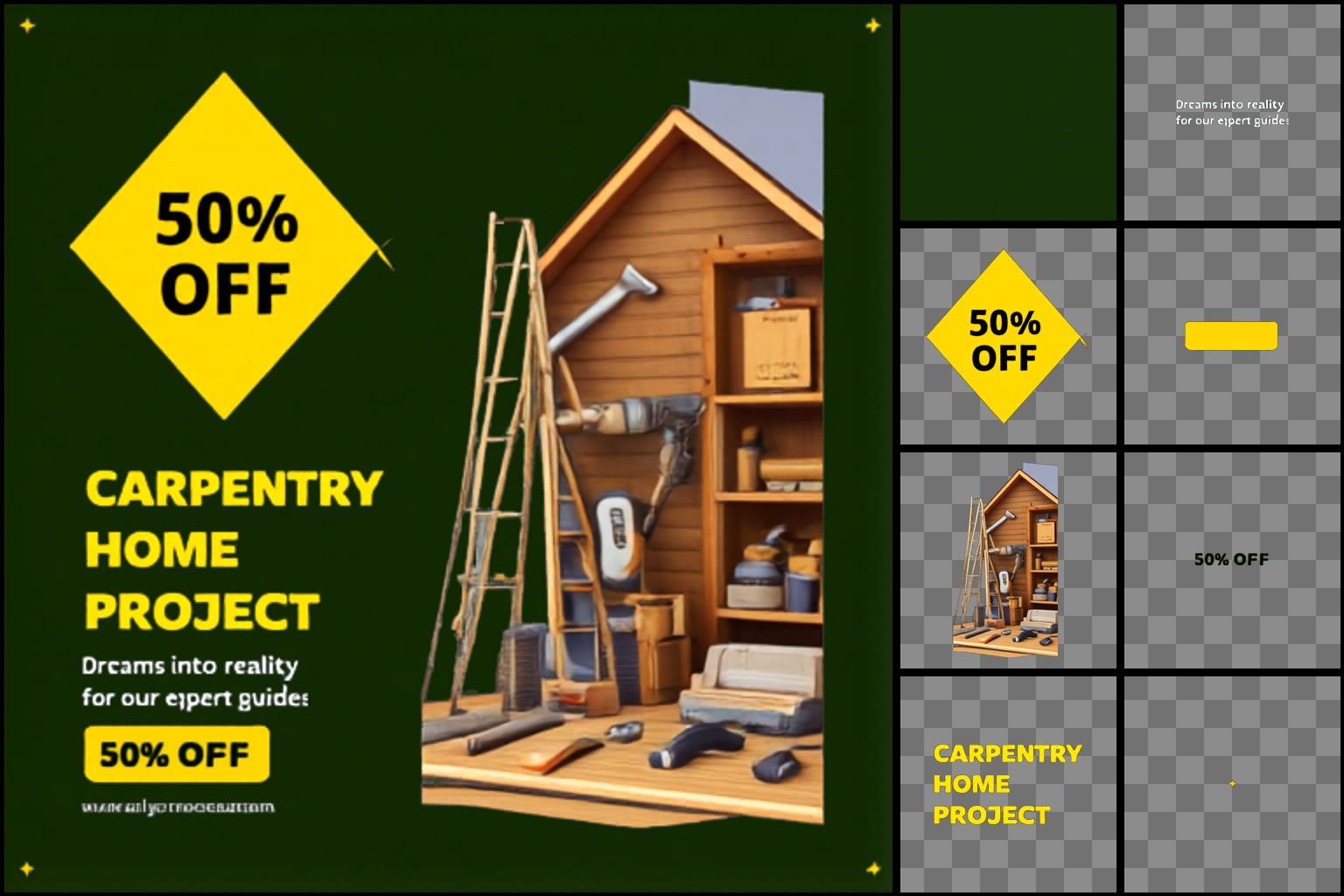} \\
        \small{Generated A} & \small{Generated B} & \small{Generated C} \\
        \bottomrule
    \end{tabular}
    
    \caption{Generated results conditioned on the same prompt and variant layouts. We show the prompt at the first row, three different layouts (the background index `\#0' is omitted) at the second row and the generated results at the last row. (Case 3)}
    \label{tab:variant_layout3}
\end{table}

\begin{table}[htbp]
    \centering
    \begin{tabular}{p{\textwidth}}
    \midrule
    \textbf{Prompt:} \small{The image features a graphic design with a stylized illustration of an urban landscape. The illustration includes various buildings of different shapes and sizes, some with red roofs, and a few trees. The buildings are depicted in a simplified manner, with flat colors and minimal detail, giving the image a modern and clean aesthetic. At the top of the image, there is text that reads "Urban Vision Architects" in bold, capital letters. Below this, in a smaller font, it says "Innovative architectural solutions." To the right of the text, there is a graphic element resembling a star or a sun with rays emanating from it. In the lower left corner, there is a discount offer indicated by the text "15\% OFF" in a bold, sans-serif font. The overall style of the image suggests it could be an advertisement or promotional material for an architectural firm. The color palette is limited, with a dominant beige background that contrasts with the red and black elements of the illustration and text.}
    \vspace{1em}
    \end{tabular}
    \begin{tabular}{ccc}
        \includegraphics[width=0.25\textwidth]{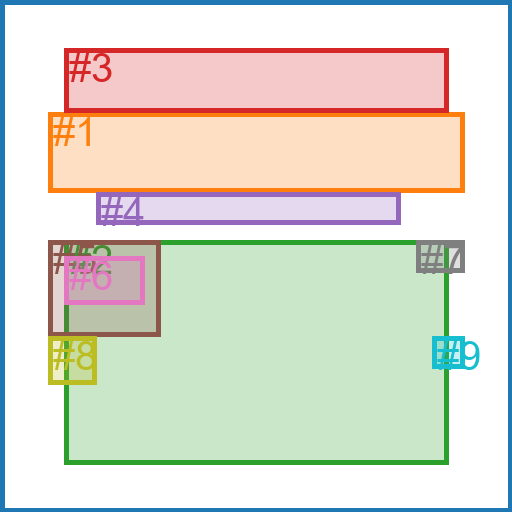} &
        \includegraphics[width=0.25\textwidth]{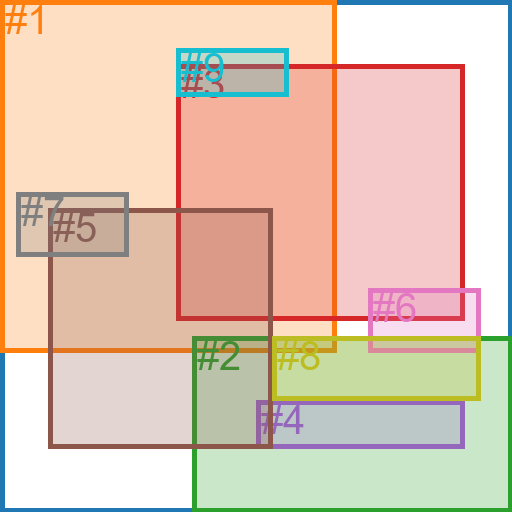} &
        \includegraphics[width=0.25\textwidth]{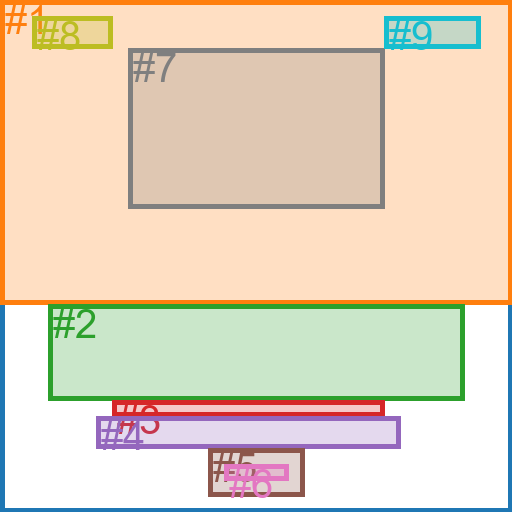} \\
        \small{Layout A} & \small{Layout B} & \small{Layout C} \\[1em]
        \includegraphics[width=0.3\textwidth]{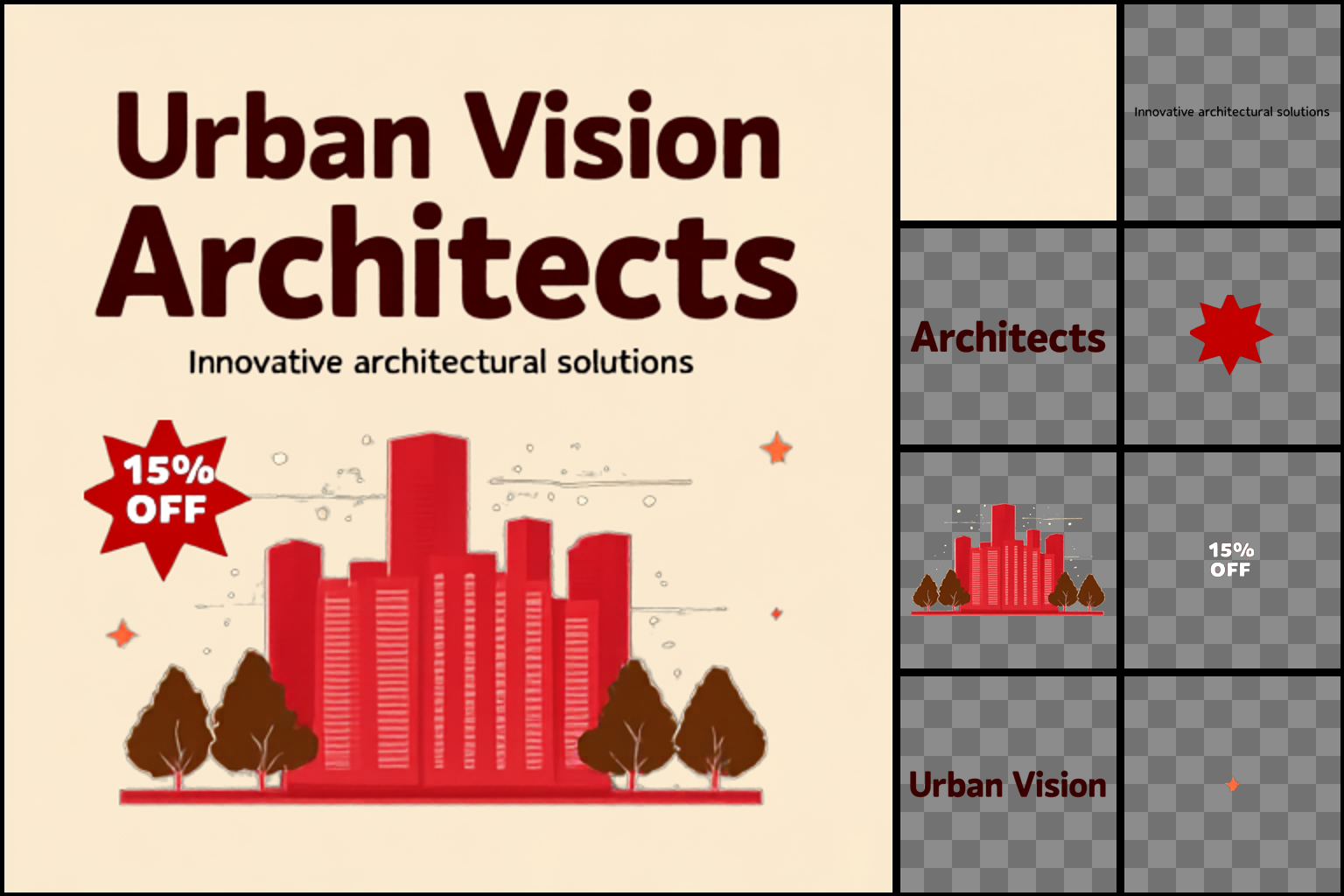} &
        \includegraphics[width=0.3\textwidth]{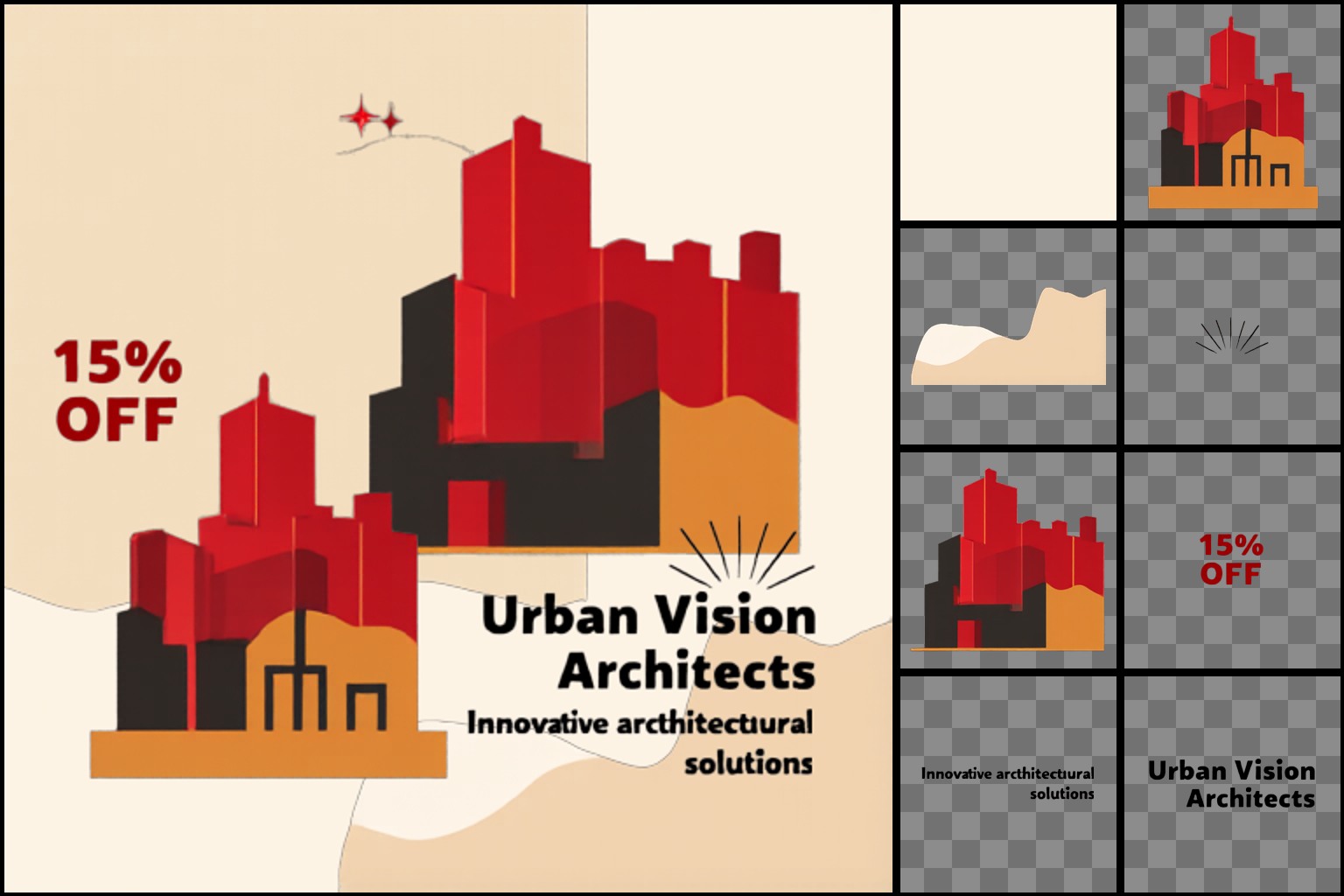} &
        \includegraphics[width=0.3\textwidth]{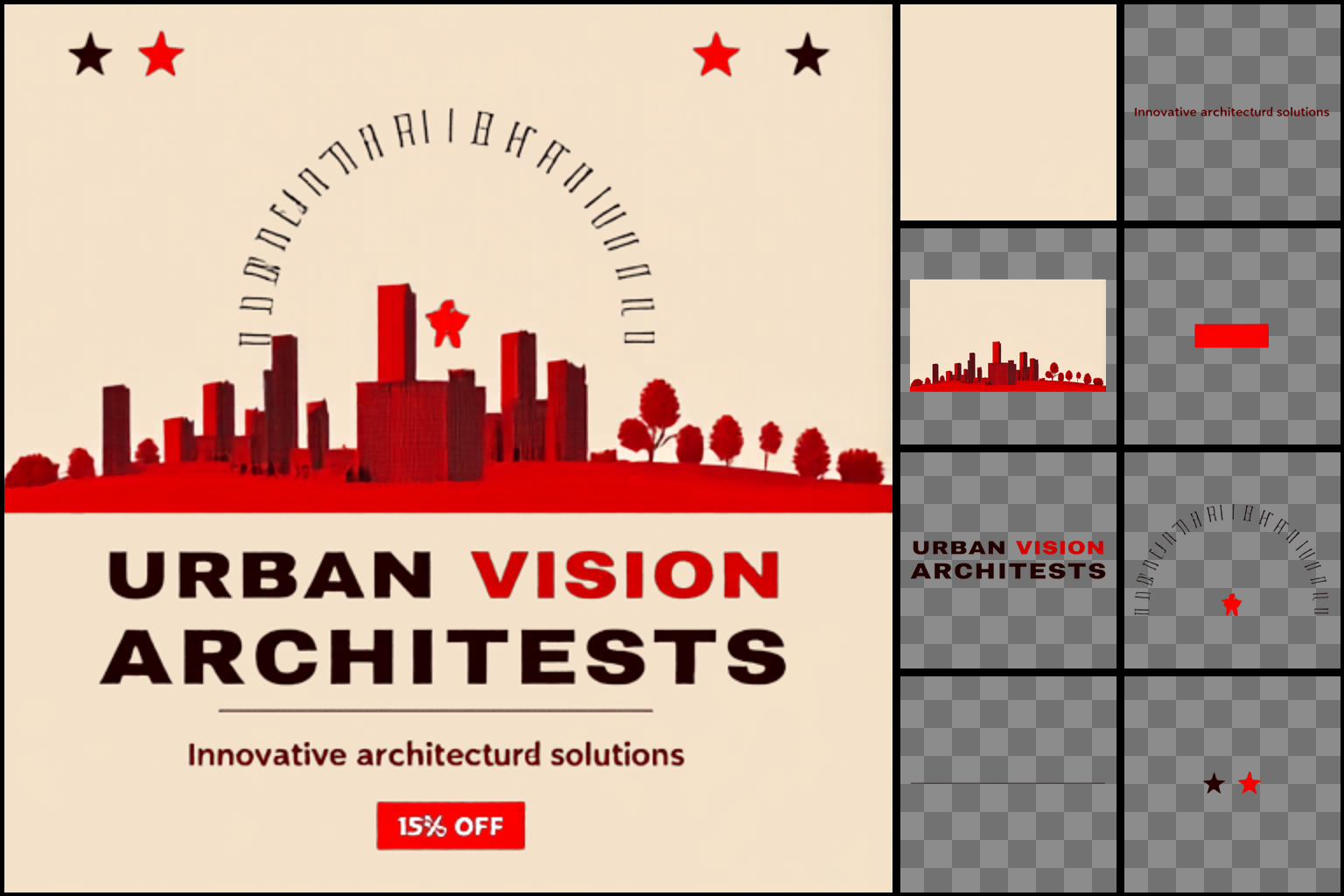} \\
        \small{Generated A} & \small{Generated B} & \small{Generated C} \\
        \bottomrule
    \end{tabular}
    
    \caption{Generated results conditioned on the same prompt and variant layouts. We show the prompt at the first row, three different layouts (the background index `\#0' is omitted) at the second row and the generated results at the last row. (Case 4)}
    \label{tab:variant_layout4}
\end{table}

\begin{table}[htbp]
    \centering
    \begin{tabular}{p{\textwidth}}
    \midrule
    \textbf{Prompt:} \small{The image features a collection of lipsticks. There are five lipsticks in total, each with a different color. From left to right, the first lipstick is a light pink, the second is a darker pink, the third is a bright red, the fourth is a deep red, and the fifth is a deep purple. Each lipstick is encased in a silver tube with a clear cap, allowing the color to be visible. The lipsticks are arranged in a straight line, and the background is a neutral beige color. At the top of the image, there is text that reads "NEW PRODUCT LIPSTICK COLLECTION," and at the bottom, there is a promotional message that says "SAVE UP TO 30\% SHOP NOW." The overall style of the image is promotional and designed to attract customers to the new lipstick collection.}
    \vspace{1em}
    \end{tabular}
    \begin{tabular}{ccc}
        \includegraphics[width=0.25\textwidth]{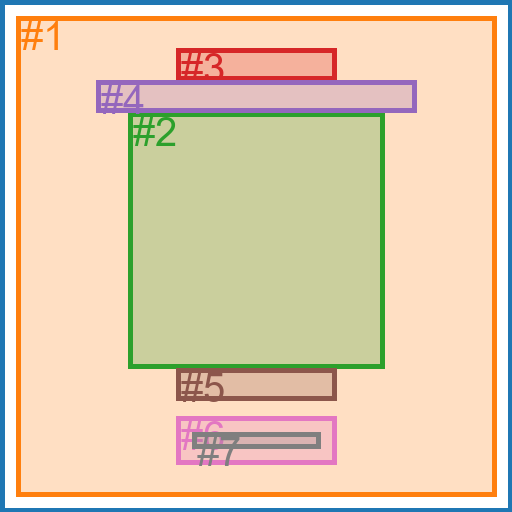} &
        \includegraphics[width=0.25\textwidth]{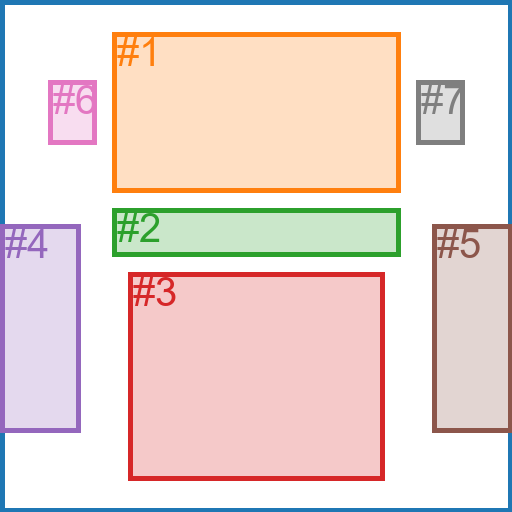} &
        \includegraphics[width=0.25\textwidth]{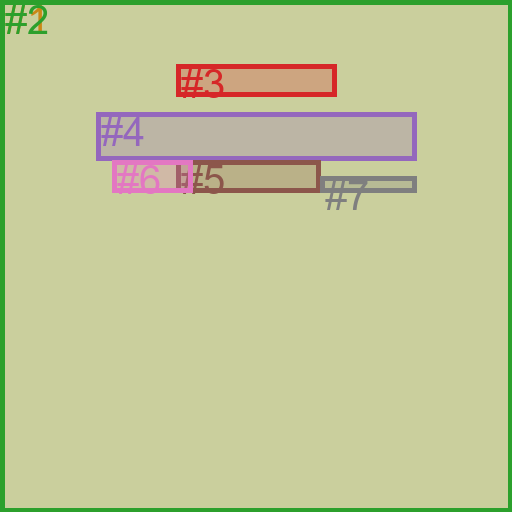} \\
        \small{Layout A} & \small{Layout B} & \small{Layout C} \\[1em]
        \includegraphics[width=0.3\textwidth]{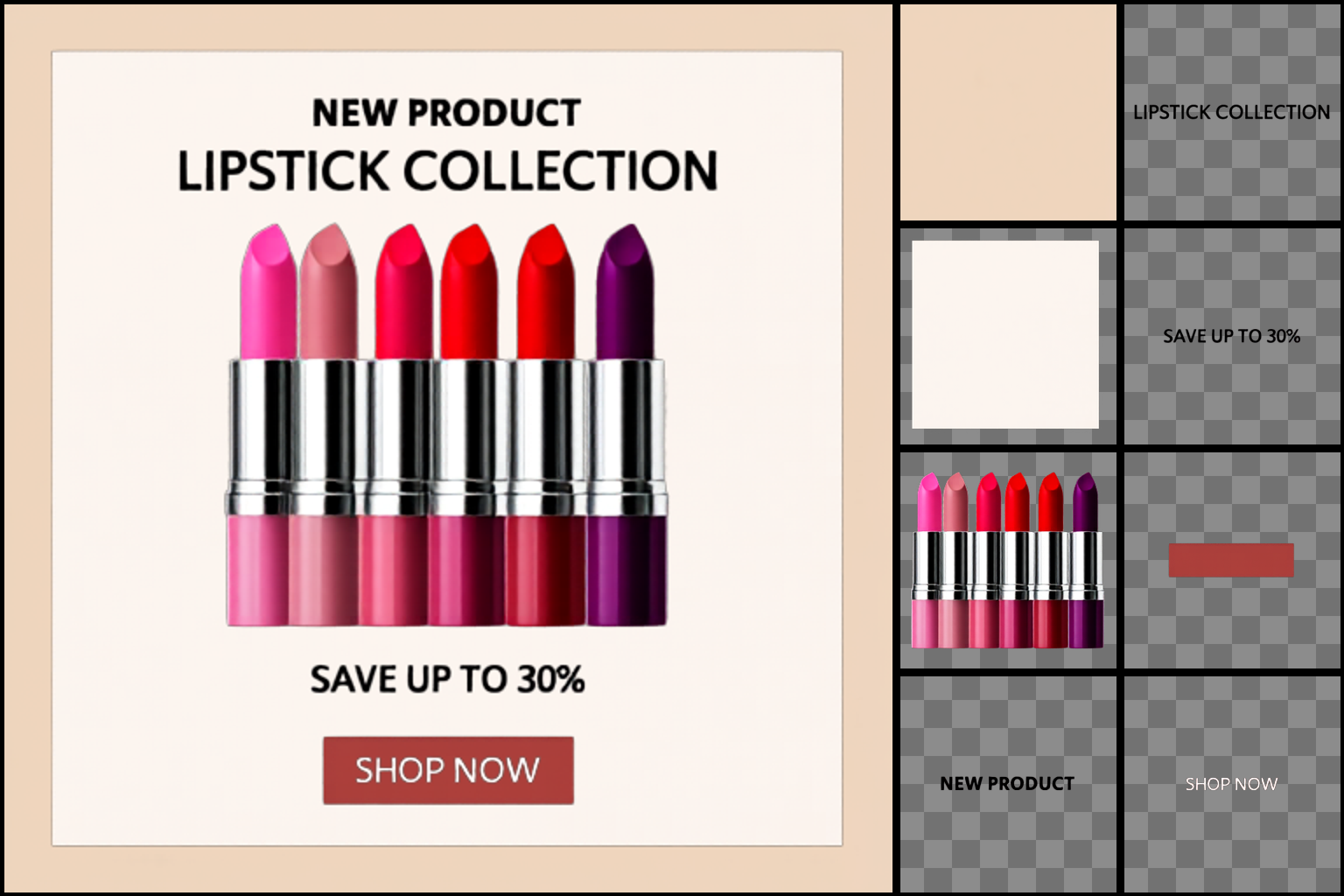} &
        \includegraphics[width=0.3\textwidth]{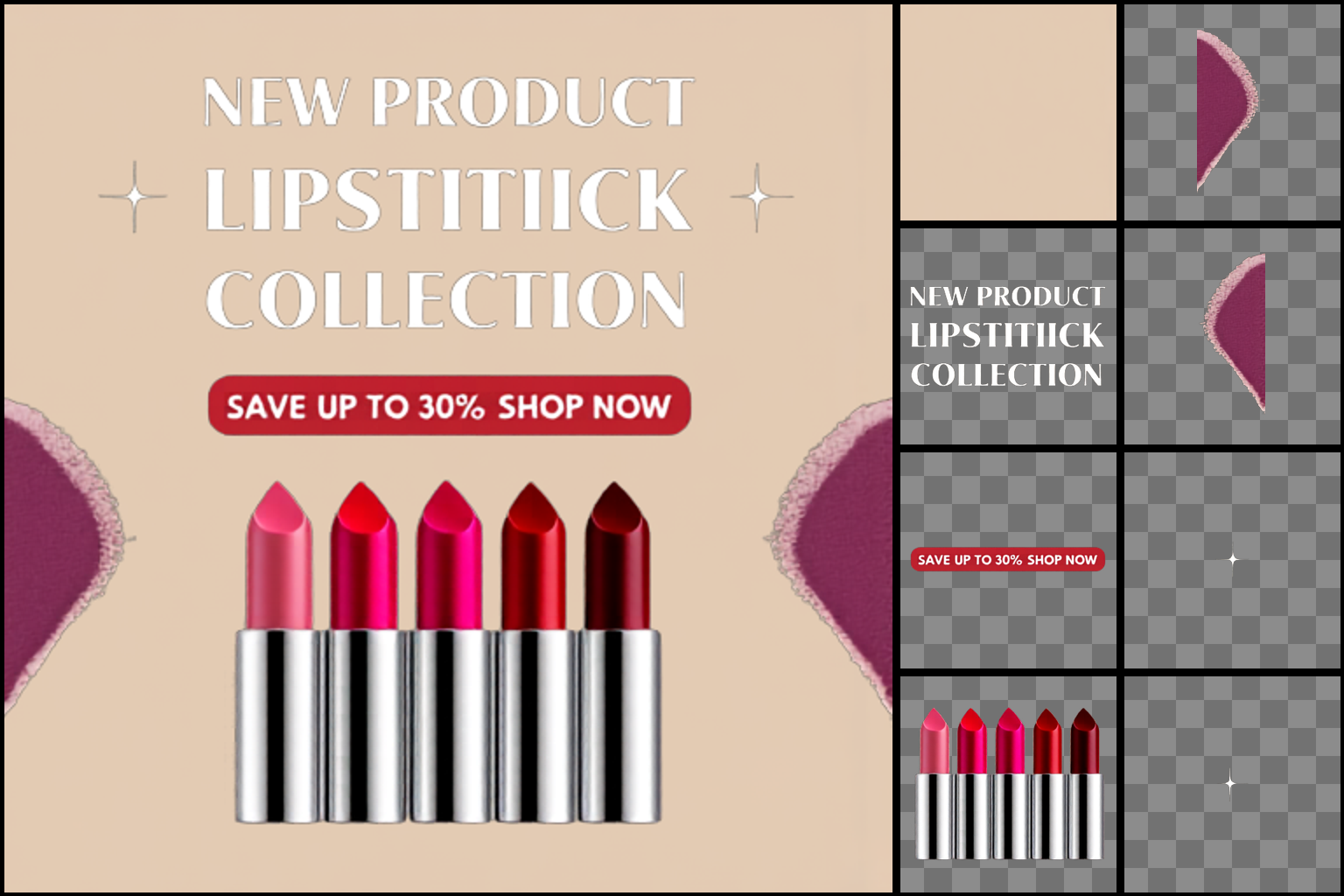} &
        \includegraphics[width=0.3\textwidth]{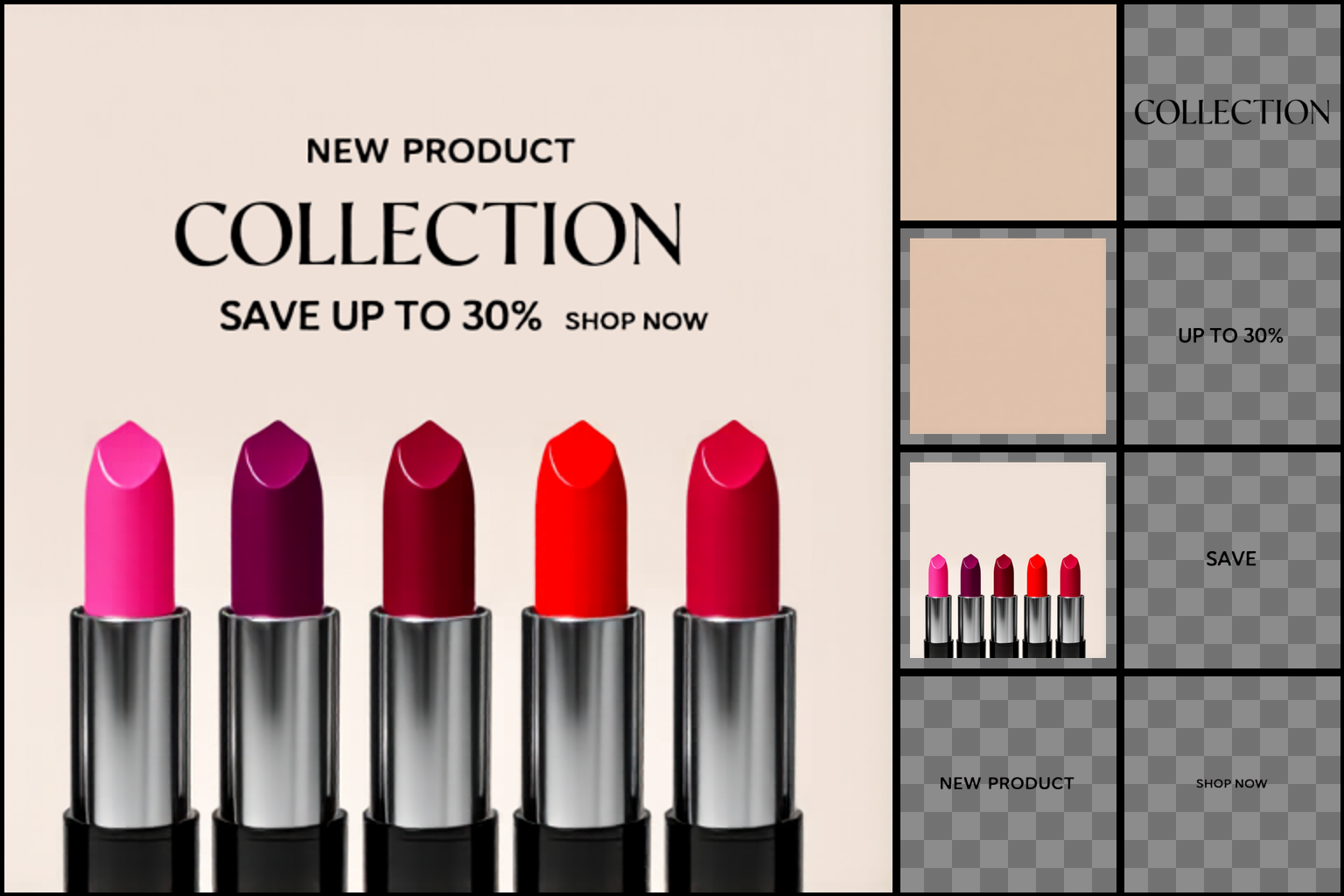} \\
        \small{Generated A} & \small{Generated B} & \small{Generated C} \\
        \bottomrule
    \end{tabular}
    
    \caption{Generated results conditioned on the same prompt and variant layouts. We show the prompt at the first row, three different layouts (the background index `\#0' is omitted) at the second row and the generated results at the last row. (Case 5)}
    \label{tab:variant_layout5}
\end{table}

\begin{table}[htbp]
    \centering
    \begin{tabular}{p{\textwidth}}
    \midrule
    \textbf{Prompt:} \small{The image features a stylized illustration of a person in a martial arts pose. The individual is depicted in a dynamic stance with one leg extended straight out to the side, while the other leg is bent at the knee, supporting the body. The person is wearing a white martial arts uniform, commonly known as a gi, and a black belt, which signifies a high level of proficiency in the martial art. The belt is tied around the waist, and the person's hands are clenched into fists, suggesting a state of readiness or combat. Above the illustration, there is text that reads "BLACK BELT CLUB" in bold, capital letters, indicating the name of the organization or program being advertised. Below this, there is a slogan that says "Elevate Your Skill to The Next Level!" which is a motivational statement encouraging individuals to improve their martial arts abilities. At the bottom of the image, there is a call to action that says "CONTACT US TODAY," suggesting that interested individuals should reach out to the club for more information or to join. The overall style of the image is clean and modern, with a limited color palette that focuses on the martial arts theme. The illustration is likely intended for promotional purposes, aiming to attract potential members to the Black Belt Club.}
    \vspace{1em}
    \end{tabular}
    \begin{tabular}{ccc}
        \includegraphics[width=0.25\textwidth]{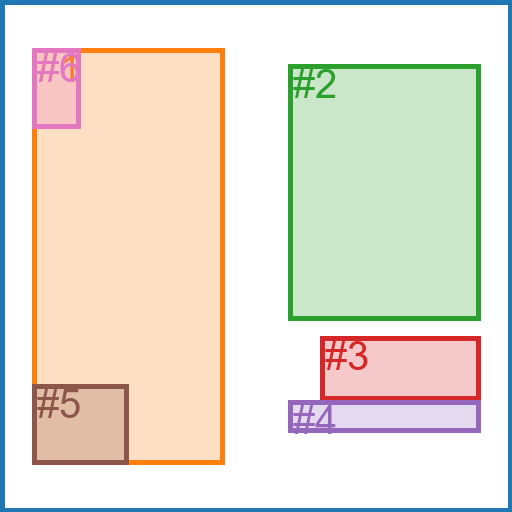} &
        \includegraphics[width=0.25\textwidth]{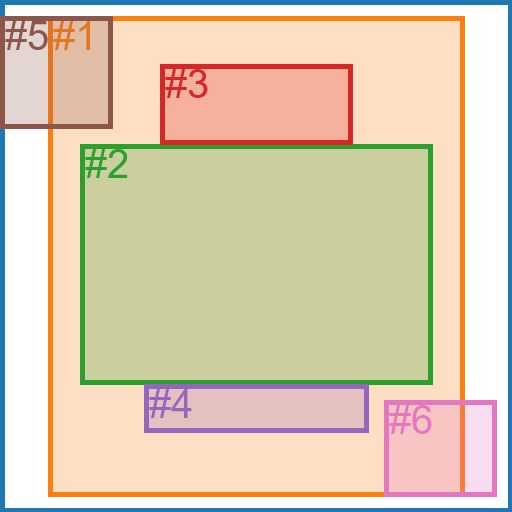} &
        \includegraphics[width=0.25\textwidth]{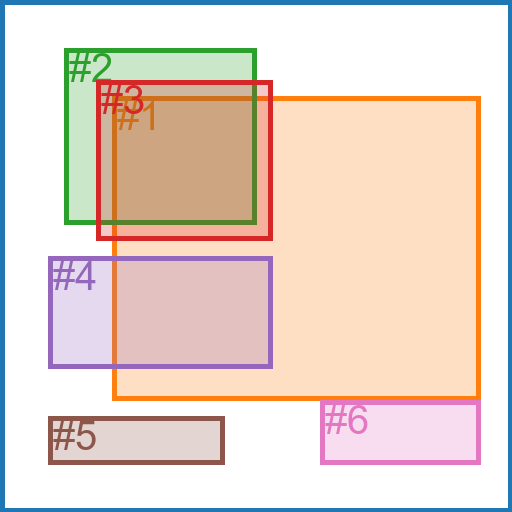} \\
        \small{Layout A} & \small{Layout B} & \small{Layout C} \\[1em]
        \includegraphics[width=0.3\textwidth]{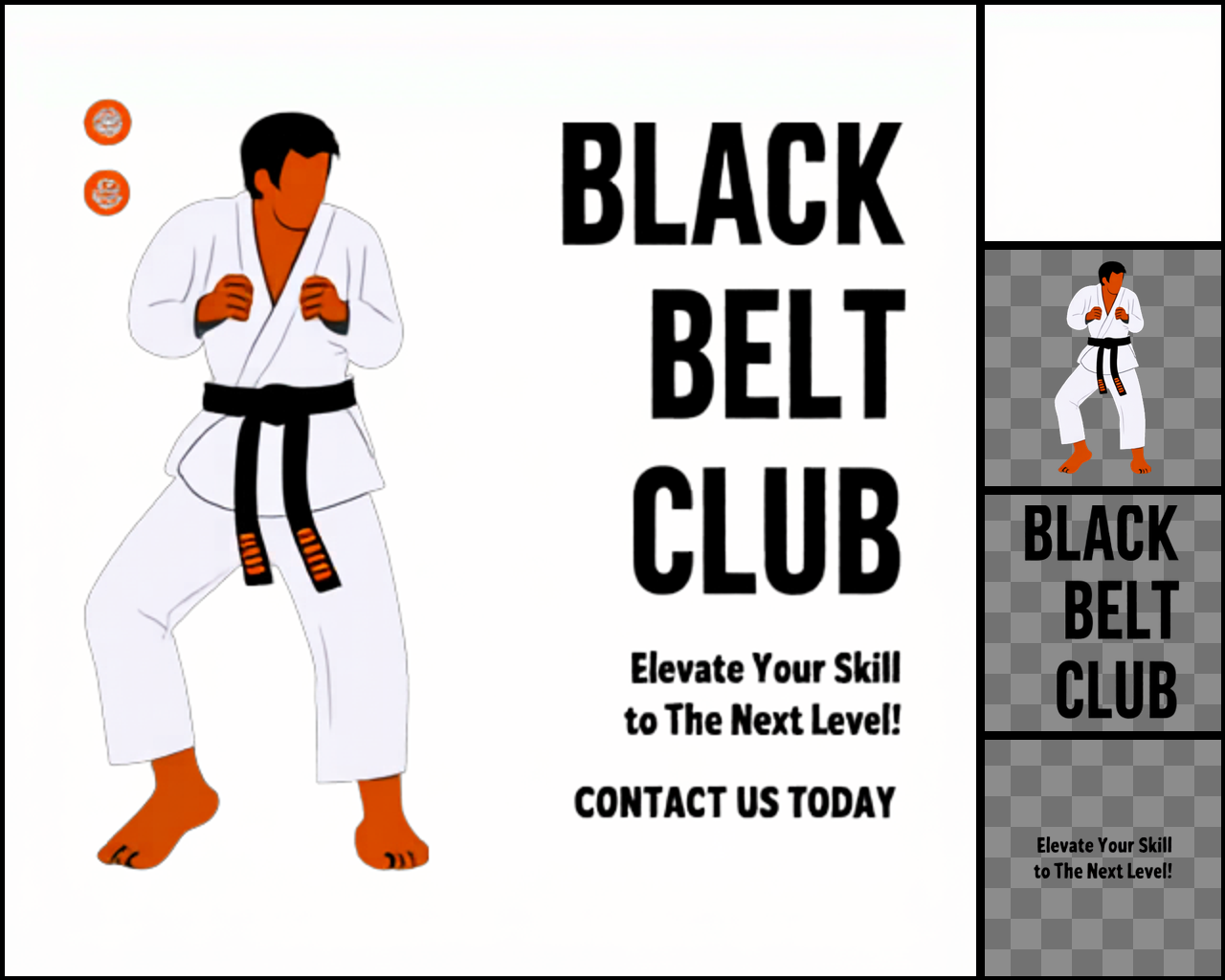} &
        \includegraphics[width=0.3\textwidth]{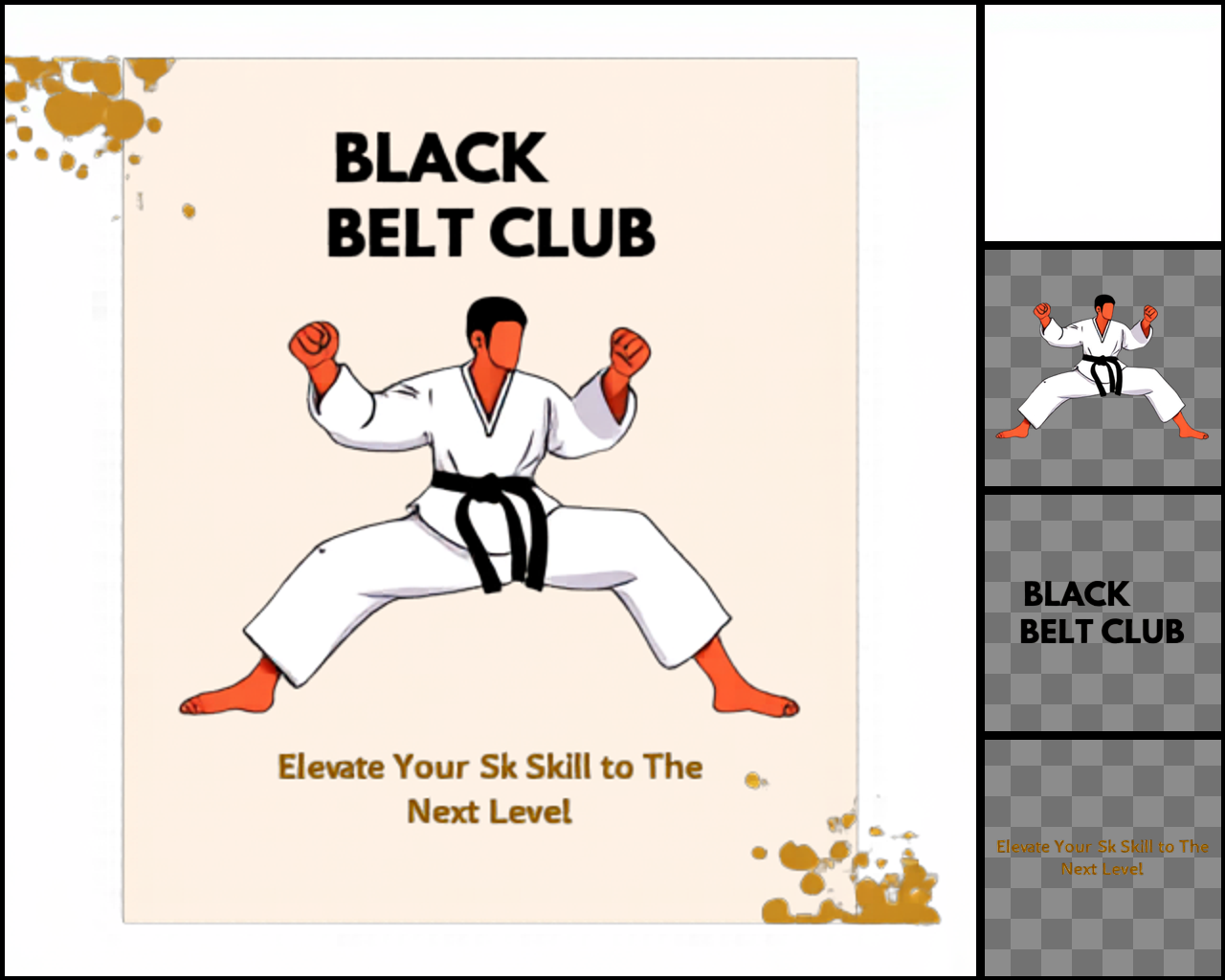} &
        \includegraphics[width=0.3\textwidth]{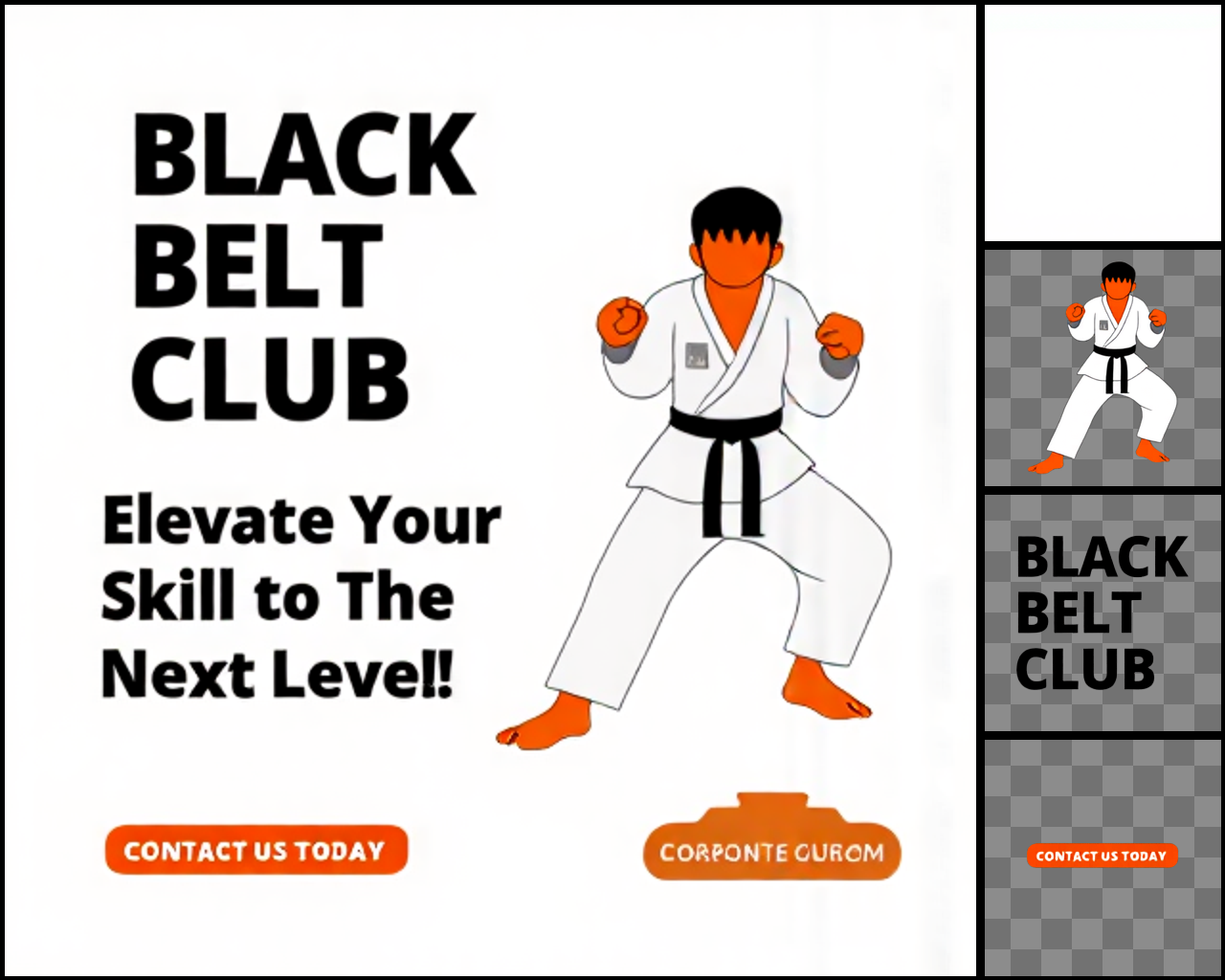} \\
        \small{Generated A} & \small{Generated B} & \small{Generated C} \\
        \bottomrule
    \end{tabular}
    
    \caption{Generated results conditioned on the same prompt and variant layouts. We show the prompt at the first row, three different layouts (the background index `\#0' is omitted) at the second row and the generated results at the last row. (Case 6)}
    \label{tab:variant_layout6}
\end{table}

\begin{table}[htbp]
    \centering
    \begin{tabular}{p{\textwidth}}
    \midrule
    \textbf{Prompt:} \small{The image is a promotional graphic for a knitting service. It features a warm, inviting design with a wooden table as the central focus. On the table, there are various knitting tools and materials, including a pair of hands actively knitting with yarn, a pair of scissors, a cup of coffee, and a bowl of cookies. The background is a rich, dark brown, and there are decorative elements such as swirls and dots in lighter shades of brown and beige. At the top of the image, in large, bold white letters, the text reads "HOW WE KNIT YOUR SWEATERS." Below this, in smaller white font, it says "Learn the ins and outs of all stages." At the bottom of the image, there's a pink banner with white text that states "MADE FOR YOU - MADE WITH CARE." The overall style of the image is cozy and crafty, designed to appeal to those interested in handmade knitwear.}
    \vspace{1em}
    \end{tabular}
    \begin{tabular}{ccc}
        \includegraphics[width=0.25\textwidth]{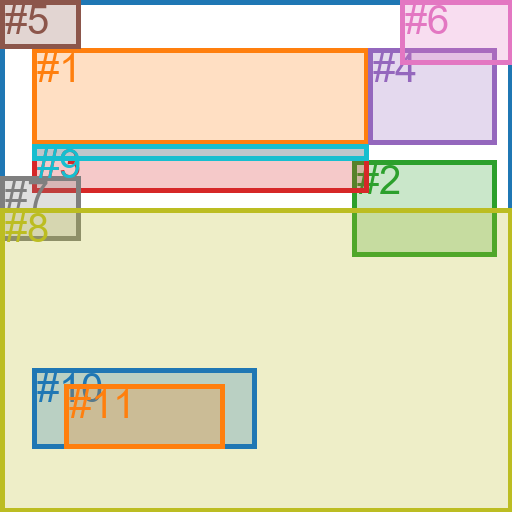} &
        \includegraphics[width=0.25\textwidth]{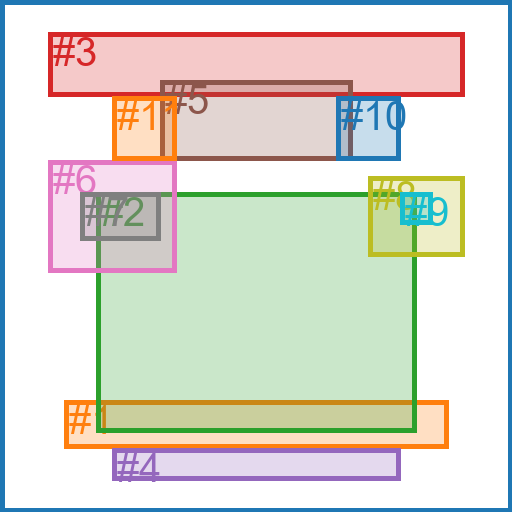} &
        \includegraphics[width=0.25\textwidth]{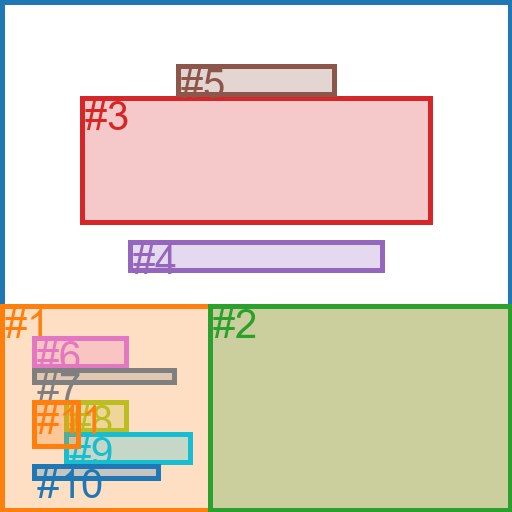} \\
        \small{Layout A} & \small{Layout B} & \small{Layout C} \\[1em]
        \includegraphics[width=0.3\textwidth]{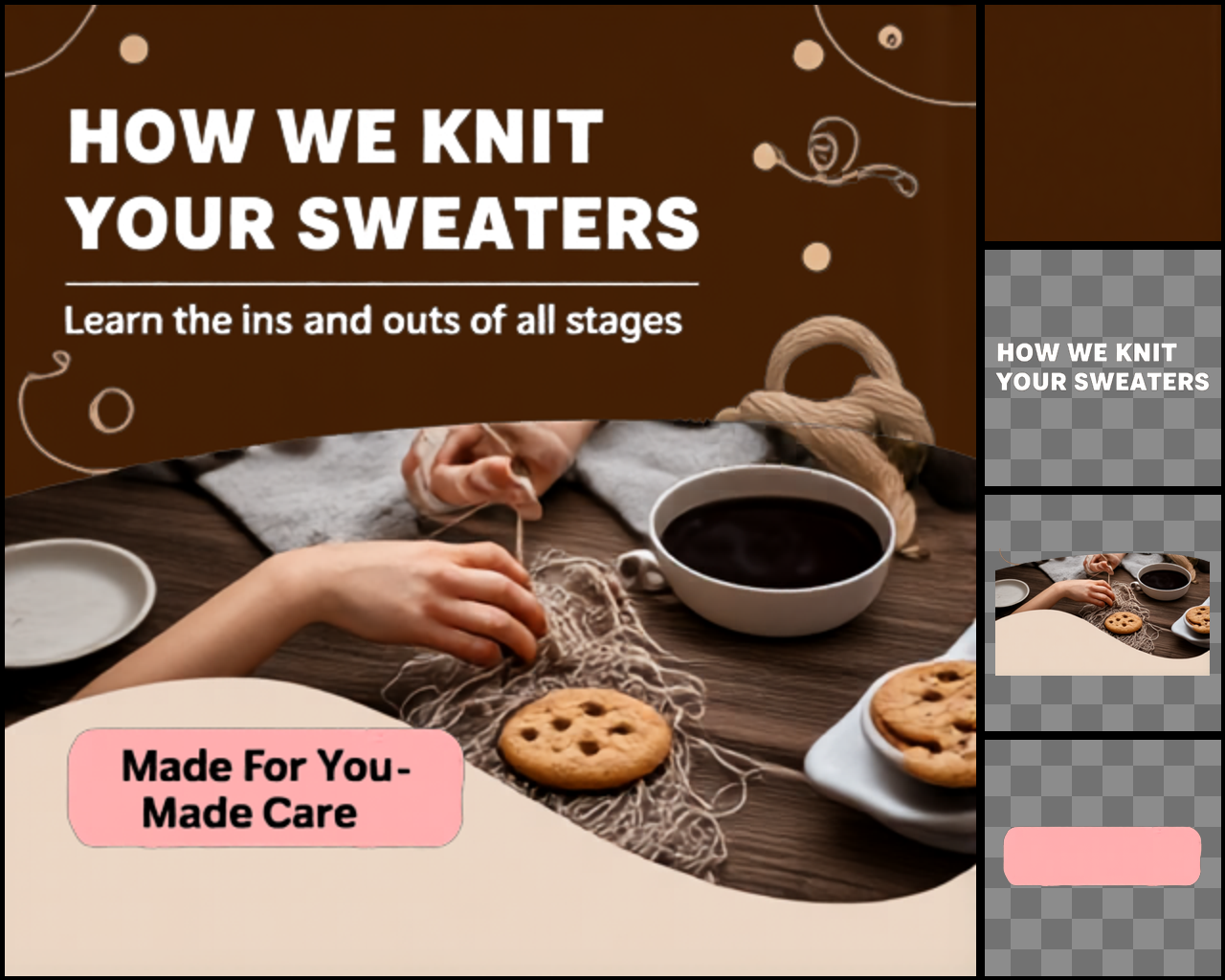} &
        \includegraphics[width=0.3\textwidth]{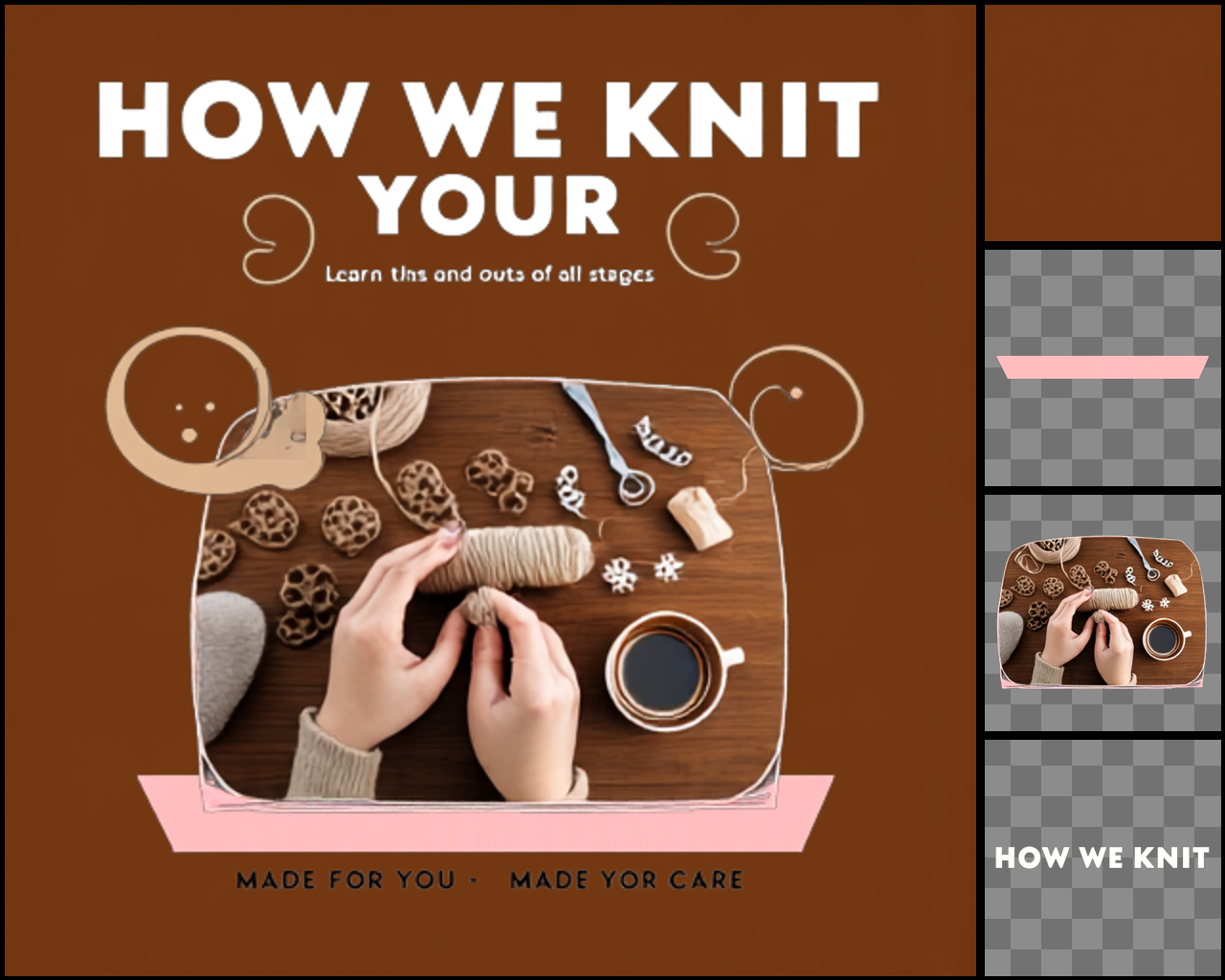} &
        \includegraphics[width=0.3\textwidth]{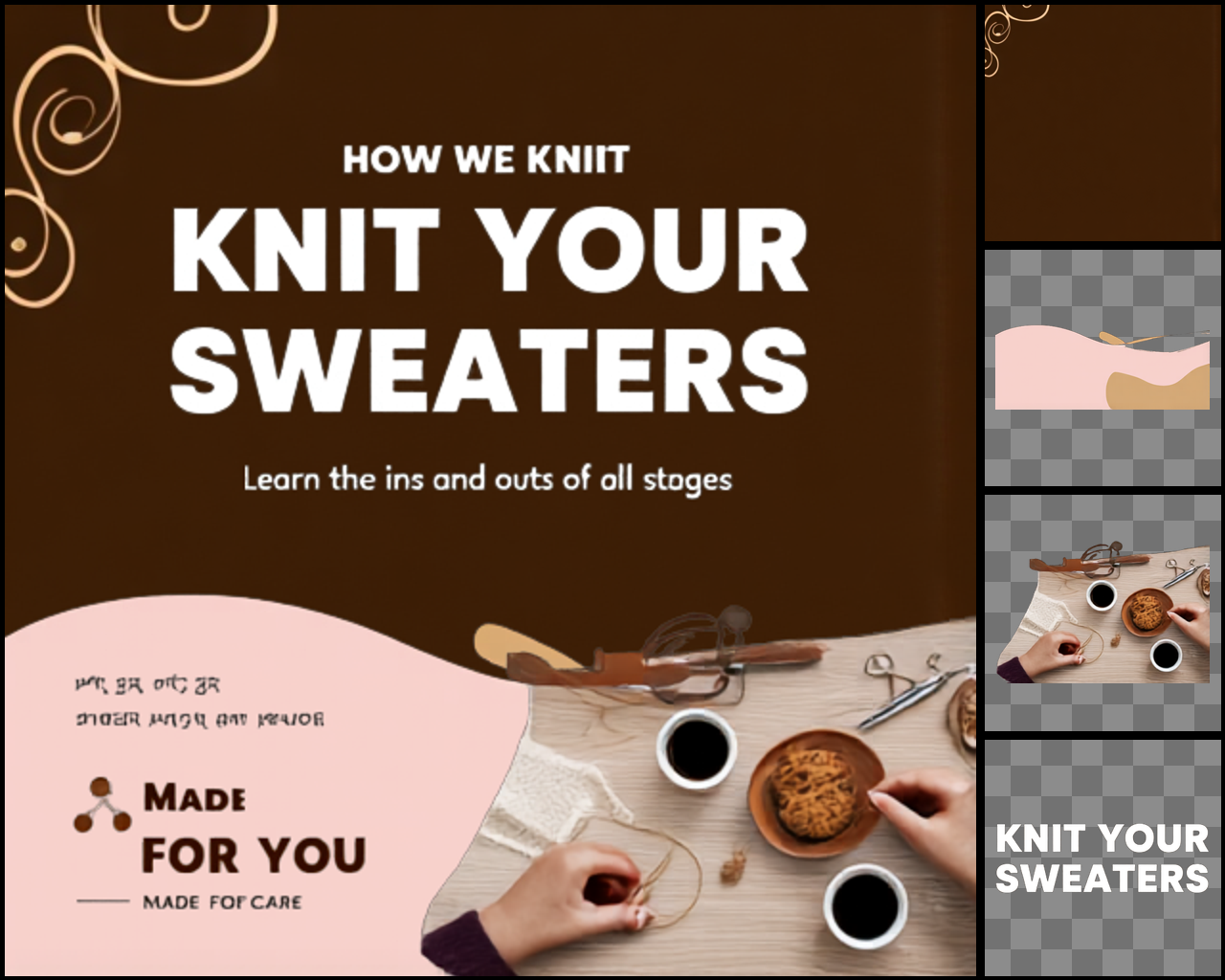} \\
        \small{Generated A} & \small{Generated B} & \small{Generated C} \\
        \bottomrule
    \end{tabular}
    
    \caption{Generated results conditioned on the same prompt and variant layouts. We show the prompt at the first row, three different layouts (the background index `\#0' is omitted) at the second row and the generated results at the last row. (Case 7)}
    \label{tab:variant_layout7}
\end{table}

\begin{table}[htbp]
    \centering
    \begin{tabular}{p{\textwidth}}
    \midrule
    \textbf{Prompt:} \small{The image is a collage of three separate photographs, each depicting a different scene related to hiking and nature. In the top left photograph, there is a text overlay that reads "EXPLORE VIRGINIA'S HIKING TRAILS" in a bold, sans-serif font. The text is green with a slight shadow effect, making it stand out against the white background. The top right photograph features a man wearing a wide-brimmed hat and a light-colored shirt. He is smiling and looking directly at the camera. A green parrot is perched on his shoulder, adding a vibrant splash of color to the scene. The man appears to be outdoors, surrounded by lush greenery, suggesting a natural, possibly tropical, environment. The bottom left photograph shows two individuals, a man and a woman, who are engaged in a hiking activity. The man is wearing a hat and is holding a large, rolled-up map or document, which he seems to be examining. The woman is standing next to him, also wearing a hat, and is looking in the same direction as the man. They are both dressed in casual, outdoor-appropriate clothing. The background is filled with dense foliage, indicating that they are in a forested area. The bottom right photograph contains text that reads "EXO TRAVEL BOOKING ONLINE" in a similar style to the text in the top left photograph. The text is green with a slight shadow effect, and it is positioned against a white background. Overall, the collage seems to be promoting outdoor activities, specifically hiking in Virginia, and is likely associated with a travel company or service. The images are designed to evoke a sense of adventure and connection with nature.}
    \vspace{1em}
    \end{tabular}
    \begin{tabular}{ccc}
        \includegraphics[width=0.25\textwidth]{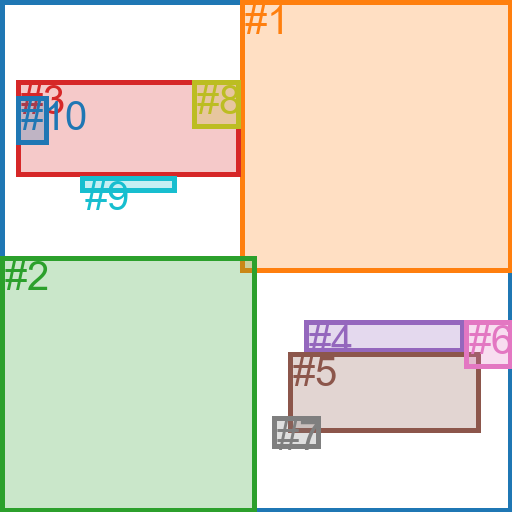} &
        \includegraphics[width=0.25\textwidth]{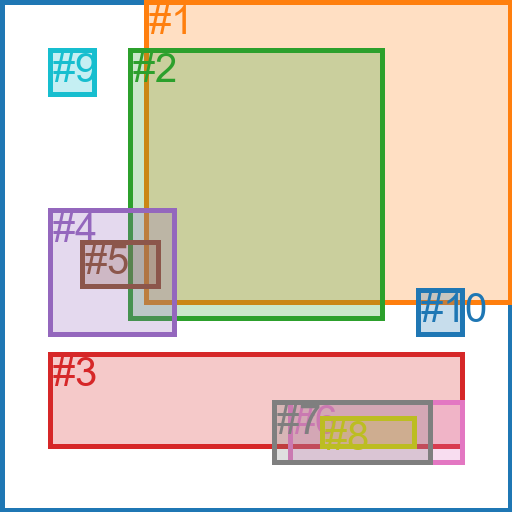} &
        \includegraphics[width=0.25\textwidth]{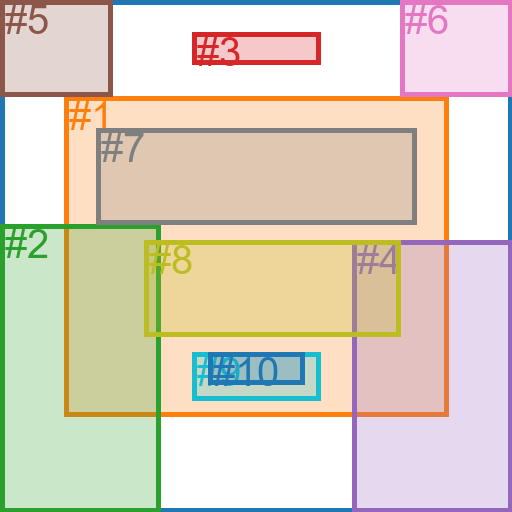} \\
        \small{Layout A} & \small{Layout B} & \small{Layout C} \\[1em]
        \includegraphics[width=0.3\textwidth]{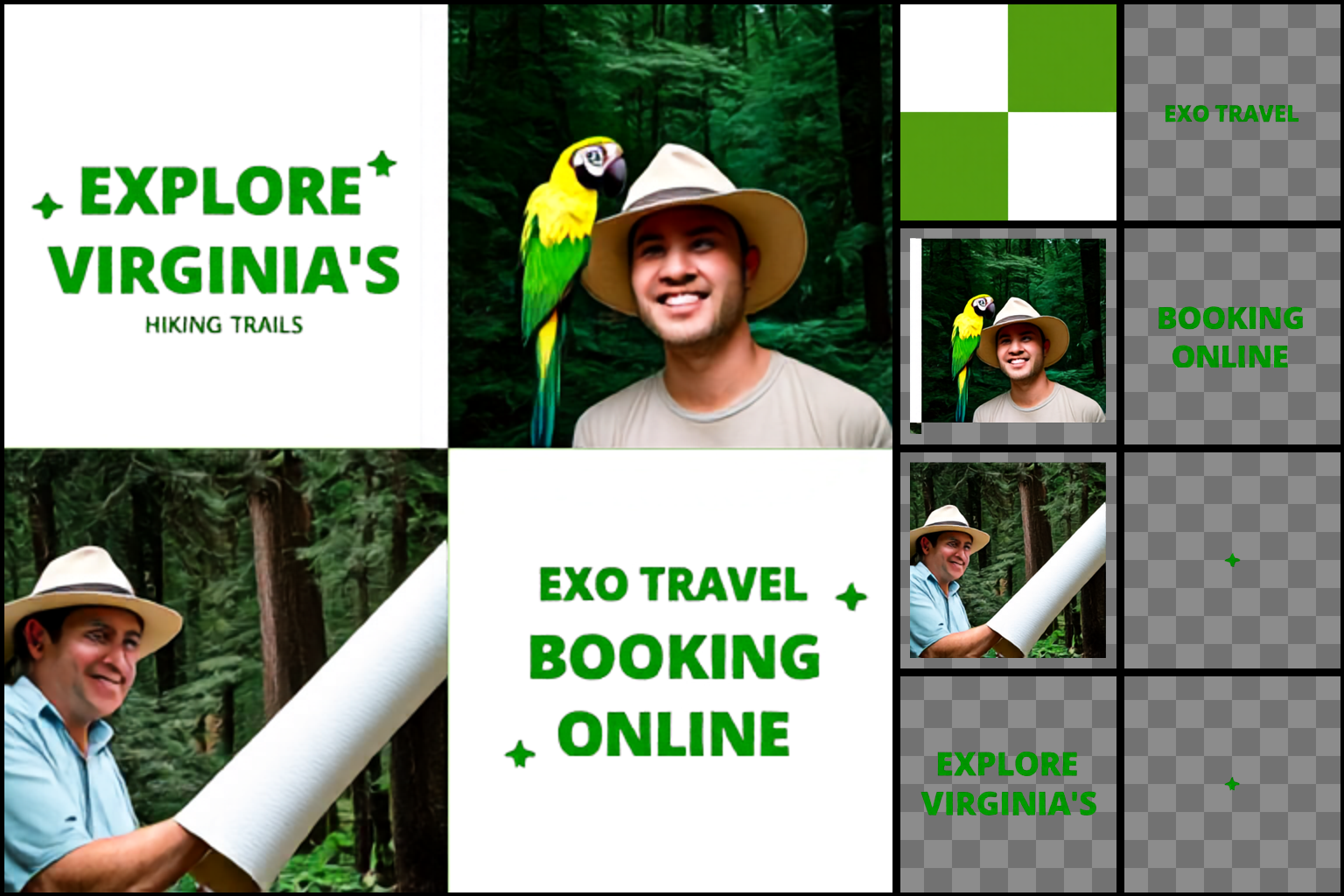} &
        \includegraphics[width=0.3\textwidth]{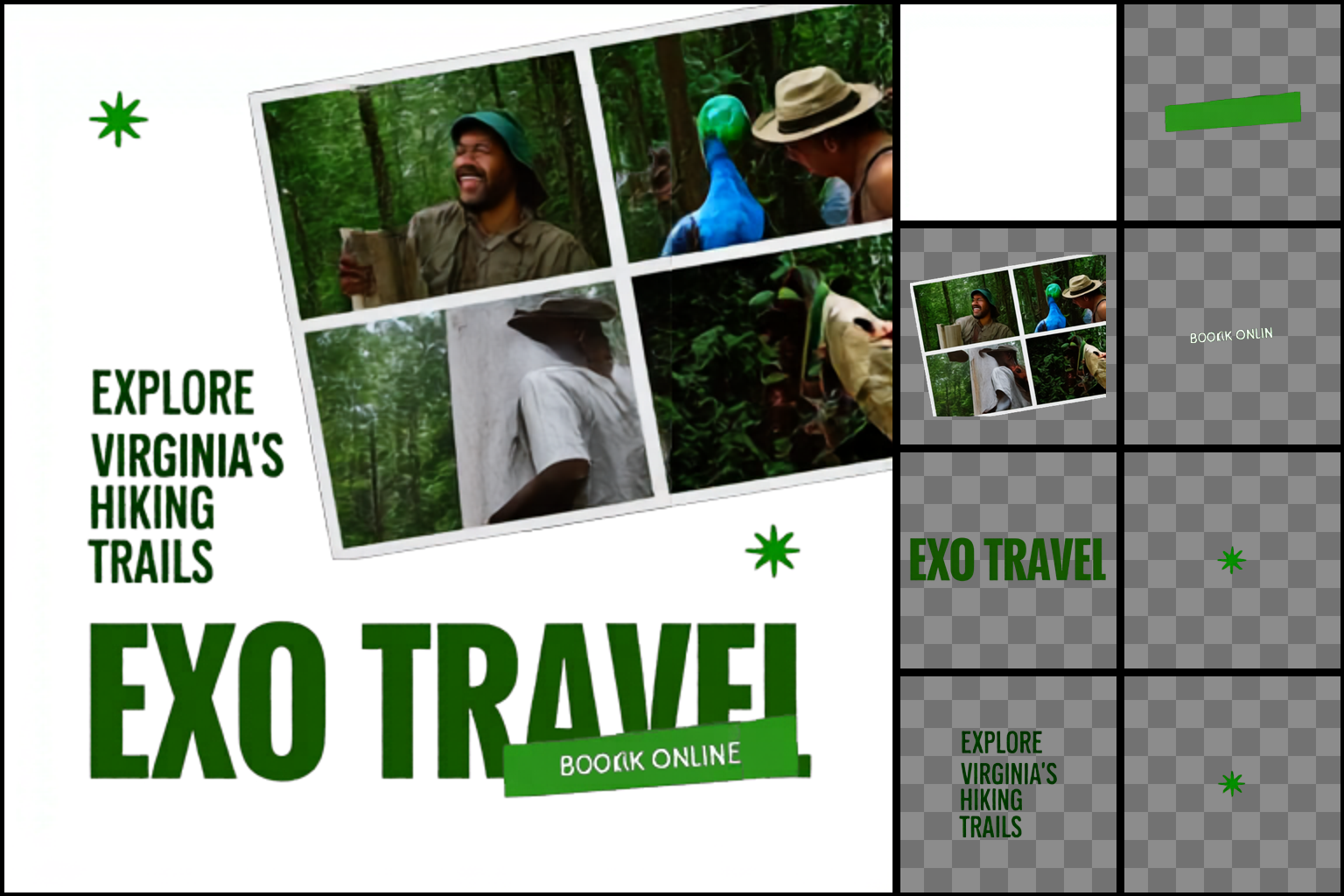} &
        \includegraphics[width=0.3\textwidth]{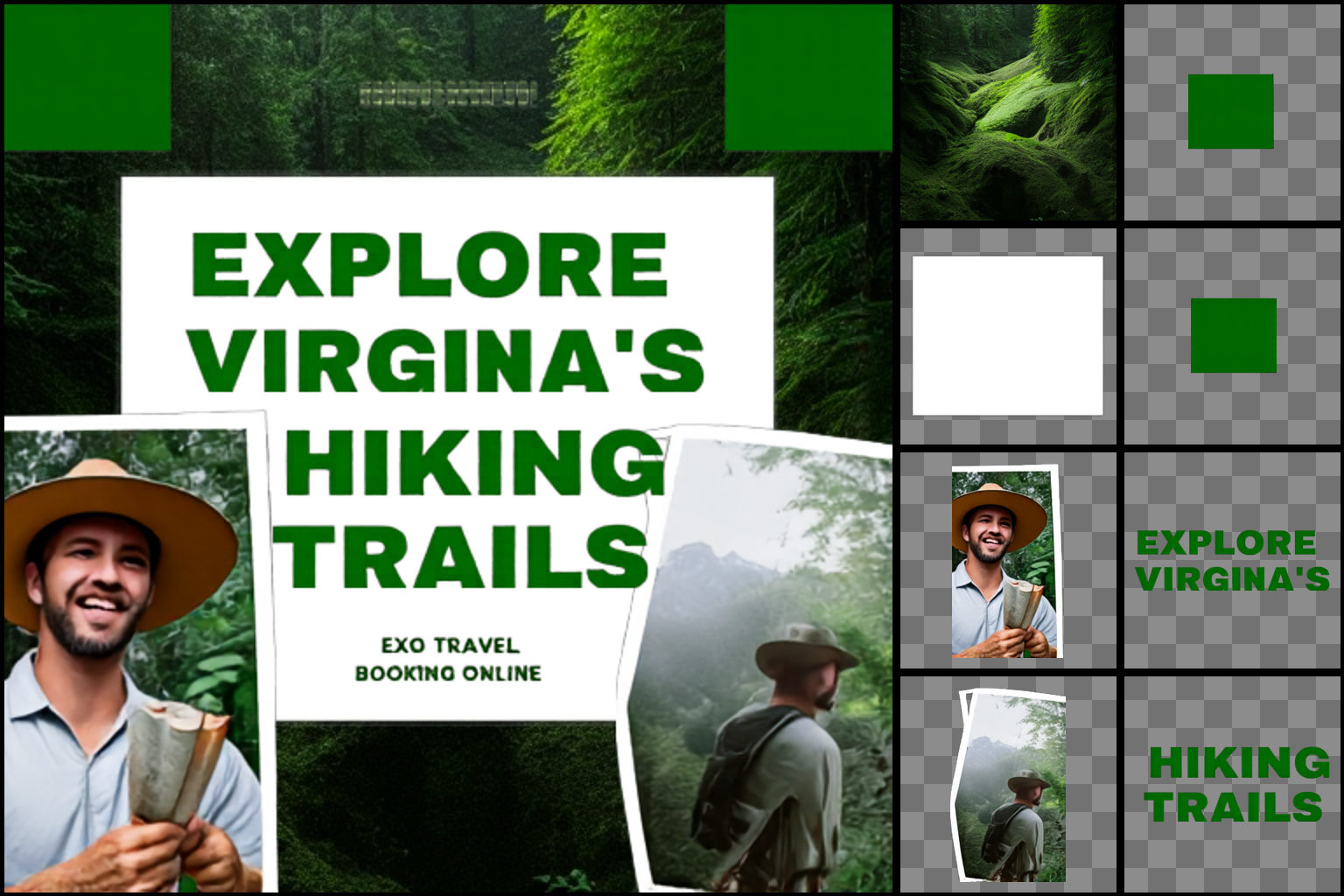} \\
        \small{Generated A} & \small{Generated B} & \small{Generated C} \\
        \bottomrule
    \end{tabular}
    
    \caption{Generated results conditioned on the same prompt and variant layouts. We show the prompt at the first row, three different layouts (the background index `\#0' is omitted) at the second row and the generated results at the last row. (Case 8)}
    \label{tab:variant_layout8}
\end{table}

\begin{table}[htbp]
    \centering
    \begin{tabular}{p{\textwidth}}
    \midrule
    \textbf{Prompt:} \small{The image features a logo for a flower shop named "Estelle Darcy Flower Shop." The logo is designed with a stylized flower, which appears to be a rose, in shades of pink and green. The flower is positioned to the left of the text, which is written in a cursive font. The text is in a brown color, and the overall style of the image is simple and elegant, with a clean, light background that does not distract from the logo itself. The logo conveys a sense of freshness and natural beauty, which is fitting for a flower shop.}
    \vspace{1em}
    \end{tabular}
    \begin{tabular}{ccc}
        \includegraphics[width=0.25\textwidth]{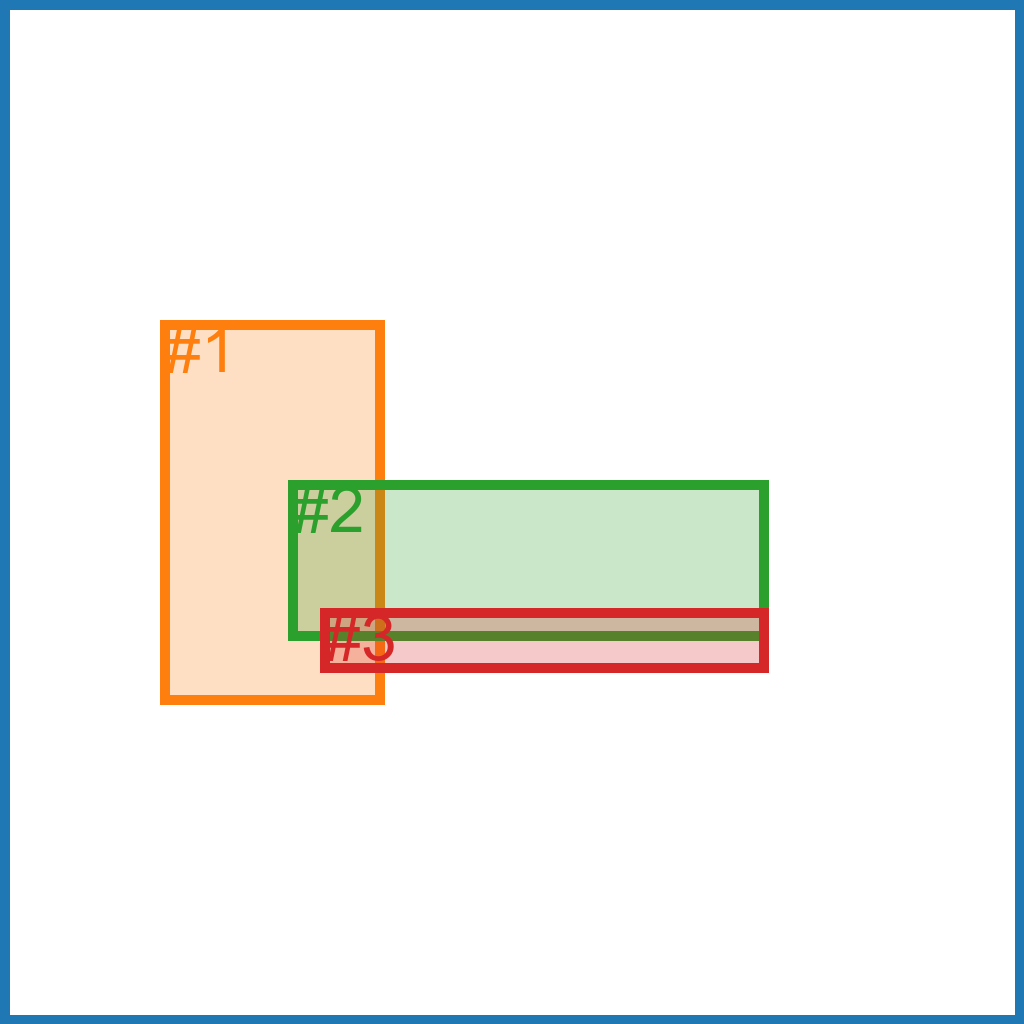} &
        \includegraphics[width=0.25\textwidth]{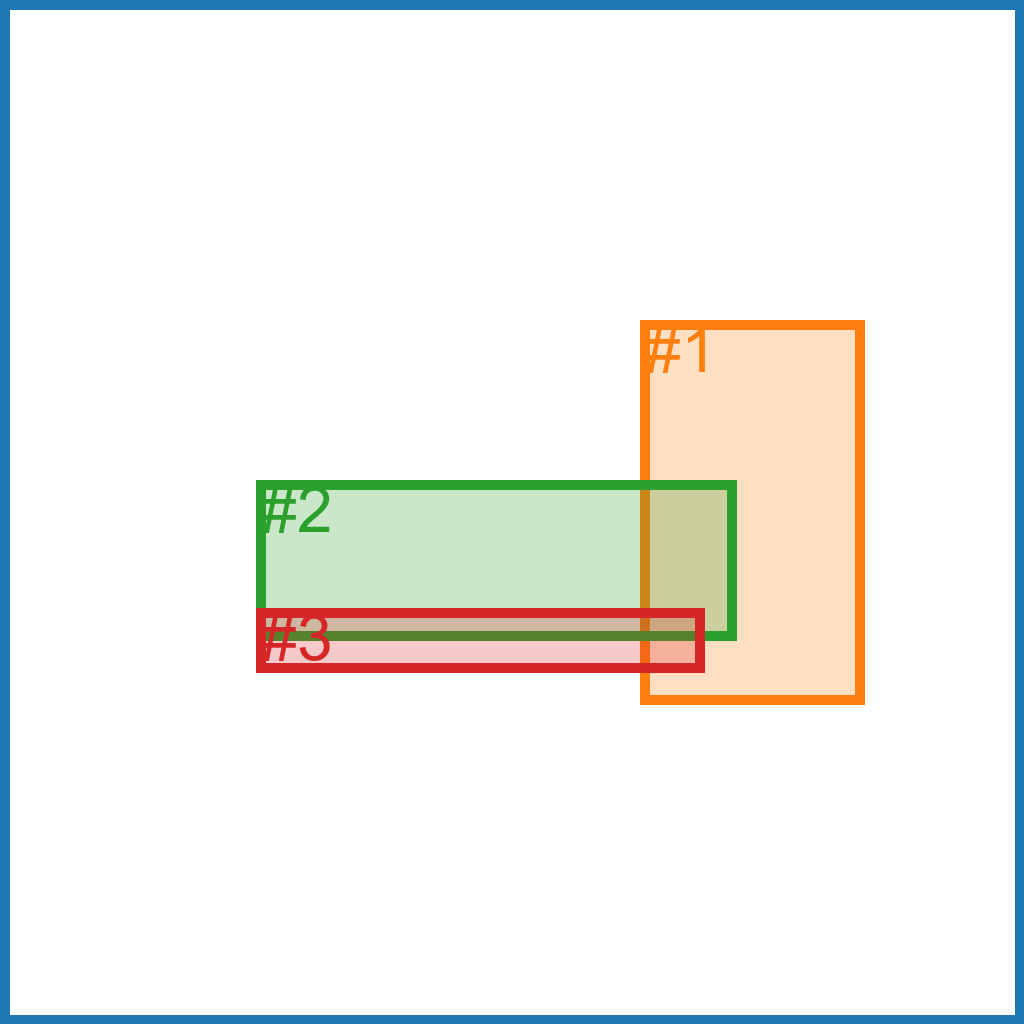} &
        \includegraphics[width=0.25\textwidth]{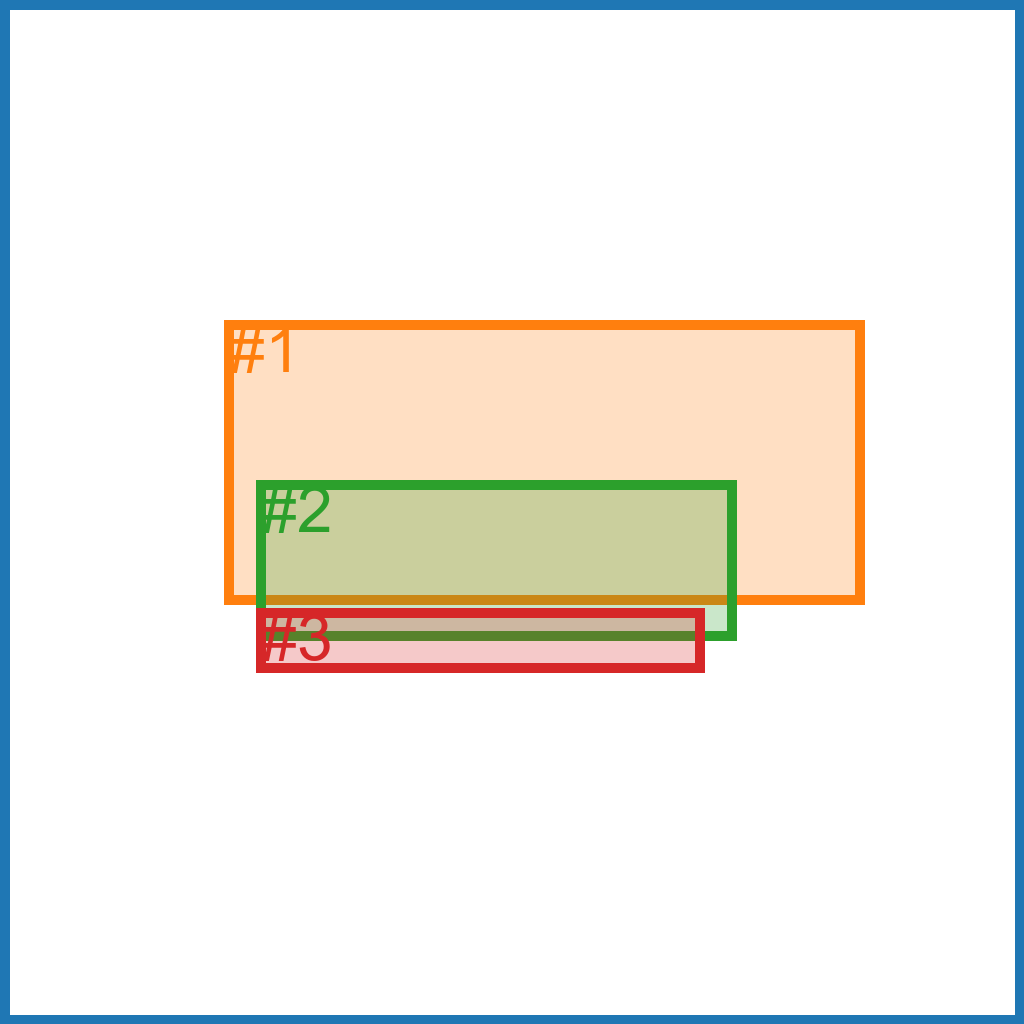} \\
        \small{Layout A} & \small{Layout B} & \small{Layout C} \\[1em]
        \includegraphics[width=0.3\textwidth]{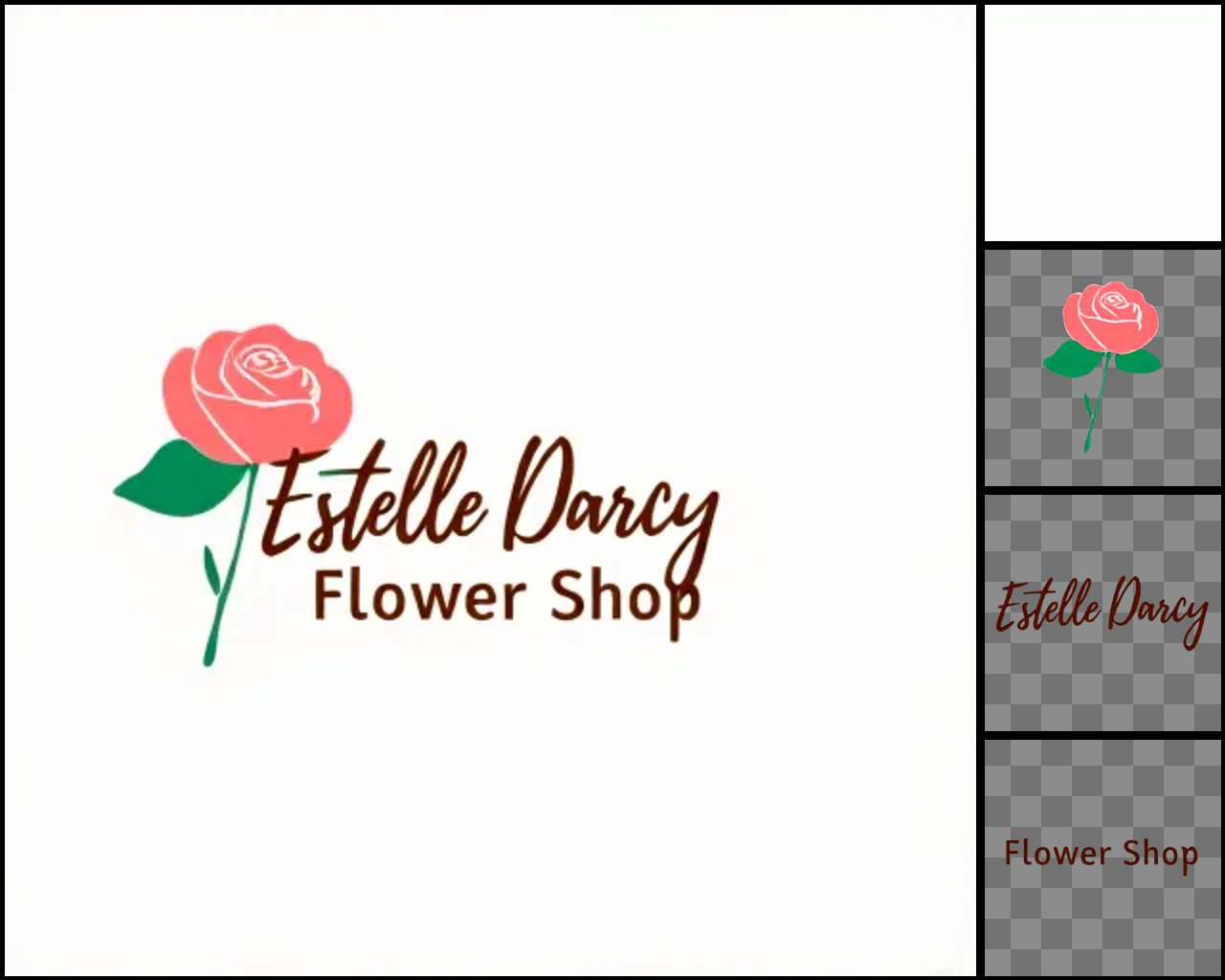} &
        \includegraphics[width=0.3\textwidth]{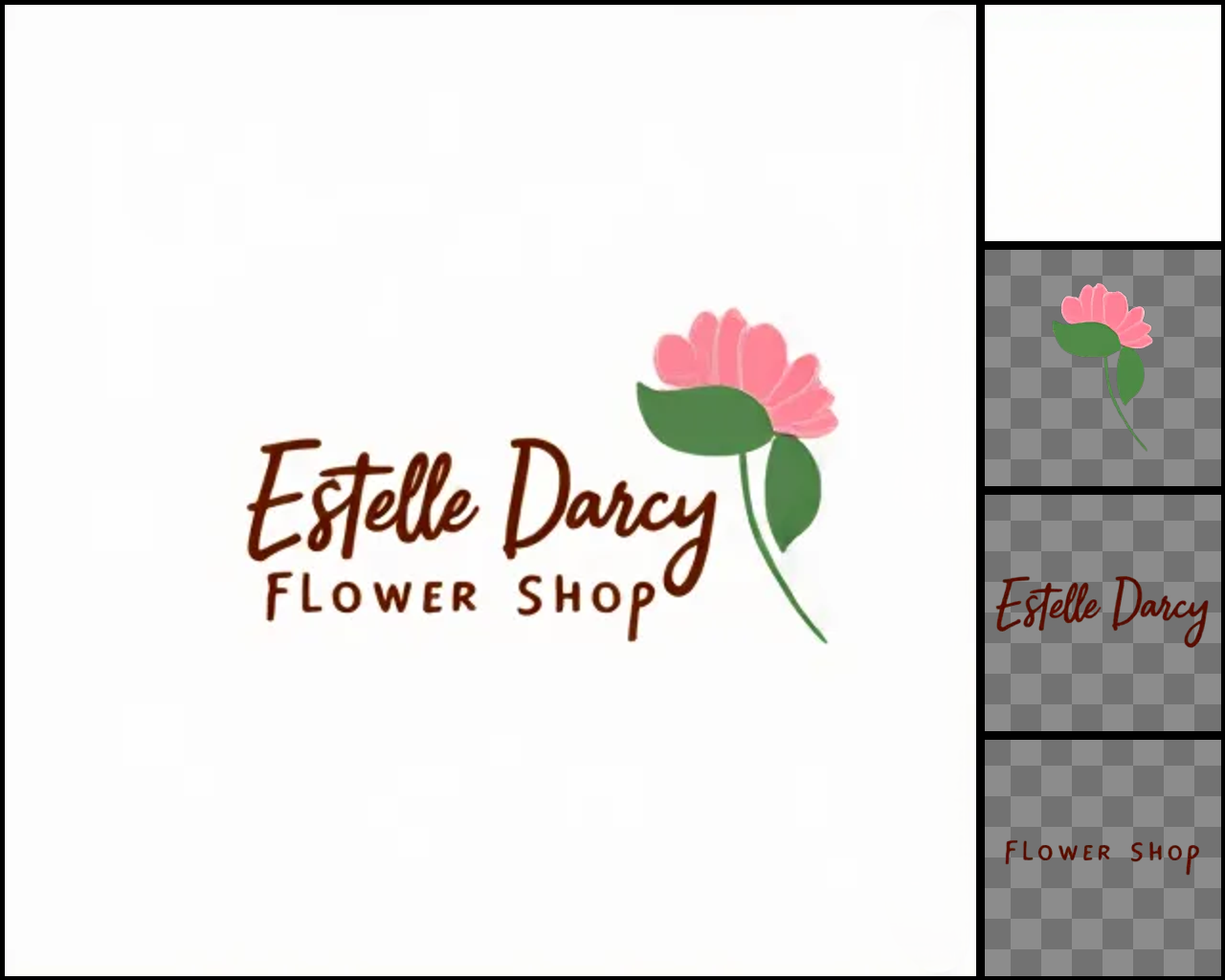} &
        \includegraphics[width=0.3\textwidth]{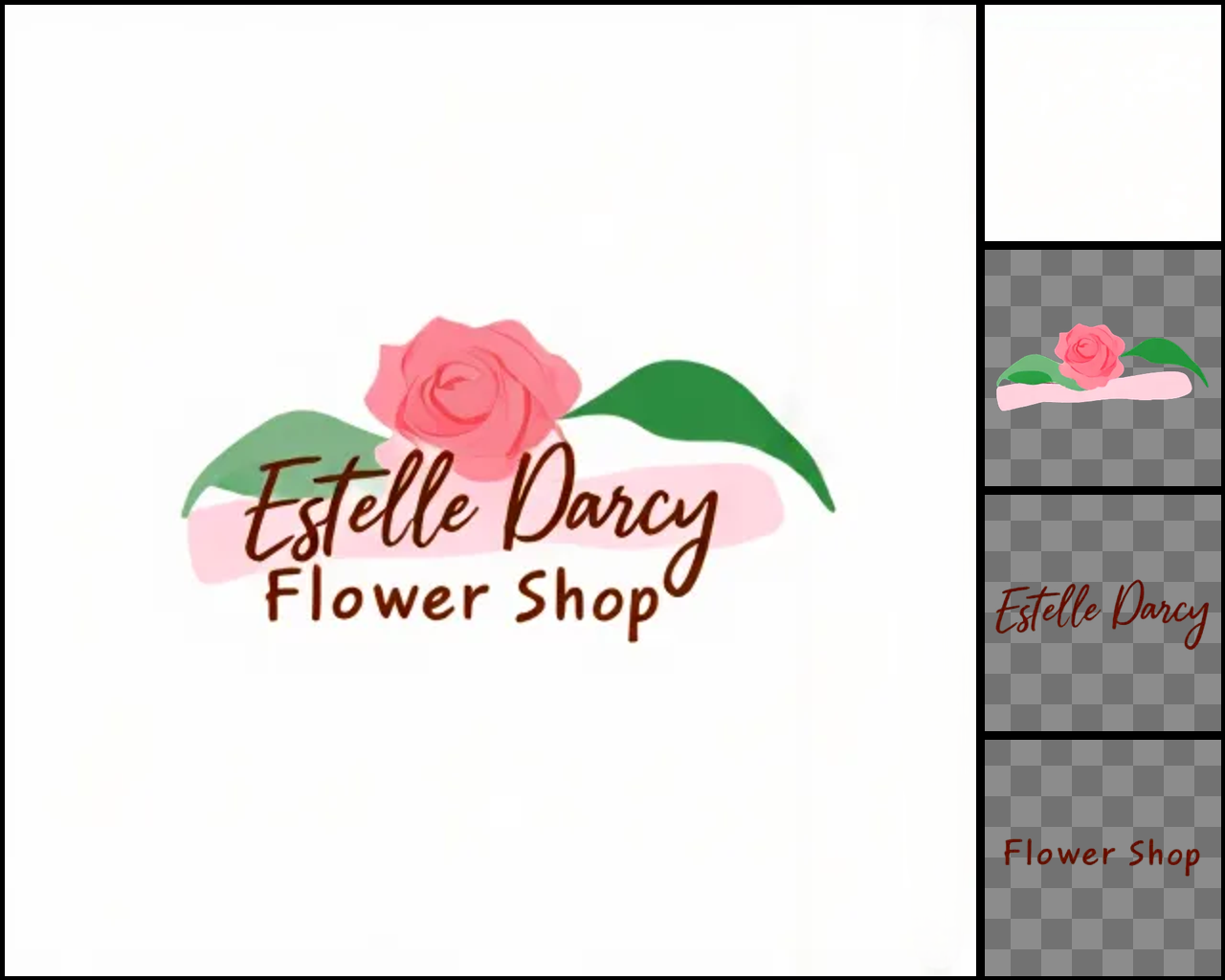} \\
        \small{Generated A} & \small{Generated B} & \small{Generated C} \\
        \bottomrule
    \end{tabular}
    
    \caption{Generated results conditioned on the same prompt and variant layouts. We show the prompt at the first row, three different layouts (the background index `\#0' is omitted) at the second row and the generated results at the last row. (Case 9)}
    \label{tab:variant_layout9}
\end{table}

\clearpage
\newpage

\end{document}